\documentclass[table]{ai2style/ai2}

\usepackage{microtype}
\usepackage{hyperref}
\usepackage{url}
\usepackage{booktabs, tabularx} 
\usepackage{graphicx}
\usepackage{lineno}
\usepackage{enumitem}
\usepackage{listings} 
\usepackage{svg}

\definecolor{ darkblue}{rgb}{0, 0, 0.5}
\hypersetup{colorlinks=true, citecolor=darkblue, linkcolor=darkblue, urlcolor=darkblue}

\usepackage{amssymb}
\usepackage{multirow}
\usepackage{bigdelim}
\usepackage{todonotes}
\usepackage{longtable}
\usepackage{tabularray}
\usepackage{wrapfig}
\usepackage[most]{tcolorbox}
\usepackage{url}
\usepackage{xspace}
\usepackage{svg}
\usepackage[absolute]{textpos} 

\usepackage{fdsymbol}   

\usepackage[utf8]{inputenc} 
\usepackage[T1]{fontenc}    
\usepackage{url}            
\usepackage{booktabs}       
\usepackage{amsfonts}       
\usepackage{nicefrac}       
\usepackage{microtype}      
\usepackage[table,xcdraw]{xcolor}         
\usepackage{amsmath}
\usepackage[most]{tcolorbox}
\usepackage{csquotes}
\usepackage{tablefootnote}

\usepackage{siunitx}
\usepackage{graphicx}
\usepackage{arydshln}
\usepackage{wrapfig}
\usepackage{enumitem}
\usepackage{soul} 

\definecolor{baselinecolor}{gray}{.9}
\newcommand{\baseline}[1]{\cellcolor{baselinecolor}{#1}}

\usepackage{multirow}
\usepackage{graphicx}
\usepackage{listings}
\lstdefinelanguage{Prompt}{
  morestring=[b]",
}
\definecolor{codebg}{RGB}{245,245,245}
\usepackage{multirow}
\usepackage{xspace}
\usepackage{adjustbox}
\usepackage{pifont}
\usepackage{caption}
\usepackage{makecell}
\usepackage{subcaption}
\usepackage{bold-extra}
\usepackage{url}

\usepackage{pgf-pie}

\usepackage{hyperref}
\definecolor{linkcolor}{RGB}{0, 0, 128}
\hypersetup{
     colorlinks   = true,
     citecolor    = linkcolor,
     linkcolor    = linkcolor,
     urlcolor     = linkcolor,
}
\usepackage{pifont}
\usepackage{listings}

\setlist[itemize]{leftmargin=*,itemsep=0em,parsep=0.3em,topsep=0.3em}

\DeclareUnicodeCharacter{2212}{\ensuremath{-}}

\definecolor{maroon}{HTML}{F26035}
\definecolor{yellow}{HTML}{FDBC42}
\definecolor{lavender}{HTML}{734f96}
\definecolor{darkergrey}{HTML}{444444}
\definecolor{midgrey}{HTML}{e6eded}
\definecolor{ai2pink}{HTML}{f0529c}
\definecolor{ai2midpink}{HTML}{fad3e5}
\definecolor{ai2lightpink}{HTML}{fbecf3}
\definecolor{ai2midwhite}{HTML}{f2e5d9}
\definecolor{ai2offwhite}{HTML}{fbf4ee}
\definecolor{ai2green}{HTML}{0fcb8c}
\definecolor{ai2lightgreen}{HTML}{e7f9f3}
\definecolor{ai2darkgreen}{HTML}{105257}
\definecolor{ai2purple}{HTML}{B932EB}
\definecolor{ai2lightpurple}{HTML}{f7e8fc}
\definecolor{neutralEight}{HTML}{343434}
\definecolor{neutralFive}{HTML}{838383}
\definecolor{neutralThree}{HTML}{bebebe}
\definecolor{neutralOne}{HTML}{dedede}
\definecolor{lightgrey}{HTML}{fafcfc}
\definecolor{plum}{rgb}{0.56,0.27,0.52}

\usepackage{tikz}

\definecolor{maroon}{HTML}{F26035}
\definecolor{yellow}{HTML}{FDBC42}
\definecolor{darkred}{RGB}{156, 39, 33}
\definecolor{darkblue}{RGB}{31, 90, 153}
\definecolor{forestgreen}{rgb}{0.13, 0.55, 0.13}
\definecolor{brickred}{rgb}{0.8, 0.25, 0.33}
\definecolor{olmoDarkBlue}{HTML}{012e59}
\definecolor{olmoBlue}{HTML}{265ed4}
\definecolor{olmoLightBlue}{HTML}{012e59}
\definecolor{olmoTeal}{HTML}{00d5ff}
\definecolor{olmoYellow}{HTML}{ffbb00}
\definecolor{olmoOrange}{HTML}{ff9100}

\newcommand{\app}{\raise.17ex\hbox{$\scriptstyle\sim$}}
\renewcommand{\paragraph}[1]{\vspace{0.5mm}\noindent\textbf{#1}}
\newcommand{\tablestyle}[2]{\setlength{\tabcolsep}{#1}\renewcommand{\arraystretch}{#2}\centering\footnotesize}

\newcommand{\x}{{\times}}

\definecolor{molmocolor}{RGB}{240, 82, 156}
\definecolor{tablegray}{RGB}{223, 242, 252}
\definecolor{tablegreen}{RGB}{15, 203, 150}
\definecolor{tableyellow}{RGB}{250, 242, 233}
\definecolor{tableblue}{RGB}{240, 82, 156}
\definecolor{darkpink}{RGB}{139, 14, 98}

\usepackage{setspace}

\usepackage{nicematrix}
\newcolumntype{L}[1]{>{\raggedright\let\newline\\\arraybackslash\hspace{0pt}}m{#1}}
\newcolumntype{C}[1]{>{\centering\let\newline\\\arraybackslash\hspace{0pt}}m{#1}}
\newcolumntype{R}[1]{>{\raggedleft\let\newline\\\arraybackslash\hspace{0pt}}m{#1}}
\newcolumntype{P}[1]{>{\centering\let\newline\\\arraybackslash\columncolor{ai2lightpink}}m{#1}}
\newcommand{\aitoo}{\raisebox{-1.5pt}{\includegraphics[height=1.05em]{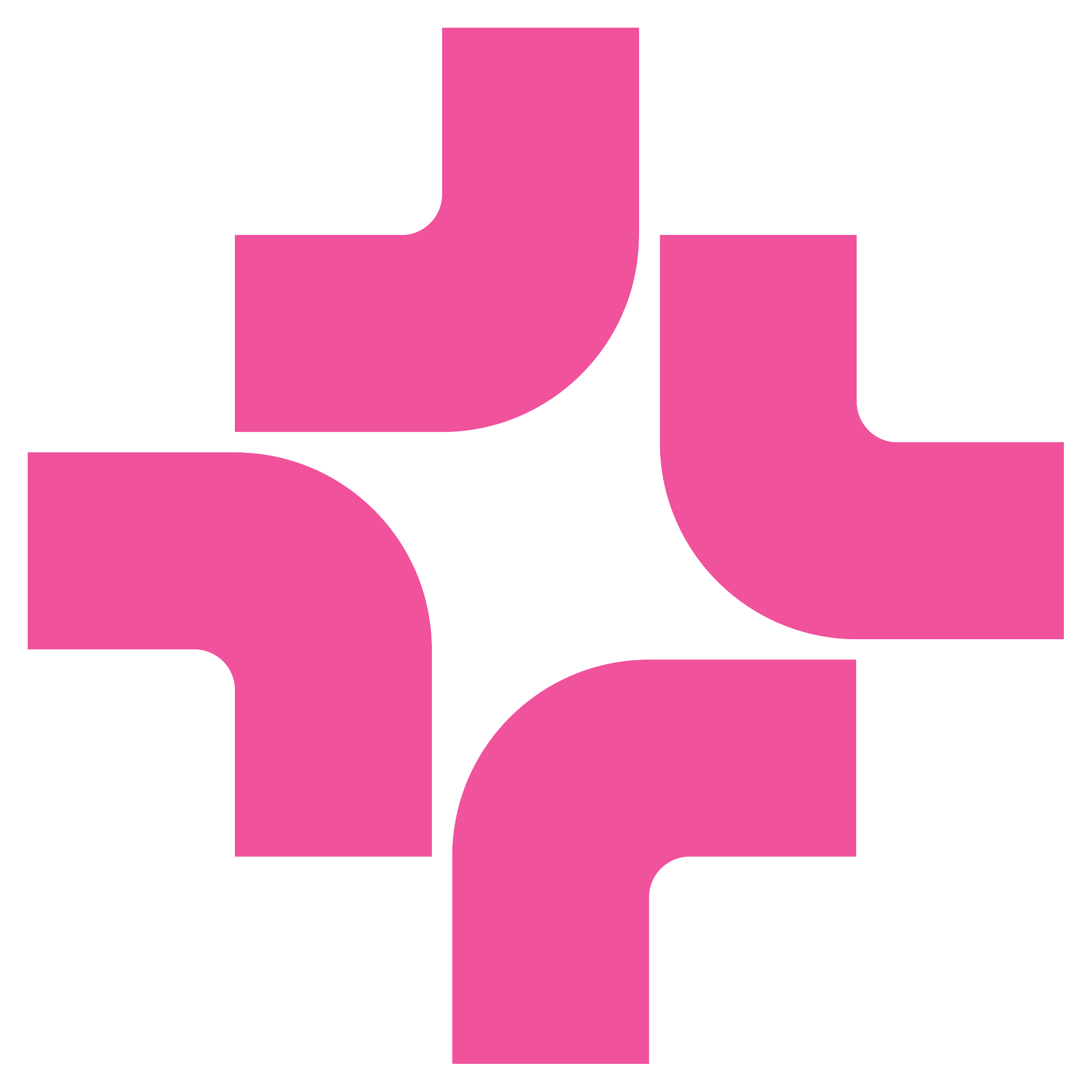}}\xspace}

\newcommand{\huggingface}{\raisebox{-1.5pt}{\includegraphics[height=1.05em]{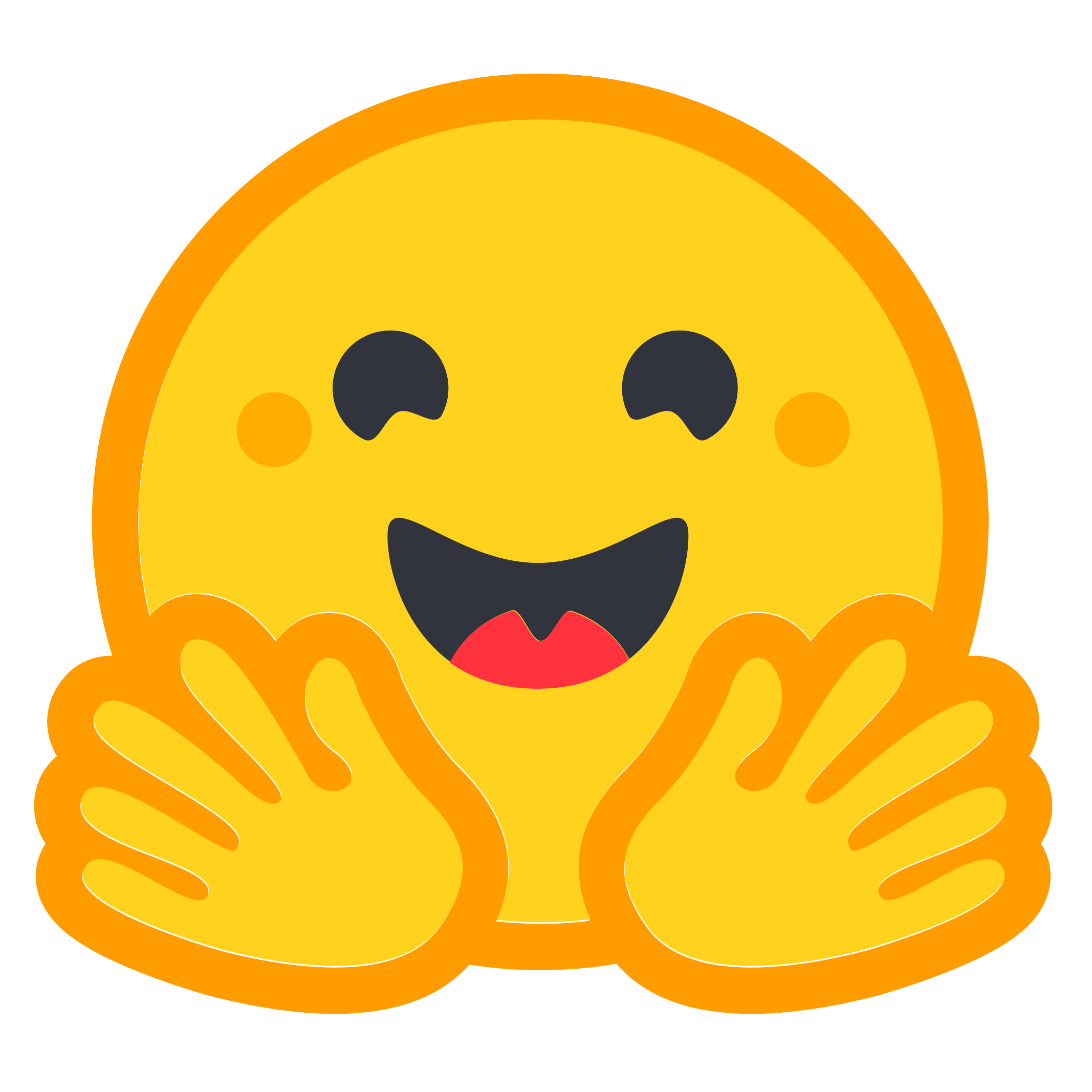}}\xspace}
\newcommand{\hfdataset}{\raisebox{-1.5pt}{\includegraphics[height=1.05em]{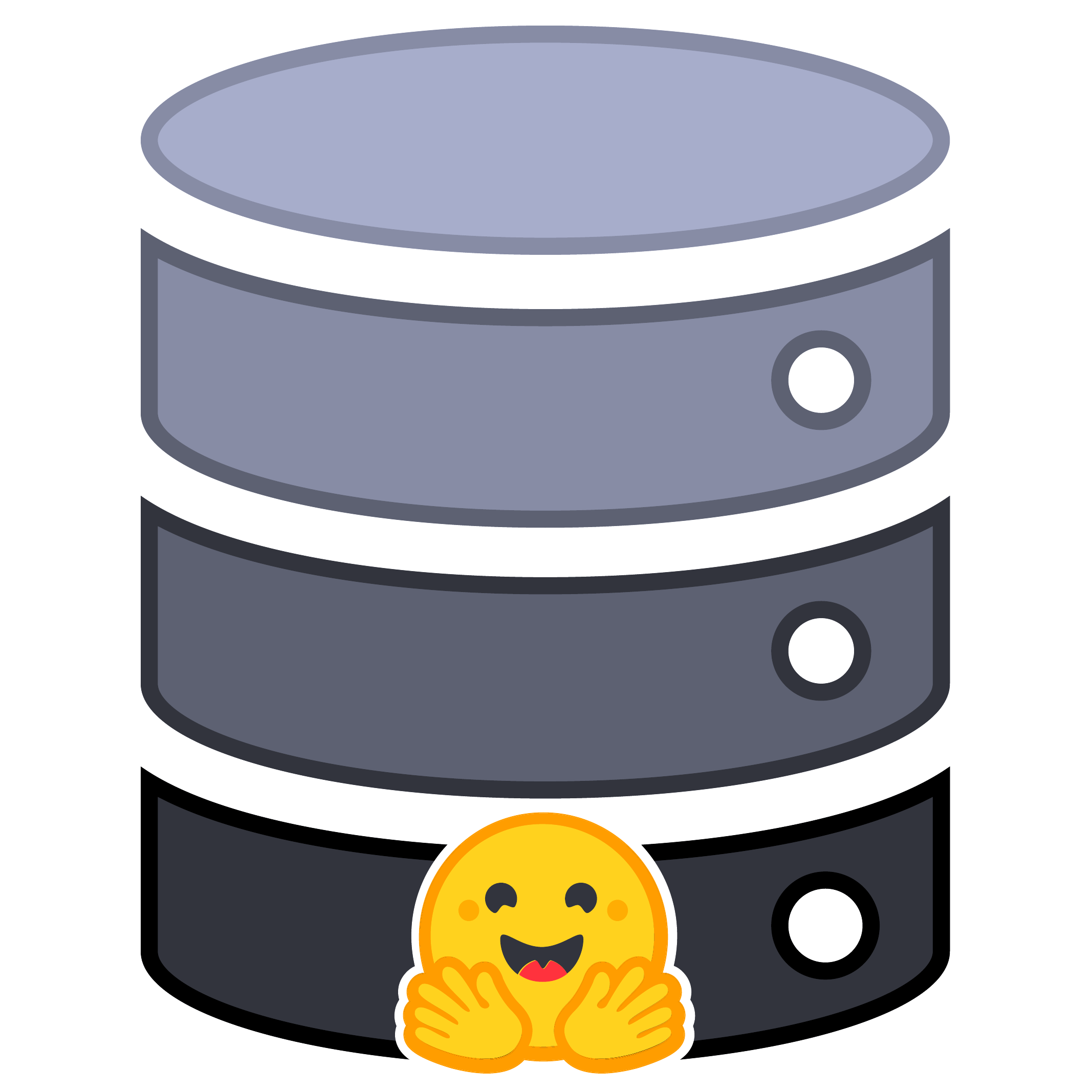}}\xspace}
\newcommand{\emailLogo}{\raisebox{-1.5pt}{\includegraphics[height=1.05em]{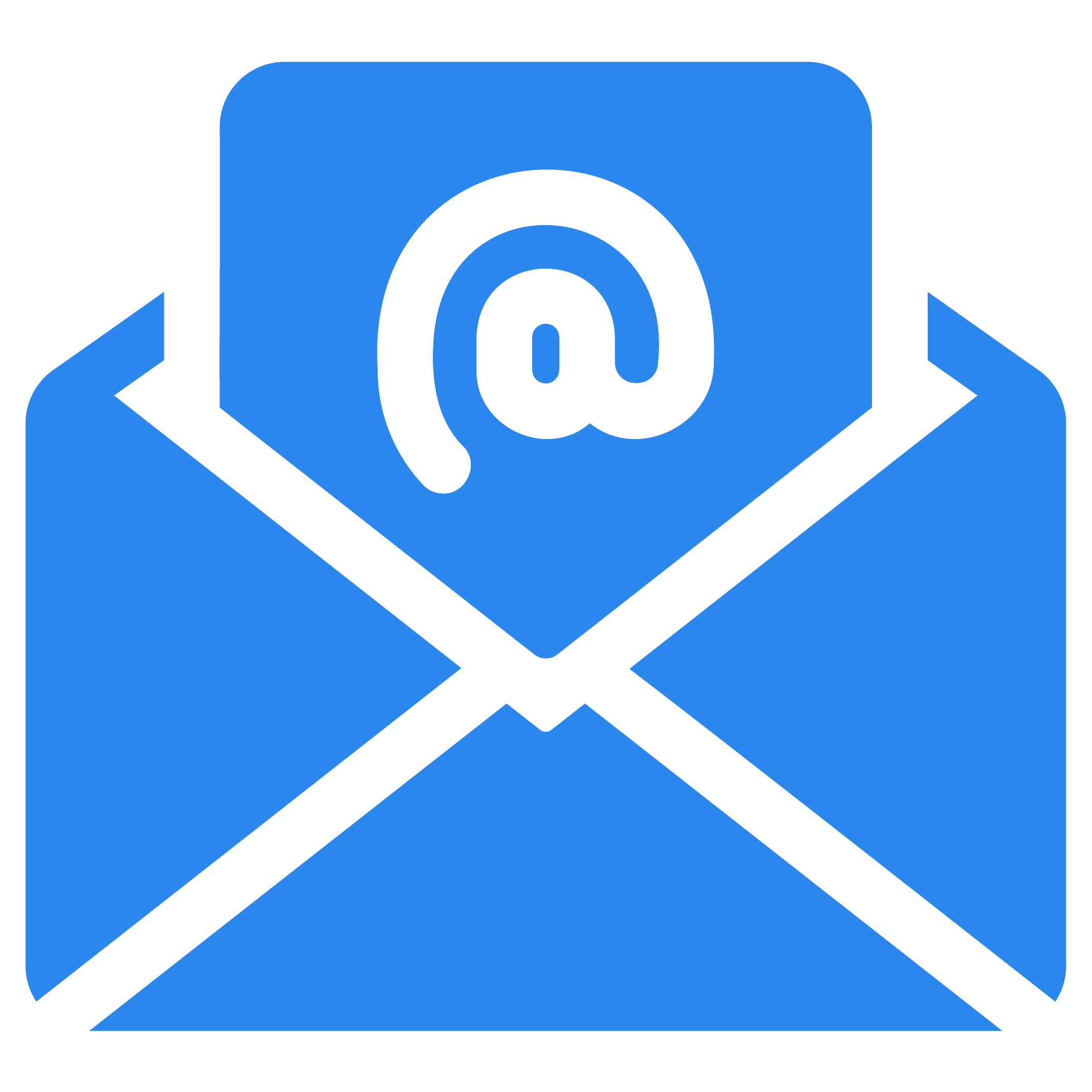}}\xspace}
\newcommand{\github}{\raisebox{-1.5pt}{\includegraphics[height=1.05em]{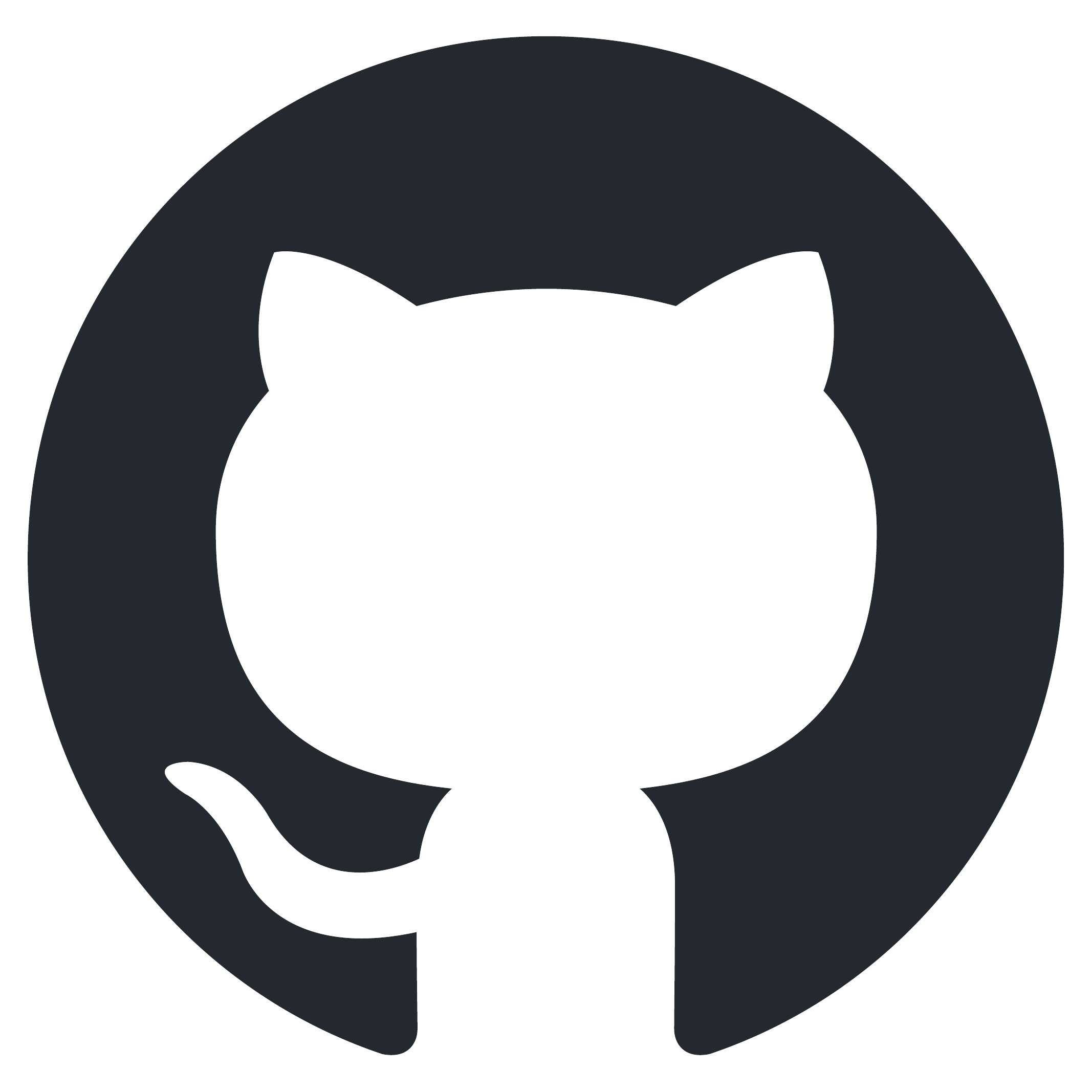}}\xspace}

\makeatletter
\DeclareRobustCommand\onedot{\futurelet\@let@token\@onedot}
\def\@onedot{\ifx\@let@token.\else.\null\fi\xspace}
\def\eg{\emph{e.g}\onedot}

\def\ie{\emph{i.e}\onedot}

\def\etc{\emph{etc}\onedot}

\makeatother

\title{Molmo2\\{\fontsize{18pt}{12pt}\selectfont  Open Weights and Data for Vision-Language Models with Video Understanding and Grounding}}

\newcommand{\core}{\textsuperscript{\textcolor{ai2pink}{\ding{170}}}}

\authorOne[1*]{Christopher Clark\core}
\authorOne[1,2*]{Jieyu Zhang\core}
\authorOne[1,2*]{Zixian Ma\core}
\authorOne[1,2*]{Jae Sung Park\core}
\authorOne[1,2]{Mohammadreza Salehi\core}
\authorOne[1]{Rohun Tripathi\core}
\authorOne[1]{Sangho Lee\core}

\authorTwo[1,2]{Zhongzheng Ren}
\authorTwo[1]{Chris Dongjoo Kim}
\authorTwo[2]{Yinuo Yang}
\authorTwo[2]{Vincent Shao}
\authorTwo[1]{Yue Yang}
\authorTwo[2]{Weikai Huang}
\authorTwo[1]{Ziqi Gao}
\authorTwo[1]{Taira Anderson}
\authorTwo[1]{Jianrui Zhang}
\authorTwo[1]{Jitesh Jain}
\authorTwo[1]{George Stoica}
\authorTwo[1]{Winson Han}
\authorThree[1,2]{Ali Farhadi}
\authorThree[1,2]{Ranjay Krishna\core}

\affiliation[1]{Allen Institute for AI}
\affiliation[2]{University of Washington}

\contribution[*]{denotes equal contribution. $\textcolor{ai2pink}{\varheartsuit}$ marks core contributors, who were all integral to the project\\ See full author contributions \hyperref[sect:contrib]{here}.}

\newcommand{\model}{Molmo2\xspace}

\definecolor{molmocolor}{RGB}{240, 82, 156}
\definecolor{tablegray}{RGB}{223, 242, 252}
\definecolor{tablegreen}{RGB}{15, 203, 150}
\definecolor{tableyellow}{RGB}{250, 242, 233}
\definecolor{tableblue}{RGB}{240, 82, 156}
\definecolor{darkpink}{RGB}{139, 14, 98}
\definecolor{baselinecolor}{gray}{.9}

\abstract{

Today’s strongest video-language models (VLMs) remain proprietary.
The strongest open-weight models either rely on synthetic data from proprietary VLMs, effectively distilling from them, or do not disclose their training data or recipe.
As a result, the open-source community lacks the foundations needed to improve on the state-of-the-art video (and image) language models.
Crucially, many downstream applications require more than just high-level video understanding; they require grounding—either by pointing or by tracking in pixels. Even proprietary models lack this capability.
We present Molmo2, a new family of VLMs that are state-of-the-art among open-source models and demonstrate exceptional new capabilities in point-driven grounding in single image, multi-image, and video tasks.
Our key contribution is a collection of 7 new video datasets and 2 multi-image datasets, including a dataset of highly detailed video captions for pre-training, a free-form video Q\&A dataset for fine-tuning, a new object tracking dataset with complex queries, and an innovative new video pointing dataset, all collected without the use of closed VLMs.
We also present a training recipe for this data utilizing an efficient packing and message-tree encoding scheme, and show bi-directional attention on vision tokens and a novel token-weight strategy improves performance. 
Our best-in-class 8B model outperforms others in the class of open weight and data models on short videos, counting, and captioning, and is competitive on long-videos. 
On video-grounding Molmo2 significantly outperforms existing open-weight models like Qwen3-VL (35.5 vs 29.6 accuracy on video counting) and surpasses proprietary models like Gemini 3 Pro on some tasks (38.4 vs 20.0 F1 on video pointing and 56.2 vs 41.1 $\mathcal{J}\&\mathcal{F}$ on video tracking).

}

\metadata[\quad\huggingface Models:]{
\href{https://huggingface.co/allenai/Molmo2-4B}{\texttt{Molmo2-4B}} \quad
\href{https://huggingface.co/allenai/Molmo2-8B}{\texttt{Molmo2-8B}} \quad
\href{https://huggingface.co/allenai/Molmo2-O-7B}{\texttt{Molmo2-O-7B}}}

\metadata[\vspace{.5em}\quad\hfdataset Data:]{
    \href{https://huggingface.co/collections/allenai/molmo2-data}{\texttt{Molmo2 Data}}
}

\metadata[\vspace{.5em}\quad\github Code:]{
    \href{https://github.com/allenai/molmo2}{\texttt{https://github.com/allenai/molmo2}} \quad
}

\metadata[\vspace{.5em}\quad\aitoo Demo:]{\href{https://playground.allenai.org/?model=molmo2-8b}{\texttt{playground.allenai.org}}}

\metadata[\vspace{.5em}\quad\emailLogo Contact:]{\color{ai2accent}{\texttt{molmo@allenai.org}}}

\begin{document}

\maketitle

\begin{figure*}[t]
  \centering
  \includegraphics[width=\textwidth]{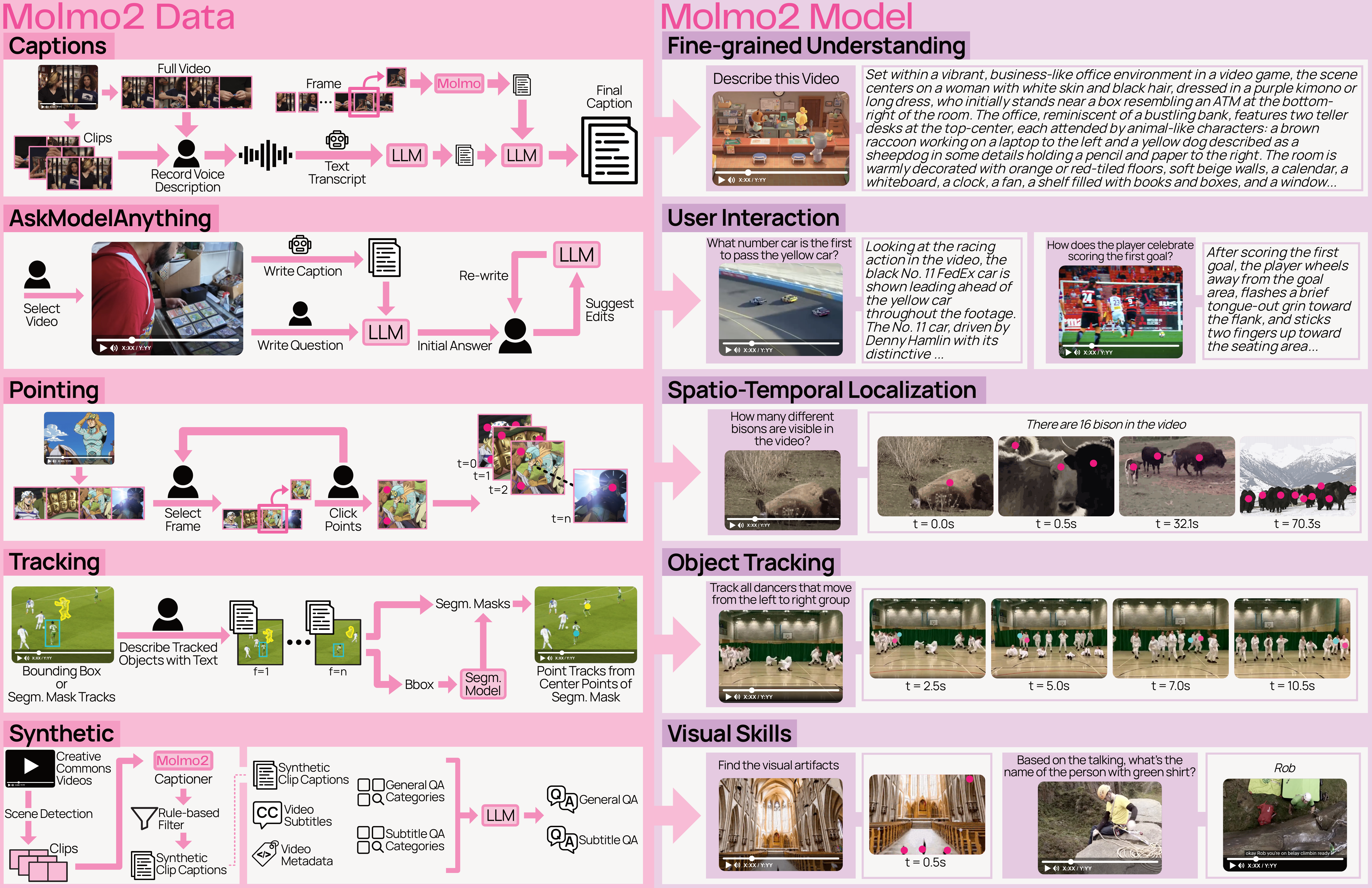}
  \caption{\model{} is trained on one of the largest fully open video-centric multimodal corpus to date, including nine new datasets for dense video captioning, long-form and long-video QA, and open-vocabulary pointing and tracking over images, multi-images, and videos. \model{} accepts single images, image sets, and videos as input and can produce both free-form language and grounded outputs such as spatio-temporal points, object tracks, and grounded chain-of-thoughts that localize objects and events over time. Across diverse video-language and grounding benchmarks, \model{} matches or surpasses prior open models, approaches proprietary systems, and remains fully open.}
  \label{fig:capabilities}
\end{figure*}

\section{Introduction}

Visual data (especially videos) is now ubiquitous, streaming continuously from phones, home cameras, social media, autonomous systems, and industrial sensors~\cite{do2025survey}. Understanding this video is fundamental for applications such as video search, household and industrial robotics, assistive technologies, sports analytics, security and traffic monitoring, and autonomous driving~\cite{liang2024vehicle,licardo2024intelligent,linlin2024cam}. 
Yet the strongest video–language models remain proprietary~\cite{team2024gemini,gpt4omini,eagle2_5,wang2024qwen2}, with closed weights, data, and training recipes.

A key missing capability in current video–language models is \textit{grounding}. Grounding would allow models to answer ``How many times does the robot grasp the red block?'', by emitting \textit{points} for each grasp event in space and time. It would identify ``When did the cup fall off the table?" by returning a \textit{track} of the cup so users can precisely locate the event.
Although image grounding is now standard~\cite{pointarena}, video grounding
is only supported in some proprietary systems, and even there in a limited form.

We present the \textbf{\model} (\textbf{M}ultimodal \textbf{O}pen \textbf{L}anguage \textbf{Mo}del), a family of \textit{fully open} state-of-the-art vision-language models.
\model supports single image, multi-image, as well as video, bridging the aforementioned gap by bringing grounding capabilities to video understanding.
To promote open research, \textit{we release our training data, model weights, and training code.}
To ensure our work is transparent and fully open, all our data is constructed without distilling from proprietary models.

A core contribution of this work is \textit{a suite of 9 novel datasets} targeting crucial skills underrepresented in existing open data for video and multi-image inputs.
This includes: (1) two open-vocabulary video pointing and tracking datasets (520k instances), enabling models to pinpoint when and where events or objects occur in videos; 
(2) a dense video captioning corpus (104k videos) with captions far longer and more detailed than in any prior work (e.g., GPT-generated video captions in LLaVA-Video~\cite{llava_video} and ShareGPT4Video~\cite{sharegpt4video}); 
(3) two long-form QA datasets (212k instances), including user questions on multi-image/video inputs with rich human-crafted answers (without distilling proprietary models); and 
(4) two long-video question answering datasets (around 1.3M instances) that tackle videos longer than those in current benchmarks (addressing a known weakness of open models on long-duration content~\cite{qian2024videostreaming}); (5) two multi-image datasets to improve multi-image pointing and document understanding. 

Our data collection uses multiple innovative pipelines (Figure~\ref{fig:capabilities}). For \textit{dense video captioning}, we devised a multi-stage process: human annotators first narrate each video clip in detail via spoken descriptions (allowing much more detail than text typing), which are transcribed and then enriched with frame-level visual details sourced from Molmo~\cite{molmov1} to ensure no detail is overlooked. 

Because existing large-scale datasets for video or multi-image are largely distilled from proprietary models~\cite{maaz2024videochatgpt, liu2024mmdu, jiang2024mantis, li2025smir, sharegpt4video}, we develop a human-and-LLM collaboration pipeline to create high-quality, long-form QA data from scratch. 
To add more data for medium (1-3 minutes) length videos, we introduce a synthetic data generator that uses our own captioning model to summarize and annotate extended videos (segmented into clips) and then formulates questions from those captions and the video’s transcript. 

\textit{Grounding capabilities} are vital. We extend the 2D pointing paradigm popularized in image-based VLMs~\cite{molmov1, jiang2025rexomni, yuan2024robopoint} into the temporal domain. Our models can not only point to objects in a frame, but also identify the moment an action happens or continuously track an object across a video. We created dedicated datasets for both video-pointing in space and time (\eg ``click the moment and location where X occurs”), and video-tracking (continuously indicating an object's position whenever it appears). 

Existing video grounding datasets tend to be narrow in scope or vocabulary, which is insufficient for training general models that can respond to arbitrary user input~\cite{ahmad2025videomolmo, munasinghe2025videoglamm}. We address this by generating large-scale video grounding data covering diverse actions and objects (including many high-frequency everyday objects and complex referring expressions), and we complement it with data converted from several academic sources (\eg reference video segmentation benchmarks) to ensure broad coverage. 
Finally, we construct a multi-image pointing dataset using PixMo-Points~\cite{molmov1}, enabling our model to output points on multiple images.


All \model variants are trained in a three-stage pipeline: (1) an image-captioning and image-pointing pre-training stage, (2) a joint supervised fine-tuning stage on our integrated multimodal dataset mixture (images, videos, and multi-image inputs), and (3) a short long-context training stage on the same data. We introduce several training innovations that further boost performance: a \textit{novel token-weighting scheme} during fine-tuning to balance learning from diverse tasks, as well as efficient training techniques like \textit{sequence packing} and a \textit{message-tree schedule} that dramatically increase training throughput. We also show that enabling \textit{bi-directional attention} between visual tokens yields notable gains. 

We evaluate \model across a broad spectrum of established benchmarks, and also propose new evaluation sets for the less-explored capabilities we target (such as dense video captioning and open-vocabulary video pointing). On short-video understanding, \model achieves results on par with or better than existing models; for example, it outperforms previous open models on benchmarks like MVBench~\cite{mvbench} and MotionBench~\cite{motionbench}, and even challenges some proprietary models’ performance on these tasks.
In tasks like visual counting and captioning, \model (even at 4B scale) is only outperformed by the strongest closed-source systems (\eg Gemini 3.0~\cite{gemini3}), demonstrating the benefits of our fine-grained grounding data. \model also establishes new state-of-the-art results in video grounding (both tracking and pointing), substantially ahead of prior open models~\cite{ahmad2025videomolmo, munasinghe2025videoglamm}, all while maintaining strong performance on traditional image and multi-image benchmarks~\cite{jiang2024mantis, liu2024mmdu}. A human preference evaluation ranks \model as equal or better than existing open-weight models and ahead of a few proprietary models, including GPT-5~\cite{gpt5} and Claude Sonnet 4.5~\cite{anthropic2025sonnet}, showing its general-purpose capabilities.

We release three versions of \model{}: 4B and 8B models based on the Qwen3 LLMs~\cite{qwen3technicalreport}, and a 7B model based on the OLMo LLM~\cite{olmo3}, to demonstrate what can be achieved with a fully-open language model. All our code, data, and models will be made open source.


\begin{table*}[!t]
    \centering
    \tablestyle{2pt}{1.3}
    \begin{tabular}{l p{9cm} c c c}
        \textbf{Dataset Group} & \textbf{Description} & \textbf{Rate}(\%) & \textbf{Datasets} & \textbf{Examples} \\
        \toprule
         Captions/Long QA & Captioning and long-form question answering data on images and videos, including 
         \textcolor{molmocolor}{\model{}-Cap, }
         \textcolor{molmocolor}{\mbox{-AskModelAnything}, \mbox{-MultiImageQA}} and \mbox{PixMo-Cap}, \mbox{-AskModelAnything} and \mbox{-CapQA}. & 13.6 & 6 & 1.2m \\
         Image QA & Multiple-choice and short answer image QA data, including \textcolor{molmocolor}{\model-SynMultiImageQA}, open-source image datasets~\cite{vqa2,textqa,okvqa,chartqa,docqa,infoqa,ai2_diagram,a_okvqa,science_qa,tab_wmp,st_qa,tally_qa,dv_qa,figure_qa,plot_qa} following Molmo with CoSyn~\cite{cosyn} instead of PixMo-Docs, and open-source multi-image datasets~\cite{suhr2018nlvr2,liu2023llava,jhamtani2018std}. & 22.7 & 32 & 2.4m\\
         Video QA & Multiple-choice and short answer video QA, including \textcolor{molmocolor}{\model{}-CapQA, -SubtitleQA}, and various open video datasets~\cite{tgif,tvqa,paxion,llava_video,perception_test,nextqa,news_video_qa,how2qa,sutd,clevrer,social_iq2,star,road_text_vqa,countix,camerabench_qa,motionbench,ssv2,moments_in_time_qa,kinetics_qa,charades_sta,coin,youcook2,activitynet,ego4d,epic_kitchens,video_localized_narratives_caption,qv_highlights,intentqa,funqa,cinepile,sportqa}. Downsampled since video-benchmarks converge quickly. & 18.2 & 32 & 2.4m\\
         Image Pointing & PixMo-Points and PixMo-Count, CoSyn-Point~\cite{cosyn}, and \textcolor{molmocolor}{\model{}-MultiImagePoint}. PixMo-Points is weighted to emphasize high counts. Downsampled since it was seen during pre-training. & 9.1 & 4 & 1.1m\\
         Video Pointing & \textcolor{molmocolor}{\model-VideoPoint} and AcademicVideoPoint. Upsampled since this task is slow to converge. & 13.6 & 7 & 0.37m\\
         Video Tracking & \textcolor{molmocolor}{\model-VideoTrack} and AcademicVideoTrack. Re-weighted to emphasize tail concepts. & 13.6 & 22 & 0.80m \\
         NLP & Text-only SFT data from Tulu~\cite{tulu3} to preserve performance on natural language understanding. & 9.1 & 1 & 0.99m\\
    \end{tabular}
    \caption{We create nine new datasets (in \textcolor{molmocolor}{pink}) to train \model{}. We also include a suite of image and language data from academic datasets into our training mix. We categorize all datasets into categories and show each categories' sampling rate, dataset count, and total training examples after filtering and formatting the data into message trees. See Section~\ref{sect:data} and the appendix for details.}
    \label{tab:data_top_level_categories}
\end{table*}

\section{Data}
\label{sect:data}

We create five human-annotated datasets and four synthetic datasets, and additionally curate two datasets by repurposing existing open-source data. We summarize their design and collection pipelines below; see the appendix for details.

\paragraph{\model-Cap (human).} \label{data:molmo2_cap}
We collect 104k video-level and 431k clip-level dense captions from annotators, targeting both high detail and broad diversity. Videos are drawn from multiple large-scale sources~\cite{zellers2022merlotreserve,wang2024koala36mlargescalevideodataset,wang2023internvid,llava_video}, starting from a pool of over 10M clips, then filtered for informativeness and sampled for diversity to obtain a balanced subset.

Obtaining dense video captions is challenging because annotators must describe dynamic events alongside fine-grained visual details~\cite{krishna2017dense}. We use a two-stage pipeline: annotators first describe short clips, then summarize the entire video. As in PixMo-Cap~\cite{molmov1}, annotators speak their descriptions, which are transcribed with Whisper-1~\cite{radford2023robust} and then rewritten by a text-only LLM for coherence.
We condition annotators to describe dynamic visual details (\eg object or event changes over time) by prompting them with a set of predefined questions.
To add any missing low-level details, we use Molmo to generate frame-level captions and an LLM to merge the clip and frame captions into a single long caption. 
This produces the densest video caption dataset to date, averaging 924 words per video, compared to 75 words in Video Localized Narratives~\cite{video_localized_narratives_caption}, 89 and 100 in RCap and RDCap~\cite{cho2025PerceptionLM}, 280 in ShareGPT4-Video~\cite{sharegpt4video}, and 547 in LLaVA-Video-178K~\cite{llava_video}.

\paragraph{\model-AskModelAnything (human).} \label{data:molmo2_ama}
We collect 140k human-authored video QA pairs. Using video captions, we cluster videos into 31 categories and sample them evenly to promote data diversity. Annotators then write specific, fine-grained questions (\eg about text, actions, or temporal relations), while we discourage counting questions (handled separately by pointing data), overly generic prompts, or questions requiring expert knowledge.
For each question, we first obtain an initial answer from an LLM (Claude Sonnet 4.5) conditioned on a caption generated by an early \model{} captioner. Annotators either accept the answer or iteratively refine it through dialogue with the LLM. Finally, we post-process all QA pairs with an LLM filter to remove non-English, mismatched, or counting questions. We remove counting questions since the model should point for those questions instead of producing a pure text response.

\paragraph{\model-CapQA and -SubtitleQA (synthetic).}
To build large-scale synthetic video QA, we use a video captioner trained on \model-Cap to caption videos from YT-Temporal~\cite{zellers2022merlotreserve} and YouTube keyword search. We segment each video into multiple scenes and caption each scene instead of the entire video to encourage detailed descriptions. An LLM then uses these captions and video metadata to generate 1M QA pairs (200k videos, 5 QA per video). For SubtitleQA, we transcribe the video audio with Whisper-1 and additionally prompt the LLM with the transcript to create 300k QA pairs (100k videos, 3 QA per video) that require reasoning over both visual content and language.



\paragraph{\model-VideoPoint (human).} To improve \model's counting and spatial-temporal localization, we collect over 650k video pointing queries on 280k videos, with an average of 6 points per video, targeting eight diverse categories: objects, animals, actions/events, referring expressions, indirect references, spatial references, comparative references, and visual artifacts/anomalies (for generative videos only). We generate queries by using LLM on video captions from an early version of \model{}. Annotators first identify the frame where an object appears and then click on its exact location in the frame. Frames were obtained at 2 fps. 

\paragraph{\model-VideoTrack (human).}
We collect point-based object-tracking data covering 3.6k video clips and 15k complex natural language queries, with an average of 2.28 objects per query. Our dataset collection follows Ref-VOS~\cite{refvos} by asking users to re-label existing tracking annotations. For each video, we display either segmentation or bounding box object tracks, and ask annotators to craft non-trivial text queries that apply to a subset of objects. The queries are then validated in a separate validation round. We source videos and tracks from diverse open-source segmentation tracks~\cite{refvos,mosev2,vpseg,ravi2024sam2} and bounding-box tracks{~\cite{sun2022dancetrack, zhang2023animaltrack, scott2024teamtrack, wang2022sportstrack, giancola2018soccernet, dendorfer2020mot20, zheng2024nettrack, du2018unmanned, varga2022seadronessee, yu2020bdd100k}}. 

\noindent\textbf{AcademicVideoPoint and AcademicVideoTrack (curated).}
For pointing, we convert existing object tracking annotations from six datasets~\cite{athar2023burst,lvvis,qi2022ovis, refvos,refdavis,ding2023mevis} into 49k pointing and counting QAs. We first obtain the timestamp of the first frame in which an object appears and then randomly sample a point in the object’s mask with a Gaussian distribution around the mask center. For tracking, we repurpose 7 existing Ref-VOS datasets~\cite{refdavis, seo2020urvos, ding2023mevis, athar2023burst, lvvis, yan2024visa, athar2025vicas} to obtain point tracking supervision data. In addition, we process 11 bounding-box based tracking datasets~\cite{zhang2024webuot, huang2019got, peng2024vasttrack, muller2018trackingnet, hong2023lvos, fan2019lasot, 10004511, lamdouar2020betrayed, wang2021tnl2k, Wang2025ReasoningTrackCR, zhu2023tiny} by using SAM-2 to generate segmentation masks and corresponding point tasks. 


\paragraph{\model-MultiImageQA (human).}
We collect QA data on semantically related image sets to support real-world multi-image queries. We form image sets by grouping images whose captions (generated by a PixMo-Cap–trained model) have high sentence-level similarity; each set contains 2–5 images (2.73 on average). Human annotators then write questions over each set, and answers are refined through the same human–LLM loop as above. In total, we construct 45k image sets from 96k unique images and 72k QA pairs.

\paragraph{\model-MultiImagePoint and -SynMultiImageQA (synthetic).}
To improve multi-image grounding, we construct a dataset of over 470k pointing and counting examples by applying soft clustering over images in PixMo-Points. Image sets are formed using a combination of single-token and sentence-level label embedding similarities, producing sets of 2–5 semantically related images (mean set size: 3.24).
For each image set, we first normalize all human-provided labels via lowercasing, punctuation, and whitespace normalization, and synonym consolidation. We then use a large language model to resolve these normalized labels into a single canonical description that is semantically consistent across the set. This canonical label defines the shared entity or concept to be pointed to and counted across all images in the set.
During training, we stochastically sample from the original (pre-canonicalized) human annotations rather than always using the canonical label, thereby preserving lexical diversity and improving robustness to annotation variability.

For \model-SynMultiImageQA, we adapt CoSyn~\cite{cosyn} to create 188k synthetic multi-image examples with text-rich images such as charts, tables, and documents.

\section{Training}

\label{sec:training}
This section provides an overview of our model and training pipeline. See the appendix for additional details.

\subsection{Architecture}
\begin{figure}[!t]
  \centering
  \begin{minipage}{0.58\linewidth}
    \includegraphics[width=\linewidth]{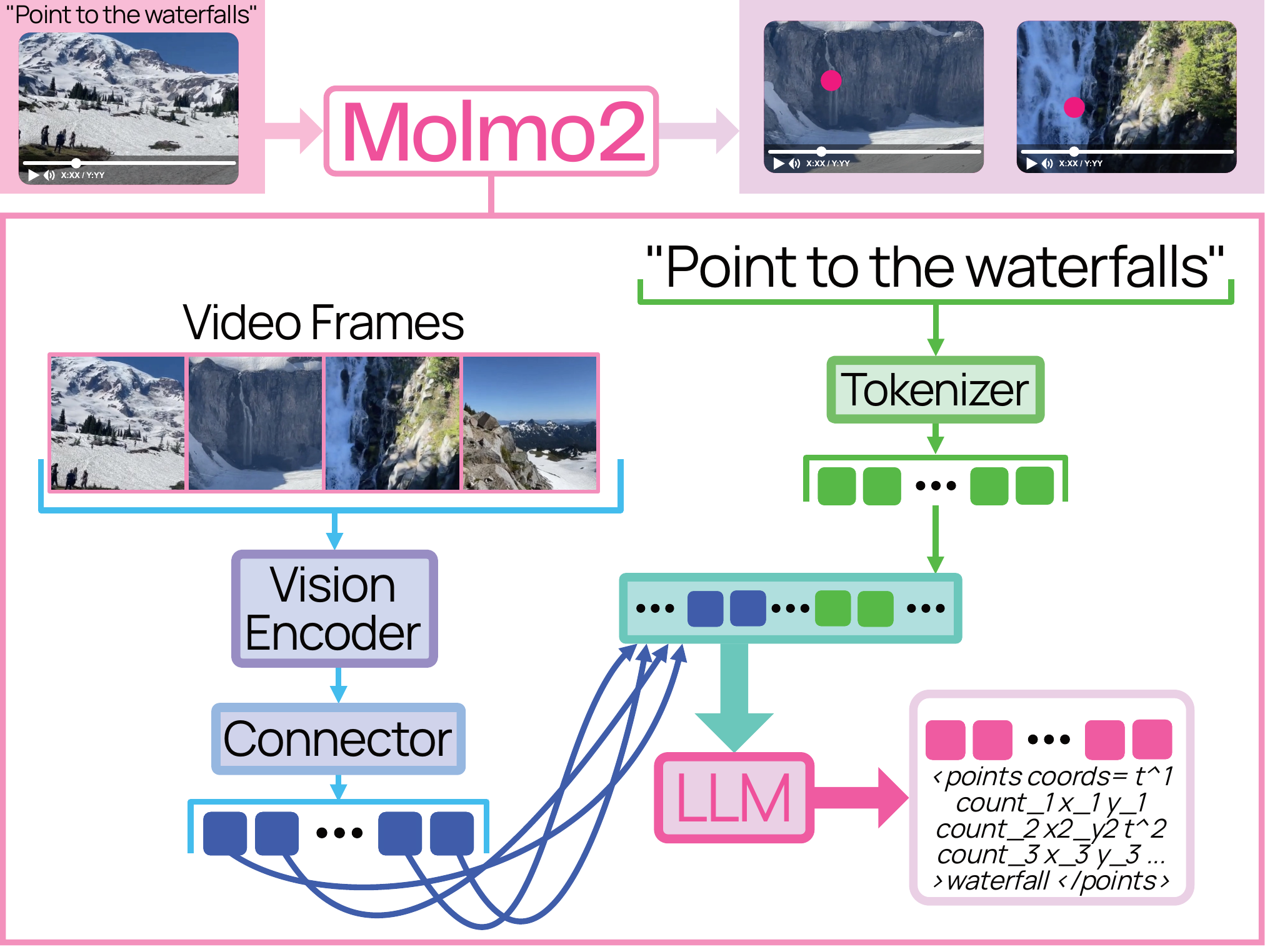}
    \caption{\model follows the standard design of connecting a vision encoder and a language model to process video inputs.}
    \label{fig:model}
  \end{minipage}\hfill
  \begin{minipage}{0.38\linewidth}
  \centering
    \includegraphics[width=0.9\linewidth]{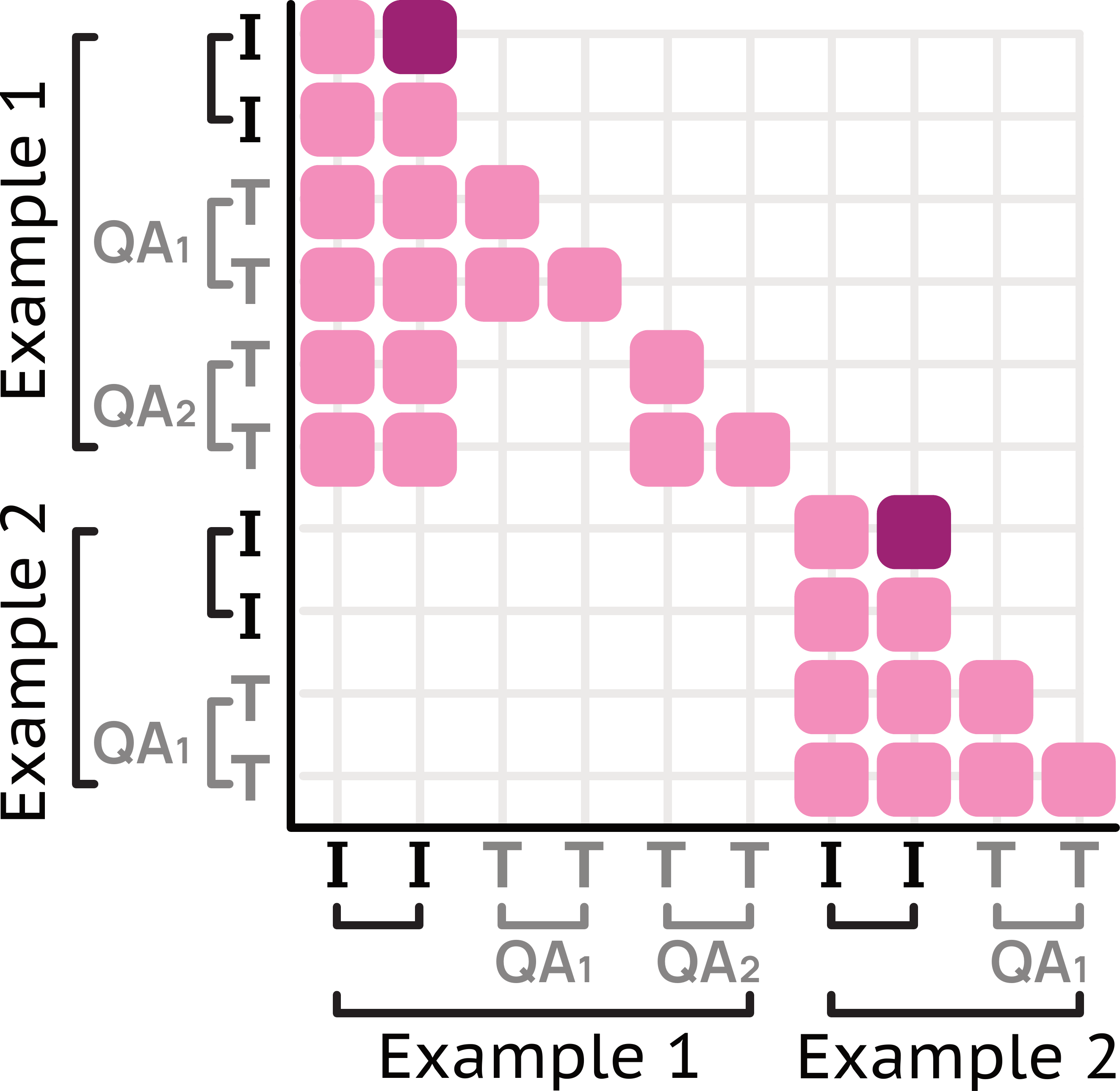}
    \caption{Attention mask for a \textit{packed} sequence with two examples. The first contains two QA pairs for one image. Frame tokens (\textcolor{darkpink}{dark pink}) have forward attention, while masking blocks cross-attention between different examples (lower-left empty block) and between distinct QA pairs within the same example (upper empty block).}
    \label{fig:attention_mask}
  \end{minipage}
\end{figure}

Our model architecture follows the common design of combining a pre-trained LLM and a vision transformer (ViT)~\cite{dosovitskiy2021vit} via a connector module~\cite{molmov1, liu2023llava}. Visual inputs are split or resized into fixed-size crops, which are encoded into patch-level features by the ViT. The patch-level features are then pooled, projected by the connector, and passed as visual tokens, along with any text inputs, to the LLM. Figure~\ref{fig:model} provides an overview.

\noindent\textbf{Cropping.}
For input images, we use a single crop of the down-scaled image as well as up to $K$ overlapping crops tiling the image to allow higher-resolution processing~\cite{molmov1}. Images that cannot be tiled by $K$ crops are downscaled. We use $K=8$ during training and $K=24$ during inference.
For videos, we sample frames at $S=2$ fps as single crops (downscaling if needed) to reduce computational costs when processing long videos. 
We set a maximum of $F=128$ frames (or $F=384$ for long-context training). If the video length is longer than $F / S$, we uniformly sample $F$ frames. In both cases, the last frame is always included since most video players will display the last frame after the video finishes playing, and it therefore might have special importance to users.

\noindent\textbf{Vision-language connector.}
The connector uses features from the third-to-last and ninth-from-last ViT layers, following~\cite{molmov1}. For images, 2$\times$2 patch windows are pooled into a single vector using a multi-headed attention layer, where the mean of the patches serves as the query. For video frames, a 3$\times$3 patch window is used instead to reduce the token count. We use the same shared parameters for the connector for both image and video frame pooling. Finally, the pooled features are projected using a shared MLP.

\noindent\textbf{LLM.}
The LLM takes as input the visual tokens interleaved with text timestamps (for videos) or image indices (for multi-image input). For multi-crop images, we include column tokens~\cite{molmov1} to indicate the image's aspect ratio. We do not include column-tokens for single-crop images since they are always square. We also add image and frame start tokens and include subtitles (marked with text timestamps) as text after the visual input if available. We allow image tokens (even if they are from different frames/images) to forward-attend to one another~\cite{gao2025tc,team2025gemma}, which we find can increase performance.

\subsection{Training}
We use a simple three-stage design: a light-weight image-only pre-training stage, a joint video/image supervised fine-tuning (SFT) stage, and then a short long-context SFT stage.
We train on the \model{} data, image data from PixMo, and various open-source datasets. We review those stages and additional training details here, but leave most details to the appendix.

\noindent\textbf{Pre-training.} Our pre-training stage includes dense captioning with length conditioning and transcript prediction using PixMo-Cap, following~\cite{molmov1}. We add NLP data using the supervised fine-tuning data from Tulu~\cite{tulu3}, filtered to remove non-English content and code, to better preserve language capabilities. Additionally, we add pointing data from PixMo-Points, PixMo-Count, and CoSyn-Point~\cite{cosyn}. We find that adding pointing data during pre-training leads to better and more stable pointing performance. We use 60\% captioning, 30\% image pointing, and 10\% natural language for the mixing ratios. We train for 32k steps with a batch size of 128, which results in about 4 epochs of training on PixMo-Cap. All parameters are fine-tuned, and we use separate learning rates for the ViT, connector, and LLM following~\cite{molmov1}.

\noindent\textbf{SFT.} 
Our data mixture combines PixMo~\cite{molmov1}, the \model{} datasets, Tulu, and other open-source video and image datasets. We divide these datasets into categories and manually assign each category a sampling rate based on empirical tests; see Table~\ref{tab:data_top_level_categories}.
Within each category, we sample datasets proportionally to the square root of each dataset size, with the addition of some manual rebalancing, such as downsampling large synthetic datasets. We train for 30k steps with a batch size of 128 and a max sequence length of 16,384.

\paragraph{Long-context SFT.}
Finally we do a third stage of training with a longer context length~\cite{eagle2_5,team2024gemini} on the same SFT data mixture. 
During this stage we increase the sequence length to 36,864, set $F=384$, train for 2k steps, and use context parallelism (CP) on the LLM so each example is processed by a group of 8 GPUs. We employ Ulysses attention~\cite{jacobs2024system} for the LLM context parallelism as its all-gather offers flexibility with the custom attention masks used by our packing and message tree system \cite{llama3}. We also distribute video frame processing by the vision encoder and the attentional pooling after that across each context parallel group and find it very effective in reducing the memory footprint of the model. We only do long-context training as a short final training stage since its adds significant overhead to the training.

\noindent\textbf{Pointing and tracking.}
We represent point coordinates with a compressed plain-text format that includes normalized x and y coordinates, a timestamp (for video) or an image index (for images), and an integer ID that is unique for each distinct object to enable tracking and counting.
Points are sorted based on time/image index, then x, y coordinates. 
During SFT, we use a maximum of 24 crops instead of 8 for 30\% of images with pointing annotations to ensure that pointing can generalize to high-resolution images.
For video pointing, we train with examples with up to 60 points annotated. Additionally, we construct and train on multi-turn conversations with multiple pointing or counting queries for the same videos. 
For tracking, we also add auxiliary tasks of predicting only the first and last frames in which the objects appear, or tracking from an input query and point.

\paragraph{Token weighting.}
Our data includes both multiple choice questions with a single output token and long video captions with 4,000+ output tokens. These long-output examples can easily become the large majority of loss tokens even if they are sampled rarely, which can cause degradation on short-answer or multiple-choice tasks. As a solution, we adjust the weighting of some examples when they are used with the loss. We use a fixed weight of 0.1 for video captions and 0.2 for pointing, since both of these tasks can have very long, dense outputs. For other tasks we follow the heuristic of $\frac{4}{\sqrt{n}}$ where $n$ is the number of answer tokens, which better balances long and short output training examples. 


\paragraph{Packing.}
Examples can have anywhere from hundreds (pure-text or small images) to 16k+ (videos with subtitles or long videos during long-context training) of tokens. To avoid wasteful padding when creating training batches, we use packing to merge multiple short examples into a single long sequence. Packing is non-trivial for vision-language models due to the need to efficiently pack both crops for the ViT and tokens for the LLM, and the need to support models with different approaches to converting images/videos into tokens. We develop an on-the-fly packing algorithm that builds maximally efficient packed sequences from a small pool of in-memory examples and can be integrated into standard PyTorch data loaders.

\paragraph{Message trees.}
We encode videos and images with multiple annotations as \textit{message-trees}. The visual input is encoded as the first message, and each annotation becomes a different branch. The tree is linearized as a single sequence with a custom attention mask to prevent branches from cross-attending to each other. On average, examples in our data have 4 annotations, and packing is able to fit 3.8 examples into a 16348 token sequence during SFT, leading to 15x training efficiency.
Figure~\ref{fig:attention_mask} shows the attention masking.

\section{Evaluation}
We evaluate \model on standard video academic benchmarks and on our new benchmarks for video captioning, counting, and pointing, as well as a large-scale human-preference study. 
Then we report results for ablations, task-specific \model{} variants, and test-time scaling.
See the appendix for details, additional ablations, evaluations on NLP benchmarks, and additional discussion.

\subsection{Overall results}
\newcommand{\mycell}[2]{%
  \rotatebox{90}{%
    \parbox{2.0cm}{%
      \setlength{\baselineskip}{0.5em}%
      \textbf{\scriptsize{#1}}\\
      \footnotesize{\textcolor{gray}{#2}}%
    }%
  }%
}

\newcommand{\newcell}[1]{%
  \rotatebox{90}{%
    \parbox{2.0cm}{%
      \setlength{\baselineskip}{0.5em}%
      \textbf{\scriptsize{#1}}
    }%
  }%
}

\begin{table*}[!ht]
    \renewcommand{\arraystretch}{0.98}
    \centering
    \setlength{\tabcolsep}{2.5pt}
    \resizebox{\textwidth}{!}{
    \begin{tabular}{@{}l>{\columncolor{tableyellow!50}}c
    >{\columncolor{tableyellow!50}}c
    >{\columncolor{tableyellow!50}}c
    >{\columncolor{tableyellow!50}}c
    >{\columncolor{tableyellow!50}}c
    >{\columncolor{tableyellow!50}}c
    >{\columncolor{tablegreen!10}}c
    >{\columncolor{tablegreen!10}}c
    >{\columncolor{tablegreen!10}}c
    >{\columncolor{tablegreen!10}}c
    >{\columncolor{tablegreen!10}}c
    >{\columncolor{tablegreen!10}}c
    >{\columncolor{tablegreen!10}}ccc
    >{\columncolor{tableyellow!50}}c
    >{\columncolor{tablegreen!10}}c
    >{\columncolor{tableblue!10}}crr@{}}
        \textbf{Model} & 
        \mycell{NextQA}{test~\cite{nextqa}}
        & \mycell{PerceptionTest}{test~\cite{perception_test}}
        & \mycell{MVBench}{test~\cite{mvbench}}
        & \mycell{Tomato}{test~\cite{tomato}}
        & \mycell{MotionBench}{val~\cite{motionbench}}
        & \mycell{TempCompass}{test MCQ~\cite{tempcompass}}
        & \mycell{Video-MME}{test~\cite{videomme}}
        & \mycell{Video-MME-Sub}{test~\cite{videomme}}
        & \mycell{LongVideoBench}{val~\cite{longvideobench}}
        & \mycell{MLVU}{test MCQ~\cite{mlvu}}
        & \mycell{LVBench}{test~\cite{lvbench}}
        & \mycell{VideoEvalPro}{test~\cite{videoevalpro}}
        & \mycell{Ego Schema}{test~\cite{egoschema}}
        & \mycell{Molmo2 Caption}{test F1 Score}
        & \mycell{Molmo2 Count}{val accuracy}
        & \newcell{Short QA avg.}
        & \newcell{Long QA avg.}
        & \newcell{Average}
        & \newcell{Elo Score}
        & \newcell{Elo Rank}\\
        \midrule
        
        \multicolumn{16}{@{}l}{\textbf{\textit{API call only}}} \\

        GPT-5~\cite{gpt5}              & 86.3 & 79.4 & 74.1 & 53.0 & 65.4 & 80.4 & 83.3 & 86.9 & 72.6 & 77.7 & 65.2 & 68.8 & 75.6 & 50.1 & 35.8 & 73.1 & 76.3 & 70.6 & 1031 & 10\\
        GPT-5 mini~\cite{gpt5}         & 83.2 & 72.0 & 66.5 & 44.1 & 59.9 & 74.9 & 77.3 & 82.3 & 69.7 & 69.1 & 54.7 & 60.1 & 70.9 & 56.6 & 29.8 & 66.8 & 69.8 & 65.0 & 1076 & 4\\
        Gemini 3 Pro~\cite{gemini3} & 84.3 & 77.6 & 70.4 & 48.3 & 62.6 & 82.8 & 88.6 & 87.5 & 75.9 & 75.7 & 77.0 & 78.0 & 68.9 & 36.0 &37.1 & 71.0	& 78.8 & 70.0 & 1082 & 3\\
        Gemini 2.5 Pro~\cite{comanici2025gemini}     & 85.3 & 78.4 & 70.6 & 48.6 & 62.0 & 81.9 & 87.8 & 87.8 & 76.8 & 81.5 & 75.7 & 78.4 & 72.2 & 42.1 & 35.8 & 71.1 & 80.4 & 71.2 & 1096 & 1\\
        Gemini 2.5 Flash~\cite{comanici2025gemini}   & 81.8 & 74.7 & 67.0 & 39.1 & 59.3 & 80.2 & 84.2 & 84.2 & 73.1 & 75.1 & 64.9 & 69.6 & 70.2 & 46.0 & 31.9 & 67.0 & 74.5 & 66.7 & 1084 & 2\\
        Claude Sonnet 4.5~\cite{anthropic2025sonnet}  & 79.2 & 64.3 & 62.1 & 39.6 & 58.5 & 72.8 & 74.2 & 80.5 & 65.1 & 64.0 & 50.5 & 50.5 & 73.1 & 26.0 & 27.2 & 62.8 & 66.4 & 59.6  & 1008 & 12 \\
        \midrule
        
        \multicolumn{16}{@{}l}{\textbf{\textit{Open weights only}}} \\

        InternVL3.5-4B~\cite{wang2025internvl3}     & 80.3 & 68.1 & 71.2 & 26.8 & 56.5 & 68.8 & 65.4 & 68.6 & 60.8 & 52.0 & 43.2 & 46.5 & 58.9 & 7.7 & 26.3  & 62.0 & 56.5 & 53.4 & 935 & 18\\

        InternVL3.5-8B~\cite{wang2025internvl3}     & 81.7 & 72.7 & 72.1 & 24.6 & 56.6 & 70.3 & 66.0 & 68.6 & 62.1 & 53.2 & 43.4 & 48.1 & 58.6 & 7.8 & 26.1  & 63.0 & 57.1 &  54.1 & 941 & 19\\

        Qwen3-VL-4B~\cite{qwen3technicalreport}         & 81.4 & 70.7 & 68.9 & 31.8 & 58.6 & 70.8 & 69.3 & 74.0 & 62.8 & 58.4 & \underline{56.2} & 49.8 & 68.4 & 25.2 & 25.3 & 63.7 & 62.7 & 58.1 & {1048} & {7}\\

        Qwen3-VL-8B~\cite{qwen3technicalreport}         & 83.4 & 72.7 & 68.7 & 35.7 & 56.9 & 74.3 & 71.4 & 75.2 & 62.4 & 57.6 & \textbf{58.0} & 50.3 & \underline{69.8} & 26.7 & 29.6 & 65.3 & {63.5} & 59.5 & \underline{1054} & \underline{6}\\

        Keye-VL-1.5-8B~\cite{yang2025kwai}        & 75.8 & 64.2 & 56.9 & 33.0 & 55.1 & \textbf{75.5} & \textbf{73.0} & \textbf{76.2} & 66.0 & 53.8 & 42.8 & 54.9 & 56.3 & 25.4 & 27.2 & 60.1 & 60.4 & 55.7 & 952 & 17\\

        GLM-4.1V-9B~\cite{glmv}          & 81.3 & 74.2 & 68.4 & 30.0 & 59.0 & 72.3 & 68.2 & 75.6 & 65.7 & 56.6 & 44.0 & 51.1 & 62.6 & 18.4 & 26.6 & 64.2 & 60.5 & 56.9 & 962 & 14\\

        MiniCPM-V-4.5-8B~\cite{yu2025minicpmv45cookingefficient}      & 78.8 & 70.9 & 60.5 & 29.8 & 59.7 & 72.7 & 67.9 & 73.5 & 63.9 & \underline{60.6} & 50.4 & 54.9 & 49.6 & 29.3 & 26.3 & 62.1 & 60.1 & 56.6 & 975 & 13\\

        Eagle2.5-8B~\cite{eagle2_5}             & 85.0 & 81.0 & 74.8 & 31.0 & 55.7 & \underline{74.4} & \underline{72.4} & 75.7 & 66.4 & 60.4 & 50.9 & 58.6 & \textbf{72.2} & 22.8 & 28.9 & 67.0 & \textbf{65.2} & 60.7 & 1019 & 11 \\

        \midrule
        
        \multicolumn{16}{@{}l}{\textbf{\textit{Open models}}}  \\
        PLM-3B~\cite{cho2025PerceptionLM}             &83.4 & 79.3 & 74.7 & 30.9 & 60.4 & 69.3 & 54.9 & 59.4 & 57.9 & 48.4 & 40.4 & 46.2 & 66.9 & 12.3 & 24.4 & 66.3 & 53.5 & 53.9 & 841 & 20\\

        PLM-8B~\cite{cho2025PerceptionLM}            & 84.1 & \textbf{82.7} & \textbf{77.1} & 33.2 & 61.4 & 72.7 & 58.3 & 65.4 & 56.9 & 52.6 & 44.5 & 47.2 & 68.8 & 10.9 & 26.6 & 68.5 & 56.2 & 56.2 & 853 & 21\\
        
        LLaVA-Video-7B~\cite{llava_video}     & 83.2 & 68.8 & 58.6 & 24.9 & 54.2 & 66.6 & 63.3 & 69.7 & 58.2 & 52.8 & 44.2 & 47.8 & 57.3 & 19.9 & 21.4 & 59.4 & 56.2& 52.7 & 959 & 15\\

        VideoChat-Flash-7B~\cite{li2024videochat}    & \underline{85.5} & 76.5 & 74.0 & 32.5 & 60.6 & 69.4 & 65.3 & 69.7 & 64.7 & 56.0 & 48.2 & 51.2 & 51.3 & 14.8 & 21.6 & 66.4 & 58.1 & 56.1 & 956 & 16\\

        \midrule
        
        \multicolumn{16}{@{}l}{\textbf{\textit{Molmo2 family: Open weights, Open data (no distillation), Open code}}} \\
        \textcolor{molmocolor}{Molmo2-4B}     & \underline{85.5} & 81.3 & 75.1 & \textbf{39.8} & \underline{61.6} & 72.8 & 69.6 & 75.7 & \textbf{68.0} & \textbf{63.0} & 53.9 & \underline{59.9} & 61.2 & 39.9 & \underline{34.3} & \underline{69.3} & \underline{64.5} & \underline{62.8} & 1041 & 8\\

        \textcolor{molmocolor}{Molmo2-8B}      & \textbf{86.2} & \underline{82.1} & \underline{75.9} & \underline{39.6} & \textbf{62.2} & 73.4 & 69.9 & \underline{75.8} & \underline{67.5} & 60.2 & 52.8 & \textbf{60.4} & 62.0 & \textbf{43.2} & \textbf{35.5} & \textbf{69.9}	&64.1 & \textbf{63.1} & \textbf{1057} & \textbf{5}\\

        \textcolor{molmocolor}{Molmo2-O-7B}          & 84.3 & 79.6 & 74.8 & 36.2 & 60.6 & 73.0 & 64.9 & 69.2 &	63.7 & 55.2 & 49.6 & 55.1 & 56.8 & \underline{40.1} & 33.2 & 68.1	& 59.2 & 59.7 & 1033 & 9\\
    \end{tabular}
    }%
    \caption{\textbf{Video benchmark results} for a range of proprietary APIs, open-weight baselines, video-specialized models, and our Molmo2 family across video understanding, captioning, and counting benchmarks. The result of the best-performing open-weight model is in \textbf{bold}, and the second best is \underline{underlined}.}
    \label{tab:video_benchmark_results}
\end{table*}

We evaluate captioning by constructing \model{}-CapTest, an eval set of 693 Creative Commons-licensed videos with at least four human-annotated captions. We use an LLM-as-a-judge to compute precision, recall, and F1 for statements made in the model's caption relative to statements from the annotator's captions, similar to Molmo's image captioning metric~\cite{molmov1}. For counting, we construct \model-VideoCount by using our \model-VideoPoint pipeline to collect 533 diverse examples that cover object, action, and animal queries with up to 60 points.

For the human preference study, we collect questions from human annotators and manually filter them to prioritize open-ended questions over straightforward ones, resulting in 450 questions. We added another 51 videos for captioning queries. We sample two model outputs and gather pairwise preferences on them from annotators. We collect over 105K ratings (501 per model pair). From this data, we calculate an Elo ranking using the Bradley-Terry model~\cite{chiang2024chatbot}.

We obtain results for all models on all tasks. We prioritize author-published results but fill in missing results with the best previously reported values from technical reports or papers. If data is still missing, we compute it ourselves. We try to follow the author's eval setup, but note that eval details (\eg, prompting or number of frames) are sometimes not public, so results should be interpreted carefully.

During inference, we use $384$ frames and greedy decoding. For human evaluations and video captioning, we use top\_p=0.95, temperature=0.7, and frequency\_penalty=0.1 instead, which produces more natural results when generating long outputs.

Results are in Table~\ref{tab:video_benchmark_results}; we highlight a few key takeaways:
\begin{itemize}[leftmargin=5mm,topsep=1mm]
\item \model{} is SoTA on short video benchmarks, captioning, and counting among non-proprietary models
\item \model{} outperforms previous fully-open models but lags behind the best open-weight models. We believe this is due to a lack of open-source long (10+ minutes) training data and computational limitations that made it challenging to run extensive ultra-long context training.
\item \model{} ranks equal to or better than other open-weight models on human preference, and is far ahead of previous fully-open models.
\end{itemize}

\subsection{Grounding results}
\label{sect:grounding_results}
\newcommand{\mycellg}[2]{%
  \rotatebox{90}{%
    \parbox{2.8cm}{%
      \setlength{\baselineskip}{0.5em}%
      \textbf{\scriptsize{#1}}\\
      \footnotesize{\textcolor{gray}{#2}}%
    }%
  }%
}

\begin{table}[!ht]
    \renewcommand{\arraystretch}{0.98}
    \centering
    \small
    \begin{tabular}{@{}lccccccc@{}}
     & 
        \multicolumn{2}{@{}c}{\textbf{BURST~\cite{athar2023burst} VC (test)}} 
        & \multicolumn{2}{@{}c}{\textbf{Molmo2-VC}} 
        & \multicolumn{3}{@{}c}{\textbf{Molmo2-VP}} \\ 
       \textbf{Model}  & 
       Acc.
        & Close acc.
        & Acc.
        & Close acc.
        & F1 & Recall & Precision\\
        \midrule
        \multicolumn{5}{@{}l}{\textbf{\textit{API call only}}} \\
        GPT-5~\cite{gpt5}             & 43.1	& 73.7 & \underline{35.8} & 50.3 & 4.1 &  4.4 & 4.2  \\
        GPT-5 mini~\cite{gpt5}         & 46.0 & 73.0 & 29.8 & 49.3 & 2.2 & 2.2 & 2.2  \\
          Gemini 3 Pro~\cite{gemini3}    & 44.0 & 71.7 & \textbf{37.1} & 53.1 & 20.0 & 27.4 & 19.8\\
        Gemini 2.5 Pro~\cite{comanici2025gemini}    & 41.6 & 70.0 & \underline{35.8} & \textbf{56.5} & 13.0 & 14.5 & 13.6\\
        Gemini 2.5 Flash~\cite{comanici2025gemini}   & 38.7 & 70.0 & 31.9 & 48.2 & 11.1 & 11.2 & 12.2 \\
        Claude Sonnet 4.5~\cite{anthropic2025sonnet}  & 42.4 & 72.6 & 27.2 & 45.1 & 3.5 & 3.7 & 4.3  \\
        \midrule
        
        \multicolumn{5}{@{}l}{\textbf{\textit{Open weights only}}} \\
        Qwen3-VL-4B~\cite{bai2025qwen3vltechnicalreport}        & 38.9 & 74.7 & 25.3 & 44.3 & 0.0 & 0.0 & 0.0 \\
        Qwen3-VL-8B~\cite{bai2025qwen3vltechnicalreport}         & 42.0 & 74.4 & 29.6 & 47.7 & 1.5 & 1.5 & 1.5\\
        \midrule
        
        
        \multicolumn{5}{@{}l}{\textbf{\textit{Molmo2 family: Open weights, Open data (no distillation), Open code}}} \\
        \textcolor{molmocolor}{Molmo2-4B}    & \underline{61.5} & \textbf{76.1} & 34.3 & \underline{56.1} & \textbf{39.9} & \textbf{42.7} & \textbf{39.4} \\
        \textcolor{molmocolor}{Molmo2-8B}    & 60.8 & 75.0 & 35.5 & 53.3 & \underline{38.4} & \underline{39.3} & \underline{38.7} \\
         \textcolor{molmocolor}{Molmo2-O-7B}    & \textbf{61.6} & \underline{76.0} & 33.2	& 50.5 & 35.8&35.8	 & 37.9 \\
    \end{tabular}
    \caption{\textbf{Video counting and pointing results.} 
    \model\ scores highest on BURST-VC and \model-VP and second highest on \model-VC's close accuracy, slightly behind Gemini 2.5 Pro.}
    \label{tab:video_count_and_point_results}
\end{table}

\begin{table*}[t]
    \renewcommand{\arraystretch}{0.98}
    \centering
    \setlength{\tabcolsep}{2.5pt}
    \resizebox{\textwidth}{!}{
    \begin{tabular}{@{}lccccccccccccccccc@{}}
        &{\textbf{MeViS~\cite{ding2023mevis}}} & \multicolumn{3}{@{}c}{\textbf{MeViS~\cite{ding2023mevis}}} & \multicolumn{3}{@{}c}{\textbf{Ref-YT-VOS~\cite{seo2020urvos}}} & \multicolumn{3}{@{}c}{\textbf{Ref-Davis~\cite{refdavis}}} & \multicolumn{3}{@{}c}{\textbf{ReasonVOS~\cite{bai2024one}}} \\
        & valid & \multicolumn{3}{@{}c}{valid-u} & \multicolumn{3}{@{}c}{valid} & \multicolumn{3}{@{}c}{valid} &
        \multicolumn{3}{@{}c}{test} \\
        \textbf{Model} & $\mathcal{J}\&\mathcal{F}$ & $\mathcal{J}\&\mathcal{F}$ & F1 & HOTA &  $\mathcal{J}\&\mathcal{F}$ & F1 & HOTA &  $\mathcal{J}\&\mathcal{F}$ & F1 & HOTA &  $\mathcal{J}\&\mathcal{F}$ & F1 & HOTA \\ 
        \midrule
        
        \multicolumn{10}{@{}l}{\textbf{\textit{API call only}}} \\
        GPT-5~\cite{gpt5}                            & 23.4 & 26.5 & 17.3 & 14.0 & 30.9 & 21.0 & 18.4 & 25.2 & 17.0 & 11.6 & 24.7 & 13.6 & 10.7 \\
        GPT-5 mini~\cite{gpt5}                       & 15.7 & 15.4 & 8.5  & 6.8  & 16.2 &  7.4 & 6.2  & 8.4  & 3.4  & 2.3  & 14.6 & 4.2  & 3.4 \\
        Gemini 3 Pro~\cite{gemini3}                  & 42.5 & 51.1 & 42.3 & 36.0 & 55.0 & 49.1 & 45.5 & 66.6 & 60.8 & 55.7 & 52.6 & 48.5 & 42.1\\ 
        Gemini 2.5 Pro~\cite{comanici2025gemini}     & 40.7 & 52.8 & 41.2 & 35.0 & 45.1 & 44.5 & 40.5 & 45.6 & 62.7 & 56.6 & 44.0 & 50.2 & 42.4 \\
        Gemini 2.5 Flash~\cite{comanici2025gemini}   & 27.6 & 31.8 & 24.0 & 19.9 & 36.0 & 32.8 & 30.0 & 31.6 & 36.7 & 30.0 & 26.5 & 25.8 & 21.0 \\     
        
        \midrule

        \multicolumn{10}{@{}l}{\textbf{\textit{Open weights only}}} \\
        Qwen3-VL-4B~\cite{qwen3technicalreport} & 29.7 & 30.6 & 23.3 & 18.7 & 32.1 & 29.0 & 26.5 & 44.4 & 33.1 & 26.9 & 26.5 & 17.0 & 13.5 \\
        Qwen3-VL-8B~\cite{qwen3technicalreport} & 35.1 & 34.4 & 30.1 & 23.8 & 48.3 & 42.1 & 37.6 & 41.0 & 41.6 & 33.2 & 24.9 & 22.3 & 17.5 \\
        \midrule
        
        \multicolumn{10}{@{}l}{\textbf{\textit{Specialized open models}}} \\
        VideoLISA~\cite{bai2024one}                      & 44.4 & 53.2 & -- & -- & 63.7 & -- & -- & 68.8 & -- & -- & 47.5  & -- & -- \\
        VideoGLaMM~\cite{rasheed2024glamm}               & 45.2 & 50.6 & -- & -- & 66.8 & -- & -- & 69.5 & -- & -- & 33.9 & -- & -- \\
        Sa2VA-8B~\cite{yuan2025sa2va}                    & 46.9 & 57.0 & -- & --  & \textbf{70.7} & -- & --  & \underline{75.2} & -- & -- & 55.5 & -- & -- \\
        Sa2VA-Qwen3-VL-4B~\cite{yuan2025sa2va}           & 36.7 & 57.1 & -- & --  & 68.1 & -- & -- &  \textbf{76.0}  & -- & -- & 50.0 & -- & -- \\
        Molmo~\cite{molmov1} + SAM 2~\cite{ravi2024sam2} & 46.9 & 51.5 & 53.8 & -- & 64.6 & 71.1 & -- & 65.2 & 74.5 & -- & 45.7 & 50.3 & -- \\
        VideoMolmo-7B~\cite{ahmad2025videomolmo}            & 53.9 & 57.0 & 59.4 & -- & 67.3 & 73.7 & -- & 72.5 & 75.4 & -- & 51.1 & 50.3 & -- \\

    \midrule
    \multicolumn{10}{@{}l}{\textbf{\textit{Molmo2 family: Open weights, Open data (no distillation), Open code}}} \\
    \textcolor{molmocolor}{Molmo2-4B}          &  \textbf{63.3} & \underline{70.0} & 75.5 & \underline{72.4} & \underline{70.2} & \textbf{80.4} & \textbf{78.8} & 73.5 & \textbf{83.1} & \textbf{81.1} & 61.9 & 66.5 & 64.0 \\ 
    \textcolor{molmocolor}{Molmo2-8B}          &  \underline{62.3} & \textbf{70.8} & \underline{75.9} & \textbf{72.6} & \underline{70.2} & \underline{78.7} & \underline{77.3} & 72.7 & \underline{81.3} & \underline{78.7} & \textbf{65.8} & \textbf{70.8} & \textbf{68.6} \\ 
    \textcolor{molmocolor}{Molmo2-O-7B}        &  58.4  & 69.7 & \textbf{76.1} & 72.3 & 67.9 & 77.7 & 76.1 & 70.4 & 79.2 & 76.0 & \underline{62.6} & \underline{67.5} & \underline{65.1} \\ 
    \end{tabular}
    }
    \caption{\textbf{Tracking Results on Academic Benchmark}. $\mathcal{J}\&\mathcal{F}$ is reported for specialized segmentation or points-to-segmentation models. F1 is the point accuracy measured for VLMs that can generate points per frame. HOTA~\cite{luiten2021hota} is the tracking accuracy that accounts for association accuracy for models that provide tracking IDs. }
    
    
    
    \label{tab:track_result_academic}
\end{table*}

\begin{table*}[!h]
    \renewcommand{\arraystretch}{0.98}
    \centering
    \setlength{\tabcolsep}{2.5pt}
    \resizebox{\textwidth}{!}{
    \begin{tabular}{@{}l ccc ccc ccc ccc ccc| ccc@{}}
        & \multicolumn{3}{@{}c}{\textbf{Animals}}   & \multicolumn{3}{@{}c}{\textbf{Person}} & \multicolumn{3}{@{}c}{\textbf{Sports}} & \multicolumn{3}{@{}c}{\textbf{Dancers}} & \multicolumn{3}{@{}c}{\textbf{Misc}} & \multicolumn{3}{@{}c}{\textbf{Overall}}\\
        \textbf{Model} & $\mathcal{J}\&\mathcal{F}$ & F1 & HOTA &  $\mathcal{J}\&\mathcal{F}$ & F1 & HOTA &  $\mathcal{J}\&\mathcal{F}$ & F1 & HOTA &  $\mathcal{J}\&\mathcal{F}$ & F1 & HOTA &  $\mathcal{J}\&\mathcal{F}$ & F1 & HOTA &  $\mathcal{J}\&\mathcal{F}$ & F1 & HOTA \\
        \midrule
        
        \multicolumn{10}{@{}l}{\textbf{\textit{API call only}}} \\
        GPT-5~\cite{gpt5}                            & 41.4 & 20.6 & 20.3 & 16.5 & 4.5  & 4.2 & 14.4 & 2.0 & 2.5 & 33.8 & 11.7 & 11.5 & 14.6 & 2.2 & 1.6 & 23.5 & 7.5 & 7.5 \\
        GPT-5 mini~\cite{gpt5}                       & 21.7 & 7.8  & 8.0  & 8.6  & 1.6  & 1.5 & 10.7 & 0.6 & 0.8 & 15.6 & 2.1  & 2.0  & 13.5 & 0.6 & 0.4 & 12.7 & 2.1 & 2.1 \\
        Gemini 3 Pro~\cite{comanici2025gemini}       & 70.4 & 62.3 & 60.0 & 44.5 & 30.7 &29.2 & 23.4 & 10.3& 8.8 & 55.6 & 44.3 & 37.8 & 35.3 & 18.3&	14.4 & 44.6 & 32.2 & 29.1\\
        Gemini 2.5 Pro~\cite{comanici2025gemini}     & 69.3	& 56.8 & 53.2 & 50.0 & 33.6	& 31.9 & 29.7 & 10.8 & 8.9 & 55.9& 39.4 & 32.2 & 34.7 & 17.6 & 18.3 & 47.9 & 31.2 & 27.8\\
        Gemini 2.5 Flash~\cite{comanici2025gemini}   & 58.0 & 46.6 & 44.4 & 38.9 & 21.4	& 20.1 & 13.2 & 6.2	 & 5.5 & 48.0 & 29.0 & 25.1	& 21.9 & 5.7 & 4.6 & 36.2 & 21.8 & 19.8 \\        
        \midrule

        \multicolumn{10}{@{}l}{\textbf{\textit{Open weights only}}} \\
        Qwen3-VL-4B~\cite{qwen3technicalreport} & 57.2 & 11.5 & 12.3 & 35.1 & 12.0 & 11.2 & 3.8 & 0.4 & 0.4 & 34.6 & 6.9 & 5.7 & 17.5 & 6.2 & 4.2 & 28.5 & 7.2 & 6.7\\
        Qwen3-VL-8B~\cite{qwen3technicalreport} & 63.8 & 52.3 & 50.2 & 35.4 & 20.3 & 18.9 & 5.2 & 1.7 & 1.4 & 31.3 & 19.0 & 16.7 & 16.3 & 6.2 & 4.2 & 28.7 & 18.0 & 16.5\\
        \midrule
        
        \multicolumn{10}{@{}l}{\textbf{\textit{Specialized open video models}}} \\
     
        VideoLISA~\cite{bai2024one}              & 67.8 & -- & -- & 35.8 & -- & -- & 32.9 & -- & --  & 53.6 & -- & -- & 25.8 & -- & -- & 43.3 & -- & --\\
        VideoGLaMM~\cite{rasheed2024glamm}       & 63.9 & -- & -- & 26.2 & -- & -- & 34.3 & -- & -- & 46.0 & -- & -- & 22.3 & -- & -- & 37.9 & -- & --\\
        Sa2VA-8B~\cite{yuan2025sa2va}            & 74.3 & -- & -- & 45.5 & -- & -- & 30.7 & -- & -- & 53.3 & -- & -- & \textbf{49.1} & -- & -- & 46.9 & -- & -- \\
        Sa2VA-Qwen3-VL-4B~\cite{yuan2025sa2va}   & 73.3 & -- & -- & 48.6 & -- & -- & 31.6 & -- & -- & 50.1 & -- & -- & 31.4 & -- & -- & 46.7 & -- & -- \\
        SAM 3~\cite{carion2025sam} & 41.1 & -- & -- & 35.2 & -- & -- & 43.3 & -- & -- & 29.2 & -- & -- & 36.8 & -- & -- & 36.3 & -- & -- \\
        Molmo~\cite{molmov1} + SAM 2~\cite{ravi2024sam2} & 71.8 & 76.0 &  -- & \textbf{52.7} & 7.0 & -- & 52.8 & 2.6 & -- & 51.7 & 7.55 & -- & 40.9 & \underline{37.5} & -- & 54.2 & 14.0 & -- \\
        VideoMolmo-7B~\cite{ahmad2025videomolmo} & 68.4 & 69.5 & -- & \underline{51.1} & 6.3 & -- & 43.2 & 2.1 & -- & 53.8 & 7.2 & -- & 39.9 & 30.8 & -- & 51.3 & 12.7 & -- \\

    \midrule
    \multicolumn{10}{@{}l}{\textbf{\textit{Molmo2 family: Open weights, Open data (no distillation), Open code}}} \\
    \textcolor{molmocolor}{Molmo2-4B}          &  \textbf{81.0} & \textbf{83.0} & \textbf{83.7} & 43.7 & \textbf{48.3} & \underline{47.7} & \underline{59.7} & \underline{53.1} & \underline{54.3} & \textbf{60.4} &  \textbf{64.4} & \textbf{64.4} & 43.1 & 35.1 & \underline{31.3} & \textbf{56.7} & \textbf{57.5} & \textbf{57.6} \\ 
    \textcolor{molmocolor}{Molmo2-8B}          &  \underline{80.1} & \underline{82.0} & \underline{83.0} & 43.1 & \underline{47.9} & \textbf{48.0} & \textbf{59.8} & \textbf{53.3} & \textbf{54.8} & \underline{59.9} & \underline{63.9} & \underline{63.5} & 41.6 & 31.5 & 29.7 & \underline{56.2} & \underline{57.1} & \underline{57.5} \\ 
    \textcolor{molmocolor}{Molmo2-O-7B}        &  \underline{80.1} & 81.9 & 82.8 & 41.5 & 45.5 & 45.4 & 54.1 & 47.6 & 48.6 & 57.7 & 61.0 & 60.3 & \underline{45.0} & \textbf{37.6} & \textbf{34.7} & 53.7 & 54.2 & 54.2 \\
     
    \end{tabular}
    }
    \caption{\textbf{Tracking results on \model-Track by video domain}. Overall is the accuracy across all samples.}    
    \vspace{0.5cm}
    \label{tab:molmo2_track_results}
\end{table*}

\noindent\textbf{Video counting and pointing.}
For counting, we also evaluate on BURST-VideoCount, a counting benchmark of 2.2k examples derived from the ground-truth tracks in the BURST test set~\cite{athar2023burst}. 
We report the close accuracy metric (correct if $|pred - gt| \leq \Delta$, where $\Delta = 1 + \lfloor 0.05\times gt \rfloor$), which rewards being close to the correct answer. For pointing, we build \model-VideoPointVal (\model{}-VP) by running SAM 2~\cite{ravi2024sam2} to gather object segmentation masks within a 3-second window centered around the annotated spatial-temporal points in \model-VideoPoint, and manually filter out examples with incorrect masks, leaving a total of 181 examples. For video pointing, we report the F1, recall, and prediction metrics, measuring how well the generated points match the ground-truth masks.


Results are shown in Table~\ref{tab:video_count_and_point_results}. \model{} is strong on the close metric, outperforming GPT 5.
For \model-VP, we carefully tune the prompts and try both point and bounding-box formats for our baseline models; however, we were unable to find a formulation that achieved very strong performance. Gemini Pro 3.0 reached the best score, but \model still significantly outperforms it.

\paragraph{Video object tracking.}
We evaluate video tracking on referring video object segmentation (VOS) benchmarks, where a point is considered correct if it lies within the ground truth segmentation mask. We additionally introduce \model-Track, a benchmark covering more diverse domains with complex object movements and occlusions, to evaluate \model{} on more challenging and realistic tracking tasks (see the appendix). Following~\cite{ahmad2025videomolmo}, we use SAM 2 to convert point predictions to segmentation masks for evaluation. We report the Jaccard and F-measure ($\mathcal{J}\&\mathcal{F}$) metrics for measuring segmentation quality across all frames, and the F1 score for the points at 1 fps. For API models, we generate the bounding box and extract their center points as they fail to generate accurate points. Tables~\ref{tab:track_result_academic}--\ref{tab:molmo2_track_results} show the results: 1) \model outperforms all baselines, including specialized segmentation models (in gray), across all benchmarks, particularly excelling on ReasonVOS and \model-Track, which require complex reasoning and occlusion handling skills. 2) Gemini 2.5 Pro is the strongest API model, but it still struggles to generate accurate object tracks.

\begin{table*}[!h]
    \renewcommand{\arraystretch}{0.98}
    \centering
    \setlength{\tabcolsep}{3pt}
    \resizebox{\textwidth}{!}{
    \begin{tabular}{@{}l>{\columncolor{tableyellow!50}}c
    >{\columncolor{tableyellow!50}}c
    >{\columncolor{tableyellow!50}}c
    >{\columncolor{tableyellow!50}}c
    >{\columncolor{tableyellow!50}}c
    >{\columncolor{tableyellow!50}}c
    >{\columncolor{tableyellow!50}}c
    >{\columncolor{tableyellow!50}}c
    >{\columncolor{tableyellow!50}}c
    >{\columncolor{tableyellow!50}}c
    >{\columncolor{tableyellow!50}}c
    >{\columncolor{tablegreen!10}}c
    >{\columncolor{tablegreen!10}}c
    >{\columncolor{tablegreen!10}}c
    >{\columncolor{tableyellow!50}}c
    >{\columncolor{tablegreen!10}}c
    >{\columncolor{tableblue!10}}c}
        \textbf{Model} & 
        \mycell{AI2D}{test~\cite{ai2_diagram}}
        & \mycell{ChartQA}{test~\cite{chartqa}}
        & \mycell{DocVQA}{test~\cite{mathew2021docvqa}}
        & \mycell{InfoQA}{test~\cite{infoqa}}
        & \mycell{TextVQA}{val~\cite{textqa}}
        & \mycell{VQA v2.0}{val ~\cite{goyal2017making}}
        & \mycell{RWQA}{\cite{realworldqa}}
        & \mycell{MMMU}{val~\cite{yue2024mmmu}}
        & \mycell{MathVista}{testmini~\cite{lu2024mathvista}}
        & \mycell{CountBench}{~\cite{beyer2024paligemma}}
        & \mycell{PixMoCount}{test \cite{molmov1}}
        & \mycell{MuirBench}{\cite{wang2024muirbench}}
        & \mycell{MMIU}{\cite{meng2024mmiumultimodalmultiimageunderstanding}}
        & \mycell{Blink}{val \cite{fu2024blink}}
        & \newcell{Img QA avg.}
        & \newcell{MultiImg QA avg.}
        & \newcell{Average}\\
        \midrule
        
        \multicolumn{16}{@{}l}{\textbf{\textit{API call only}}} \\

        GPT-5~\cite{gpt5}        & 97.1 & 89.6 & 88.9 & 83.0 & 78.7 & 79.7 & 80.8 & 81.8 & 82.7 & 90.8 & 67.2 & 78.6 & 71.0 & 66.5 & 83.7 & 72.1 & 81.2 \\

        GPT-5 mini~\cite{gpt5}  & 95.8 & 88.2 & 86.7 & 82.2 & 79.1 & 72.1 & 77.0 & 78.7 & 79.2 & 87.1 & 74.4 & 71.4 & 64.5 & 68.7 & 81.9 & 68.2 & 78.9 \\

        Gemini 3 Pro~\cite{gemini3} & 98.7 & 93.7 & 87.1 & 86.9 & 74.1 & 74.1 & 73.6 & 85.2 & 89.1 & 96.1 & 90.0 & 86.1 & 72.1 & 87.4 & 86.2 & 81.9 & 85.3 \\

        Gemini 2.5 Pro~\cite{comanici2025gemini} & 94.3 & 82.7 & 91.5 & 82.0 & 70.3 & 67.1 & 77.4 & 79.6 & 84.6 & 90.8 & 73.8 & 74.5 & 68.9 & 73.7 & 81.3 & 72.4 & 79.4 \\

        Gemini 2.5 Flash~\cite{comanici2025gemini}   & 95.9 & 76.8 & 91.1 & 80.9 & 73.0 & 69.4 & 74.5 & 79.0 & 81.2 & 86.7 & 63.9 & 73.5 & 61.2 & 70.2 & 79.3 & 68.3 & 76.9 \\

        Claude Sonnet 4.5~\cite{anthropic2025sonnet}  & 91.5 & 88.1 & 91.7 & 65.9 & 67.2 & 77.0 & 61.1 & 77.8 & 73.1 & 87.3 & 58.3 & 59.6 & 54.1 & 64.8 & 76.3 & 59.5 & 72.7 \\

        \midrule
        
        \multicolumn{16}{@{}l}{\textbf{\textit{Open weights only}}} \\

        InternVL3.5-4B~\cite{wang2025internvl3} & 82.6 & 86.0 & 92.4 & 78.0 & 77.9 & 78.1 & 66.3 & 66.6 & 77.1 & 82.2 & 62.4 & 53.1 & 49.2 & 58.1 & 77.2 & 53.5 & 72.1 \\

        InternVL3.5-8B~\cite{wang2025internvl3} & 84.0 & 86.7 & 92.3 & 79.1 & 78.2 & 79.5 & 67.5 & \textbf{73.4} & 78.4 & 79.6 & 61.9 & 55.8 & 49.4 & 59.5 & 78.2 & 54.9 & 73.2 \\

        Qwen3-VL-4B~\cite{qwen3technicalreport} & 84.1 & 84.6 & \underline{95.3} & 80.3 & 81.0 & 81.7 & 70.9 & 67.4 & 73.7 & 85.5 & 58.0 & 63.8 & 43.2 & \underline{65.8} & 78.4 & 57.6 & 73.9 \\

        Qwen3-VL-8B~\cite{qwen3technicalreport}      & 85.7 & \underline{89.6} & \textbf{96.1} & \textbf{83.1} & 82.8 & 82.3 & 71.5 & 69.6 & 77.2 & 90.4 & 65.0 & \underline{64.4} & 35.3 & \textbf{69.1} & \underline{81.2} & 56.3 & \underline{75.9} \\

        Keye-VL-1.5-8B~\cite{yang2025kwai}    & 89.5 & \textbf{94.1} & 93.4 & 74.9 & 81.5 & 79.3 & 73.5 & \underline{71.4} & \textbf{81.2} & 81.6 & 57.4 & 51.2 & 50.3 & 54.9 & 79.8 & 52.1 & 73.9 \\

        GLM-4.1V-9B~\cite{glmv}      & 87.9 & 70.0 & 93.3 & 80.3 & 79.6 & 68.3 & 70.7 & 68.0 & \underline{80.7} & 88.0 & 60.7 & \textbf{74.7} & \textbf{62.4} & 65.1 & 77.0 & \textbf{67.4} & 75.0 \\

        MiniCPM-V-4.5-8B~\cite{yu2025minicpmv45cookingefficient}  & 86.5 & 87.4 & 94.7 & 73.4 & 82.2 & 64.1 & 72.1 & 67.7 & 79.9 & 83.9 & 62.8 & 53.3 & 46.5 & 42.0 & 77.7 & 47.3 & 71.2 \\

        Eagle2.5-8B~\cite{eagle2_5}    & 84.5 & 87.5 & 94.1 & \underline{80.4} & 83.7 & 82.4 & \underline{76.7} & 55.8 & 67.8 & 90.2 & 90.2 & 61.8 & 48.4 & 45.8 & \underline{81.2} & 52.0 & 75.0 \\

        \midrule
        
        \multicolumn{16}{@{}l}{\textbf{\textit{Open models}}}  \\
        PLM-3B~\cite{cho2025PerceptionLM} &  90.9 & 84.3 & 93.8 & 74.6 & 84.3 & 84.4 & 72.4 & 41.2 & 59.1 & 87.1 & 63.0 & 25.7 & 40.6 & 55.4 & 75.9 & 40.6 & 68.3 \\

        PLM-8B~\cite{cho2025PerceptionLM}& 92.7 & 85.5 & 94.6 & 80.0 & \textbf{86.5} & 85.6 & 75.0 & 46.1 & 59.9 & 91.8 & 68.0 & 23.5 & 27.4 & 56.0 & 78.7 & 35.7 & 69.5 \\
        \midrule
        \multicolumn{16}{@{}l}{\textbf{\textit{Molmo1 family: Open weights, Open data (no distillation), Open code}}} \\
    \textcolor{molmocolor}{MolmoE-1B}~\cite{molmov1} &
86.4 & 78.0 & 77.7 & 53.9 & 78.8 & 83.9 & 60.4 & 34.9 & 34.0 & 87.2 & 79.6 & - & - & - & 68.6 & - & - \\
\textcolor{molmocolor}{Molmo-7B-O}~\cite{molmov1} &
90.7 & 80.4 & 90.8 & 70.0 & 80.4 & 85.3 & 67.5 & 39.3 & 44.5 & 89.0 & 83.3 & - & - & - & 74.6 & - & - \\
\textcolor{molmocolor}{Molmo-7B-D}~\cite{molmov1} &
93.2 & 84.1 & 92.2 & 72.6 & 81.7 & 85.6 & 70.7 & 45.3 & 51.6 & 88.5 & 84.8 & - & - & - & 77.3 & - & - \\
\textcolor{molmocolor}{Molmo-72B}~\cite{molmov1} &
\textbf{96.3} & 87.3 & 93.5 & 81.9 & 83.1 & 86.5 & 75.2 & 54.1 & 58.6 & 91.2 & 85.2 & - & - & - & \underline{81.2} & - & - \\
        \midrule
        \multicolumn{16}{@{}l}{\textbf{\textit{Molmo2 family: Open weights, Open data (no distillation), Open code}}} \\
        \textcolor{molmocolor}{Molmo2-4B} & 95.6 & 86.1 & 87.8 & 78.6 & 85.0 & \underline{86.6} & 75.4 & 50.9 & 56.7 & \underline{93.9} & 88.1 & 60.5 & \underline{55.5} & 57.5 & 80.4 & \underline{57.8} & 75.6 \\
        \textcolor{molmocolor}{Molmo2-8B} & \underline{95.8} & 86.0 & 93.2 & 80.1 & \underline{85.7} & \textbf{87.0} & \textbf{77.6} & 53.0 & 58.9 & 93.7 & \underline{88.5} & 63.7 & 54.2 & 51.3 & \textbf{81.7} & 56.4 & \textbf{76.3} \\
        \textcolor{molmocolor}{\model-O-7B} & 93.7 & 84.9 & 90.4 & 77.9 & 84.7 & \underline{86.6} & 73.6 & 45.8 & 54.2 & \textbf{95.1} & \textbf{88.9} & 58.4 & 51.7 & 50.5 & 79.7 & 53.5 & 74.1 \\
    \end{tabular}
    }%
    \caption{\textbf{Image benchmark results} for a range of proprietary APIs, open-weight baselines, and our Molmo2 family across image understanding and counting benchmarks. The result of the best-performing open-weight model is in \textbf{bold}. The Molmo1 models do not support multi-image input, so those evaluations are left blank.}
    \label{tab:image_benchmark_results}
\end{table*}

\subsection{Image results}

We present image and multi-image benchmark results in Table~\ref{tab:image_benchmark_results}. We follow the evaluation protocol from Molmo~\cite{molmov1} and report the same 11-benchmark average for single-image benchmarks. As with videos, we collect results for all models by testing them ourselves if needed.

Generally, \model{} robustly outperforms previous open-data models. \model{} is a bit behind the best open-weight model on OCR-heavy benchmarks (such as DocVQA or InfoQA) but performs well on general QA tasks, including state-of-the-art performance on VQA v2.0 and RealWorldQA (RWQA). Counting is also a strength, most notably on the challenging PixMo-Count test set. However, \model{} is behind on open-weight reasoning benchmarks (MathVista, MMMU), possibly due to the lack of multi-modal reasoning training data.
On multi-image tasks, \model{} performs competitively with most open-weight models, with the exception of GLM-4.1V-9B, which is notably ahead of all other models.

\begin{table*}[!h]
\renewcommand{\arraystretch}{0.98}
\centering
\small
\centering
\begin{tabular}{@{}lcccccc@{}}
\textbf{Model} & \textbf{Affordance} & \textbf{Spatial} & \textbf{Reasoning} & \textbf{Steerability} & \textbf{Counting} & \textbf{Average} \\
\midrule
Human & 92.3 & 83.6 & 87.8 & 86.3 & 95.6 & 89.1 \\
\midrule
\multicolumn{7}{@{}l}{\textbf{\textit{API call only}}}  \\
Gemini-Robotics-ER-1.5~\cite{abdolmaleki2025gemini} & 69.7 & 69.7 & 60.1 & \textbf{67.5} & \underline{68.5} & 67.1 \\
Gemini-2.5-Pro~\cite{comanici2025gemini} & 72.7 & 70.3 & 71.0 & 41.0 & 59.2 & 62.8 \\

\midrule
\multicolumn{7}{@{}l}{\textbf{\textit{Open weights only}}}  \\
Poivre-7B~\cite{poivre} & - & - & - & - & - & 67.5 \\
Qwen2.5-VL-32B-Instruct~\cite{qwen2} & 76.8 & 60.0 & 54.4 & \underline{46.5} & 57.1 & 59.0 \\
Qwen2.5-VL-72B-Instruct~\cite{qwen2} & 76.8 & 60.0 & 54.4 & \underline{46.5} & 57.1 & 59.0 \\
Qwen3VL~\cite{qwen3technicalreport} & 81.3 & 65.6 & 60.6 & 23.5 & 61.2 & 58.5 \\
Qwen3-VL-235B-A22B-Instruct~\cite{qwen3technicalreport} & - & - & - & - & - &  58.3 \\

\midrule
\multicolumn{7}{@{}l}{\textbf{\textit{Open models}}}  \\
VisionReasoner-7B~\cite{liu2025visionreasoner} & - & - & - & - & - & 64.7 \\

\midrule
\multicolumn{7}{@{}l}{\textbf{\textit{Molmo1 family: Open weights, Open data (no distillation), Open code}}} \\
\textcolor{molmocolor}{Molmo-7B-D~\cite{molmov1}} & 82.8 & 67.7 & 70.5 & 28.5 & 58.7 & 61.6 \\
\textcolor{molmocolor}{Molmo-72B~\cite{molmov1}} & \textbf{87.9} & 70.3 & 69.4 & 37.0 & 54.6 & 63.8 \\
\textcolor{molmocolor}{Molmo-7B-O~\cite{molmov1}} & \underline{84.9} & 63.1 & 63.2 & 45.5 & 59.7 & 63.3 \\
\midrule
\multicolumn{7}{@{}l}{\textbf{\textit{Molmo2 family: Open weights, Open data (no distillation), Open code}}} \\
\textcolor{molmocolor}{\model-4B} & 82.3 & \textbf{71.8} & \textbf{72.0} & 41.0 & \underline{71.4} & \underline{67.7} \\
\textcolor{molmocolor}{\model-8B} & 84.8 & \underline{71.3} & \underline{71.5} & 44.5 & \underline{71.4} & \textbf{68.7} \\
\textcolor{molmocolor}{\model-O-7B} & 81.8 & 69.7 & 69.4 & 39.0 & \textbf{72.4} & 66.5 \\
\end{tabular}
\caption{\textbf{Point-Bench results}\tablefootnote, baseline scores taken from the Point-Bench leaderboard. Qwen3-VL-235B-A22B-Instruct and VisionReasoner-7B scores were taken from their evaluation in Poivre~\cite{poivre}, which did not include sub-category scores.}
\label{tab:point_bench}
\end{table*}

\footnotetext{An older version of this report include higher scores that were the result of an evaluation bug.}

We evaluate image pointing on Point-Bench~\cite{pointarena}, results are in Table~\ref{tab:point_bench}. \model{} surpasses all other models on the Point-Bench leaderboard\footnote{As of 12/15/25} and the recent dedicated pointing model Poivre~\cite{poivre}. We attribute the gain on pointing compared to Molmo to the improved vision encoder, pointing pre-training, and token-weighting.

\subsection{Ablations and specialized models}
Next, we present ablations on our model, training strategy, and data. To avoid the high compute cost of training the full model, we train specialized 4B models on subsets of our data and use them for ablations. These tables use \colorbox{baselinecolor}{Gray} rows to show specialized models with default settings; key takeaways are in the captions.

\begin{table*}[t]
\centering
\subfloat[
\textbf{Caption Specialization}. Joint training with other video data improves the video caption performance.
\label{tab:caption_specialization}
]{
\begin{minipage}{0.45\linewidth}{\begin{center}
\small
\begin{tabular}{lcc}
Data & QA avg. & Cap. F1  \\
\hline
\baseline{Video-Only} & 64.8 & 39.5   \\
\model{}-Cap Only & - & \underline{35.8}   \\
\multicolumn{3}{c}{~} \\
\multicolumn{3}{c}{~} \\
\end{tabular}
\end{center}}\end{minipage}
}\hfill
\subfloat[
\textbf{Modeling}. Bidirectional attention, token weighting, and time tokens significantly improve performance, while a larger pool size degrades video captioning.
\label{tab:model_ablation}
]{
\begin{minipage}{0.45\linewidth}{\begin{center}
\small
\begin{tabular}{lcc}
 Model & QA avg. & Cap. F1  \\
\hline
\baseline{Video-Only} & \textbf{64.8} & \underline{39.5}  \\
No bidir & {64.4} & 38.5   \\
No token weighting & 64.0 & \textbf{40.0} \\
No time tokens & \underline{64.5} & {37.4}   \\
Video pool size 3x3 to 4x4 & 64.3 & {37.0}   \\

\multicolumn{3}{c}{~} \\
\multicolumn{3}{c}{~} \\
\end{tabular}
\end{center}}\end{minipage}
}\hfill
\subfloat[
\textbf{Video SFT data}. Both Molmo2-Cap and Molmo2-QA improve performance compared to academic datasets only. 
\label{tab:sft_data}
]{
\begin{minipage}{0.45\linewidth}{\begin{center}
\small
\begin{tabular}{lcc}
Data & QA avg. & Cap. F1  \\
\hline
Academic & 62.9 & 5.0  \\
+ QA & 64.5 & 17.2   \\
+ Cap & \textbf{65.3} & \underline{38.4} \\
\baseline{+ Cap/QA} & \underline{64.8} & \textbf{39.5} \\
\multicolumn{3}{c}{~}
\end{tabular}
\end{center}}\end{minipage}
}\hfill
\subfloat[
\textbf{Caption data}. Using the video and frame merged caption (VF) is critical, but adding video (V) and/or frame (F) captions does not bring improvements.
\label{tab:caption_data}
]{
\begin{minipage}{0.45\linewidth}{\begin{center}
\small
\begin{tabular}{lccc}
Data & Cap. R & Cap. P & Cap. F1  \\
\hline	
V & 13.3 & \textbf{66.7} & 22.1 \\
VF & \underline{25.4} & 59.5 & {35.5}  \\
\baseline{VF+V} & \textbf{25.6} & \underline{59.6} & \textbf{35.8}\\
VF + F &  {22.4} & 59.4 & 35.6\\
{VF + V + F} & 22.6 & {57.3} & \underline{35.7}\\
\end{tabular}
\end{center}}\end{minipage}
}
\caption{\textbf{Video ablations.} For ablations (a)(b)(c) we train models on only video data; ablation (d) has models with only video captions.}
\label{tab:all_video_ablations}
\end{table*}

\begin{table*}[h!]
\centering
\subfloat[
\textbf{Counting strategy}. Pointing is the key ingredient in \model's counting abilities. 
\label{tab:counting_strategy}
]{
\begin{minipage}{0.4\linewidth}{\begin{center}
\small
\begin{tabular}{lcc}
Strategy & BVC &  MVC  \\
\hline
Count & 61.3 & 28.1  \\
\baseline{Point then count} & \textbf{61.5} & \textbf{34.5}  \\
\multicolumn{3}{c}{~}\\
\end{tabular}
\end{center}}\end{minipage}
}\hfill
\vspace{.3cm}
\subfloat[
\textbf{Data source}. Including both \model- and AcademicVideoPoints performs the best overall.
\label{tab:counting_data_source}
]{
\begin{minipage}{0.45\linewidth}{\begin{center}
\small
\begin{tabular}{lccc}
Data & BVC & MVC & MVP \\
\hline
\baseline{Both} & \underline{61.5} & \textbf{34.5}  & \underline{31.8}\\
\model-VP & 60.0 & \underline{34.3} & \textbf{35.0}\\
Academic-VP & \textbf{61.6} & 9.0 & 9.0\\
\end{tabular}
\end{center}}\end{minipage}
}\hfill
\subfloat[
\textbf{Sampling strategy}. Upsampling medium and high-count examples helps on MVC and MVP.
\label{tab:counting_upsampling}
]{
\begin{minipage}{0.45\linewidth}{\begin{center}
\small
\begin{tabular}{lccc}
Upsampling & BVC & MVC & MVP\\
\hline
\baseline{Med-high} & 61.5 & \textbf{34.5} & \textbf{31.8}\\
No & \textbf{62.4} & 32.1 & 28.1 \\
\multicolumn{3}{c}{~}\\
\end{tabular}
\end{center}}\end{minipage}
}
\caption{\textbf{Counting and pointing ablations.} BVC represents Burst-VideoCount accuracy; and MVC and MVP are \model-VideoCount accuracy and \model-VideoPoint F1 on the validation sets.}
\label{tab:point_ablations_all}
\end{table*}


\begin{table*}[ht!]
\centering
\subfloat[
\textbf{Adding pointing}. Training with pointing tasks helps tracking performance.
\label{tab:tracking_mixture}
]{
\begin{minipage}{0.4\linewidth}{\begin{center}
\small
\begin{tabular}{lccc}
Model &  $\mathcal{J}\&\mathcal{F}$ &  F1 & HOTA  \\
\hline
\baseline{Tracking only} & \underline{64.9} & \underline{70.0} & \underline{68.4} \\
Tracking + Pointing & \textbf{65.7} & \textbf{71.1} & \textbf{69.4} \\
\end{tabular}
\end{center}}\end{minipage}
}\hfill
\subfloat[
\textbf{Tracking data source}. We see progressive improvements from academic VOS, bounding box (bbox) tracks, to \model data. 
\label{tab:tracking_data_source}
]{
\begin{minipage}{0.55\linewidth}{\begin{center}
\small
\begin{tabular}{lccc}
Data &  $\mathcal{J}\&\mathcal{F}$ &  F1 & HOTA  \\
\hline
Academic (VOS)  & \underline{64.3} & 68.8 & 66.7 \\
+ Academic (bbox)  & 63.9 & \underline{69.3} & \underline{67.5} \\
\baseline{+ \model (VideoTrack)} & \textbf{64.9} & \textbf{70.0} & \textbf{68.4} \\
\end{tabular}
\end{center}}\end{minipage}
}\hfill
\subfloat[
\textbf{Tracking sub-tasks} ablated on Academic VOS only. Temporal grounding helps, while single-point object tracking slightly degrades performance. 
\label{tab:tracking_tasks}
]{
\begin{minipage}{.65\linewidth}{\begin{center}
\small
\begin{tabular}{lccc}
Strategy &  $\mathcal{J}\&\mathcal{F}$ &  F1 & HOTA \\
\hline
Tracking  &  64.2 & 68.4 & 66.2 \\
+ Temporal grounding & \textbf{64.8} & \textbf{69.4} & \textbf{67.2} \\
+ Single-point object tracking  & \underline{64.3} & \underline{68.8} & \underline{66.7}  \\
\end{tabular}
\end{center}}\end{minipage}
}\hfill

\caption{\textbf{Tracking ablations.} We report average metrics across the five tracking benchmarks (the valid-u split for MeViS). HOTA~\cite{luiten2021hota} measures association accuracy.}
\label{tab:tracking_ablations_all}
\end{table*}

\paragraph{Video ablations}.
Table~\ref{tab:all_video_ablations} shows results and ablations with video-only and video-captioning-only data. We see that video QA data transfers positively to captioning (Table \ref{tab:caption_specialization}) and vice versa (Table \ref{tab:sft_data}).
Table~\ref{tab:model_ablation} shows bi-directional attention and token-weighting both boost QA performance, although token-weighting can slightly degrade caption performance.
Meanwhile, removing frame timestamps diminishes both metrics, indicating that including temporal information is important, especially for captioning.
Increasing the video pool size from 3x3 to 4x4 slightly lowers QA performance but causes a significant drop in captioning quality. 
We believe that this is because the video benchmarks are relatively high-level and do not require understanding small details, so decreasing the pooling size is not very harmful. This illustrates the importance of tracking the captioning metric in addition to the other benchmarks, which requires a much more fine-grained understanding of the video.
Finally, captioning models based solely on human transcripts (V) produce worse results than those that include frame-level captions (VF), but training on a mixture of these captions does not lead to improvements (\ref{tab:caption_data}).




\paragraph{Video counting and pointing.}
Table~\ref{tab:point_ablations_all} reports the performance of a specialized pointing model and ablating counting strategy, data, and data sampling. We observe that our two sources of pointing are complementary (Table~\ref{tab:counting_data_source}), that pointing before counting is much better than directly predicting the count (Table~\ref{tab:counting_strategy}), and that upsampling high-frequency points improves both counting and pointing (Table~\ref{tab:counting_upsampling}). 
\paragraph{Video object tracking.}
Table~\ref{tab:tracking_ablations_all} shows ablations on task mixtures and data sources for tracking with a model trained only on our tracking data. Including our video pointing data improves performance, showing a moderate transfer from pointing to tracking (Table~\ref{tab:tracking_mixture}). Using bounding box tracks and the \model-VideoTrack dataset also leads to improvements (Table~\ref{tab:tracking_data_source}). Supporting temporal grounding helps, while adding point-based single object tracking causes a slight degradation (Table~\ref{tab:tracking_tasks}).

\begin{table}[!ht]
    \centering
    \small
    \begin{tabular}{l c c c c}
        Post-training  & Short video QA & Long video QA & Molmo2 Video Cap. & Image QA\\
\hline
\baseline{With long-context SFT} & 69.4 &	67.4 & 39.9 & 80.6\\
No long-context SFT & 69.6 & 64.4 & 42.3 & 80.5 \\
    \end{tabular}
        \caption{\textbf{Long-context SFT ablation}. Columns show the average of our 12 video benchmarks divided by short/long video benchmarks, using validation sets for EgoSchema, PerceptionText, and MLVU, video captioning F1, the average of the 11 image benchmarks using validation sets for InfoQA, DocQA, ChartQA, VQA v2, and AI2D.}
    \label{tablong_context_ablation}
\end{table}
\paragraph{Long context SFT.}
We compare the \model-4B performance before and after long-context post-training in Table~\ref{tablong_context_ablation}. We find that long-context post-training significantly improves model performance on long video QA benchmarks, while the video caption performance drops and performance on short video QA benchmarks and image QA benchmarks do not significantly change.

\section{Related works}
\label{appendix:related}
\paragraph{Multimodal LLMs.} Multimodal LLM models have become popular in the last few years for image understanding and grounding tasks~\cite{molmov1, lai2023lisa, team2025gemma}. A common strategy for multimodal LLMs is to use CLIP-style image encoders and align image embeddings with the LLM input space via a connector module~\cite{molmov1, liu2024llava665k}. Video LLMs also commonly extend the CLIP-style image encoding and use image embedders to individually embed each frame in a video~\cite{eagle2_5, cho2025PerceptionLM, maaz2024videochatgpt}. Some have explored using pretrained video encoders in combination with per-frame encoding or encoding $2$ frames together~\cite{apollo, wang2024qwen2vl, glmv}, but using video encoders with more frames lags behind using image encoders (such as SigLIP 2~\cite{siglip2}). However, when encoding each frame of a video individually, the number of visual tokens increases linearly with the frame sampling rate and the length of the video. This leads to a high compute cost and has led to a rise in works exploring efficient video encodings~\cite{shen2024longvu, xu2024slowfast, yang2025kwai, li2024videochat, wang2025internvl3}.

The best performing video LLMs~\cite{gpt5, anthropic2025sonnet, comanici2025gemini} are closed-source proprietary models. While they are very capable, not much is known about how these models are trained and what data they use. By contrast, while some open weight models have been released~\cite{wang2024qwen2vl, wang2025internvl3, yang2025kwai, apollo, eagle2_5}, most don't release their training recipes or don't release their training data. A few projects do release all the training details and data~\cite{cho2025PerceptionLM, llava_video}, but use biased data generated by proprietary VLMs (such as GPT4 and LLaMA3~\cite{chen2023sharegpt4v, llama3}). Hence, there is a need for a fully open SoTA training pipeline for Video LLMs that does not use previously trained multimodal LLMs to generate data.

\paragraph{Video-language instruction tuning datasets.} The popularity of Video LLMs has also led to an increase in methods to develop instruction-tuning data for them. The current dominant paradigm involves generating synthetic instruction data by first segmenting videos into clips, generating descriptive captions for each clip, and then using a powerful LLM to synthesize video-level captions and QA pairs~\cite{llava_video, eagle2_5, cho2025PerceptionLM, sharegpt4video}. However, a critical limitation of these approaches is their reliance on closed-source Video-Language Models (VLMs) for the initial clip captioning step. This introduces an inherent, often proprietary, bias into the generated data, as the underlying VLM's training data and biases are inaccessible to the research community.

Our \model-CapQA dataset is generated through a similar pipeline but utilizes a video captioner trained on our fully open \model-Cap to generate video captions. We segment each video into multiple scenes, caption each scene, and then provide these to an LLM along with the video metadata to generate 1M QA pairs. Another strategy used for generating QA pairs is to have annotators work with an LLM provided with an image caption when generating QA pairs~\cite{molmov1}, and we extend the same to video data to generate our \model-AskModelAnything.

\paragraph{Video tracking.}
Early video tracking focused on bounding boxes for a closed set of objects~\cite{muller2018trackingnet, dendorfer2020mot20}. Since then, the field has branched into specific subtasks, including track any point (TAP) ~\cite{karaev2024cotracker3, doerchtapvid} and tracking object segmentations ~\cite{hong2023lvos, athar2023burst}. Object segmentations improved accuracy and granularity, but tracking was still limited to a closed set of objects. Moving beyond a closed set of objects to an open vocabulary has led to a rise in language-guided video object segmentation (VOS)~\cite{yan2024visa}. A variety of new specialized models have been trained to track object~\cite{bai2024one, li2025refsam, ahmad2025videomolmo}.
Unlike \model{}, these models are specialized and do not support other capabilities.

Previous methods, like Ref-VOS~\cite{refvos} and MeVis~\cite{ding2023mevis}, support the language-guided VOS task by augmenting existing tracking datasets with complex referring expressions. However, we noticed a lack of language prompts referring to multiple objects or diverse actions. For our \model-VideoTrack dataset, we similarly add to existing datasets by asking annotators to craft non-trivial text queries that apply to object tracks, with a focus on queries that describe multiple objects. For segmentation masks, we source videos and tracks from diverse open-source segmentation tracks~\cite{refvos,mosev2,vpseg,ravi2024sam2} and use a data pipeline to produce masks from bounding-box tracks{~\cite{sun2022dancetrack, zhang2023animaltrack, scott2024teamtrack, wang2022sportstrack, giancola2018soccernet, dendorfer2020mot20, zheng2024nettrack, du2018unmanned, varga2022seadronessee, yu2020bdd100k}}. 

\paragraph{Video pointing.}
Multimodal LLMs that support point grounding in an image have recently become quite common~\cite{molmov1, poivre, comanici2025gemini, abdolmaleki2025gemini, bai2025qwen3vltechnicalreport, pointarena}. The training data used in these works is collected using automated object detectors, using existing referring expression datasets~\cite{yu2016modeling, GRES, Krishna2016VisualGC} or through manual human annotation~\cite{molmov1}. We extend the human annotation pipeline approach to videos by adding a frame-selection phase. We also propose generating some queries through an LLM based on the caption to ensure the queries are complex and diverse.

\section{Conclusion} Open research needs open-source. \model{} supports open science by closing the gap between proprietary VLMs and the rest of the community.

\clearpage

\section*{Author Contributions}
\label{sect:contrib}


Christopher Clark, Jieyu Zhang, Zixian Ma, JaeSung Park, Rohun Tripathi, Sangho Lee and Mohammadreza Salehi collectively contributed to dataset construction,
model training, and conducted numerous exploratory experiments for this project. 

\textbf{Christopher Clark} led the project and focused on video modeling and training strategies, including experiments with the SFT mixture, the pre-training approach, and video modeling. He also wrote much of the core training code and implemented the packing and message tree systems.\\
\textbf{Jieyu Zhang} co-led the data effort on video datasets. He collected and filtered raw videos for Molmo2 video caption, video QA, and video pointing datasets, and contributed to the curation of these datasets. He helped the integration of other training/evaluation datasets and ran evaluations for many baseline models. He also helped add subtitle understanding to the model and ablations of the video SFT/caption models.\\
\textbf{Zixian Ma} co-led the data effort on video datasets. She designed human data collection interfaces and implemented them with help from Yinuo Yang. She collected the Molmo2-Cap, Molmo2-AskModelAnything, and Molmo2-VideoPoint datasets via Prolific. She led the training ablations on video counting and pointing and helped integrate academic training datasets. She ran the human preference and NLP evaluations. \\
\textbf{Jae Sung Park} led the effort to add tracking capability to Molmo2 as points. Together with Zhongzheng Ren and Vincent Shao, he designed the Molmo2-Track human annotation collection, curated existing academic tracking datasets for training, and built the pipeline to extract accurate point tracks. He introduced auxiliary grounding and single-point tracking objectives and performed ablations on mixtures of video tracking tasks. He and Zhongzheng Ren designed tracking evaluations across diverse VLMs and segmentation models. \\
\textbf{Mohammadreza Salehi} led the long-context post-training and co-led sourcing videos for training. He also contributed to training dataset construction, training on a mixture of images and videos, and evaluation of Molmo and API models. \\
\textbf{Rohun Tripathi} primarily worked on efficient modeling strategies. He developed learned and training free solutions to token allocation for different frames, with and without the input query. He implemented the initial training pipeline and details such as 3D position encoding and time tokens. He helped with training/evaluation set integrations, with a focus on long video understanding.\\
\textbf{Sangho Lee} led improvements to image modeling and training strategies and extended them to the multi-image setting. He also supported and directly conducted extensive ablation studies to develop effective training strategies for video modeling. In addition, he implemented the Hugging Face model and processor code and vLLM integrations.\\
\textbf{Chris Dongjoo Kim} led the data effort for multi-image datasets. In collaboration with Weikai Huang and Sangho Lee, he curated the MultiImageQA dataset. He also held full responsibility for the multi-image pointing capability, including dataset curation algorithms and model training. \\
\textbf{Yue Yang} led data curation for text-rich multi-image datasets, synthetically generating diverse question-answer pairs grounded in images such as charts, tables, and documents.\\
\textbf{Zhongzheng Ren}, \textbf{Yinuo Yang}, \textbf{Vincent Shao}, \textbf{Weikai Huang}, and \textbf{Ziqi Gao} all made significant dataset contributions.\\
\textbf{Jitesh Jain}, \textbf{Jianrui Zhang}, and \textbf{George Stoica} contributed to research discussions throughout the project and did exploratory experiments based on \model{}.\\
\textbf{Taira Anderson} managed the project.\\
\textbf{Winson Han} designed the figures in this report.\\
\textbf{Ali Farhadi} advised the project.\\
\textbf{Ranjay Krishna} was the PI for the project.

\section*{Acknowledgements}
This work would not be possible without the support of our colleagues at Ai2.

\begin{itemize}
    \item We thank David Albright, Erin Bransom, Kristin Cha, Yvonne Chou, Karen Goodfellow, Malachi Hamada, Stephen Kelman, Ryan Kiskis, Sophie Lebrecht, Kelsey MacMillan, Crystal Nam, Lauren Olvera, Carissa Schoenick, Jeremy Tryba, Tina Weiss, Kyle Lo, Kyle Wiggers, and Will Smith for their important work for the \model public release.
    \item We thank the Ai2 Playground team, including Taylor Blanton, Byron Bischoff, Jon Borchardt, David Everhart, Michal Guerquin, Paul Laskowski, Caleb Ouellette, and Michael Schmitz, for constructing the excellent \model{} demo.
    \item We thank other members of the PRIOR team, including Maximilian Argus, Jaemin Cho, Jiafei Duan, Rose Hendrix, Amita Kamath, Yejin Kim, Tanmay Gupta, Peter Sushko, Eli VanderBilt, and Piper Wolters, for providing advice and feedback on various aspects of \model{}. 
    \item We thank the Prolific team for their support and our annotators on Prolific for providing us with high-quality data that is crucial to \model{}. 
\end{itemize}

\noindent This material is based upon work supported by the National Science Foundation under Award No. 2413244.

\clearpage
\bibliographystyle{abbrvnat}
\bibliography{main,molmov1}

\clearpage

\appendix

\section*{Appendix}

The appendix includes the following sections:
\begin{itemize} 
\itemsep0em 
    \item \S\ref{supp:model} - Model details
    \item \S\ref{supp:training} - Training details
    \item \S\ref{appendix:eval} - Evaluation details
    \item \S\ref{appendix:add_eval} - Additional results
    \item \S\ref{appendix:test-time} - Test time scaling and SlowFast encoding
    \item \S\ref{appendix:data} - Data details
    \item \S\ref{appendix:qualitative_dataset} - Data examples
    \item \S\ref{appendix:limitation} - Limitations
    \item \S\ref{appendix:qualitative} - Qualitative results
\end{itemize}

\section{Model details}
\label{supp:model}
We present additional details about image encoding, hyperparameters, and implementation choices.

\paragraph{Image crops}.
Our method of encoding images largely follows Molmo~\cite{molmov1}, including the use of overlapping crops. Unlike Molmo, we do not pad crops with black. Instead, we resize them to 378 (even if that means changing the aspect ratio), following how SigLIP 2~\cite{siglip2} was trained. If the number of image patches is not evenly divisible by the pooling size, the bottom and far-right image patches are pooled with a reduced number of patches.

\paragraph{Video frames.}
We use torchcodec\footnote{\url{https://pytorch.org/blog/torchcodec/}} to extract frames from videos.
We extract frames at $S$ fps and the last frame. If that leads to more than $F$ frames, we instead extract frames uniformly, including the first and last frames.
For tracking, during training, we always sample videos at $S$ fps and trim both videos and point tracks to a maximum of $F$ frames instead. This ensures that points, which are annotated for $S$ fps, remain aligned with the sampled frames. 
We include the last frame since it is typically what is shown when the video ends and, therefore, can have special importance to users.
Frames are extracted based on timestamps (instead of frame indices) to handle variable fps videos. 

\paragraph{Formatting.}
Videos and image tokens are always inserted first, right after the BOS token. We insert different start and end special tokens for videos, tokens from a multi-crop image, and tokens for the low-resolution single-crop version of the image. Frames are interleaved with text timestamps written as seconds to one decimal point, and multi-images are interleaved with ``Image 1'', ``Image 2'', \etc, labels. Text is added after the image/video tokens following the Qwen3~\cite{qwen3technicalreport} prompt template without thinking tokens. 

\paragraph{Pointing.}
Our pointing format provides points in an HTML-like format, with the coordinates stored in a compact string. For each frame or image with points, the string contains an image index (for image input, starting at 1) or a frame timestamp (for video, shown in seconds with one decimal point), followed by a list of point coordinates. The points each have an \textit{object index}, which is unique for each distinct object being pointed at, and x and y coordinates that are normalized to be between 0 and 1000. Object indices are sequential, starting at 1. The object indices both facilitate counting, because the final object index represents the total count, and enable tracking by identifying repeating objects. Points are sorted by time/frame index and then by x and y coordinates. Values are space-separated, with semi-columns indicating a new frame/image. We elect to use this format over a format like JSON since it dramatically reduces the number of tokens needed to represent points.

An example output for a pointing and tracking task are shown below (new lines added for clarity):
\\ \\
\definecolor{object_index_color}{HTML}{D06DF4}
\definecolor{y_color}{HTML}{198621}
\definecolor{x_color}{HTML}{198621}
\colorlet{frame_id_color}{blue}
    \noindent\texttt{
<points coords="\textcolor{frame_id_color}{1} \textcolor{object_index_color}{1} \textcolor{x_color}{555} \textcolor{x_color}{169};\textcolor{frame_id_color}{2} \textcolor{object_index_color}{3} \textcolor{x_color}{649} \textcolor{x_color}{154} \textcolor{object_index_color}{4} \textcolor{x_color}{709} \textcolor{x_color}{162};\textcolor{frame_id_color}{5} \textcolor{object_index_color}{5} \textcolor{x_color}{758} \textcolor{x_color}{175} \textcolor{object_index_color}{6} \textcolor{x_color}{808} \textcolor{x_color}{183} \textcolor{object_index_color}{7} \textcolor{x_color}{852} \textcolor{x_color}{187}">\\ Inline text\\ </points>
}
    \\
    \\
    \texttt{<tracks coords="\textcolor{frame_id_color}{0.0} \textcolor{object_index_color}{1} \textcolor{x_color}{635} \textcolor{x_color}{522};\textcolor{frame_id_color}{0.5} \textcolor{object_index_color}{1} \textcolor{x_color}{606} \textcolor{x_color}{490} \textcolor{object_index_color}{2} \textcolor{x_color}{511} \textcolor{x_color}{124};\textcolor{frame_id_color}{1.0} \textcolor{object_index_color}{2} \textcolor{x_color}{515} \textcolor{x_color}{164};\textcolor{frame_id_color}{1.5} \textcolor{object_index_color}{2} \textcolor{x_color}{520} \textcolor{x_color}{168}">\\ Inline text\\ </tracks>
    }
\\ \\
\noindent Where image indices and frame timestamps are in \textcolor{frame_id_color}{blue}, object indices are in \textcolor{object_index_color}{purple}, and x and y coordinates are in \textcolor{x_color}{green}. The first example points to an object in images 1, 2, and 5. The second one tracks two different objects through several frames. The ``Inline text" is used to describe what is being pointed at.

\begin{table}
\newcommand\allmodels[1]{\multicolumn{3}{c}{\cellcolor{olive!5} #1}}
\scriptsize
\centering
 \begin{tabular}{c l c c c}
  & & 4B & 7B & 8B  \\
\midrule
\multirow{10}{*}{\rotatebox[origin=c]{90}{{\small Image Encoder}}} 
 & Params & \allmodels{380m} \\ 
 & Dim & \allmodels{1152} \\
 & MLP Dim & \allmodels{4304} \\
 & Act. & \allmodels{GELU} \\
 & Heads & \allmodels{16} \\
 & KV Heads & \allmodels{16} \\
 & Layers & \allmodels{27} \\
 & Image Size & \allmodels{384$\x$384} \\
 & Patch Size & \allmodels{14} \\
 & Dropout & \allmodels{0.0} \\
 \midrule
\multirow{7}{*}{\rotatebox[origin=c]{90}{{\small V/L Connector}}} 
 & Params & 57m & 80m & 88m \\
 & Image Pool Size & \allmodels{2$\x$2} \\
 & Video Pool Size & \allmodels{3$\x$3} \\
 & Pool Dim & \allmodels{1152} \\
 & Pool Heads & \allmodels{16} \\
 & MLP Dim & 9728 & 100352 & 12288 \\ 
 & Act. & \allmodels{SwiGLU} \\
 & Dropout & \allmodels{0.0} \\
 \midrule
\multirow{9}{*}{\rotatebox[origin=c]{90}{{\small LLM}}} 
 & Params & 4.0b & 7.3m & 8.2m \\
 & Embed & 151936 & 100352 & 151936 \\
 & Dim & 2560 & 4096 & 4096 \\
 & MLP Dim & 9728 & 11008 & 12288 \\
 & Act. & \allmodels{SwiGLU} \\
 & Heads & \allmodels{32} \\
 & KV Heads & 8 & 32 & 8\\ 
 & Layers & 36 & 32 & 36 \\
 & Theta & 1m & 0.5m & 1m \\
 & Dropout & \allmodels{0.1} \\
 \midrule
\multirow{12}{*}{\rotatebox[origin=c]{90}{{\small Pre-Train}}} & Warmup ViT & \allmodels{2000}\\
 & Warmup Con. & \allmodels{200}\\
 & Warmup LLM & \allmodels{2000}\\
 & LR ViT & \allmodels{6e-6}\\
 & LR Con. & \allmodels{2e-4}\\
 & LR LLM & \allmodels{2e-4} \\
 & Cosine Decay & \allmodels{10\%} \\
 & Eps. & \allmodels{1e-6} \\ 
 & Betas & \allmodels{0.9, 0.95} \\ 
 & Batch Size & \allmodels{128} \\ 
 & Sequence Length & \allmodels{2560} \\ 
 & Steps & \allmodels{32k} \\ 
 \midrule
 \multirow{12}{*}{\rotatebox[origin=c]{90}{{\small SFT}}} & Warmup ViT & \allmodels{200}\\
 & Warmup Con. & \allmodels{200}\\
 & Warmup LLM & \allmodels{200}\\
 & LR ViT &  \allmodels{5e-6} \\
 & LR Con. &  \allmodels{5e-6} \\
 & LR LLM &  \allmodels{1e-5} \\
 & Cosine Decay & \allmodels{10\%} \\
 & Eps. & \allmodels{1e-6} \\ 
 & Betas & \allmodels{0.9, 0.95} \\ 
 & Batch Size & \allmodels{128} \\ 
 & Sequence Length & \allmodels{16384} \\ 
 & Steps & \allmodels{30k} \\ 
\end{tabular}
    \caption{\textbf{Model and training hyper-parameters}, \model{}-O-7B is a version of \model{} with OLMo 3~\cite{olmo3}. Long-context post-training used the same parameters as SFT}
    \label{tab:hyperparameters}
\end{table}
\paragraph{Hyperparameters.}
Hyperparameters for the \model{} models are shown in Table~\ref{tab:hyperparameters}. The connector MLP uses the same intermediate dimension as the LLM, so its size depends on the LLM; otherwise, they are the same across all models. All models use the SigLIP 2 So400m/14 384px ViT~\cite{siglip2}. 

\paragraph{Implementation.}
Our implementation uses PyTorch with Fully Sharded Data Parallel (FSDP) 2~\cite{zhao2023pytorch}.
We use  PyTorch's Scaled Dot Product Attention (SDPA), not FlashAttention~\cite{dao2022flashattention,dao2024flashattention2}, since it does not support custom attention masks. We use \texttt{torch.compile} to improve throughput and ensure that the shapes in the LLM and ViT are static so the model can be statically compiled, which we find essential for maximizing throughput.

To improve throughput, we also utilize PyTorch's Automatic Mixed Precision (AMP) module\footnote{\url{https://pytorch.org/docs/stable/report/amp.html}}, which enables most operations to run in half-precision with bfloat16 numbers. Computations for layer normalization~\cite{ba2016layer} and Rotary Position Embedding (RoPE)~\cite{su2024roformer} are still carried out in full precision.

When computing gradients, each GPU computes a gradient on a small mini-batch of examples, after which the gradients are averaged across all devices. We always compute the per-device gradient by dividing the total loss on that device by the \textit{average} number of loss tokens across all devices, not the number of loss tokens on that particular device. This avoids a subtle bias that effectively up-weights examples with a small number of loss tokens (\eg, with short responses)\footnote{\url{https://unsloth.ai/blog/gradient}}~\cite{hermans2017accumulated}.

During fine-tuning, mixing is done within each batch so that the batches contain examples from a variety of datasets. We truncate examples that are longer than the max sequence length. This occurs in $<0.1\%$ of cases, usually due to videos with both subtitles and a large number of annotations. We find training to be stable, without loss spikes or NaNs.

\section{Training details}
In this section, we provide additional details about packing, the data mixture, and other components of how \model was trained.

\begin{table*}[]
\definecolor{baselinecolor}{gray}{.9}
\newcommand{\category}[1]{\cellcolor{baselinecolor}{#1}}
\renewcommand{\indent}{\hspace{0.2cm}}
\newcommand{\header}{name & rate & visual & anno. & ex. \\ \midrule}
\setlength{\tabcolsep}{2pt}
\vspace{-0.3cm}
\scriptsize
    \centering
    \begin{minipage}{0.49\linewidth}
\centering
    \begin{tabular}{l c c c c}
\header
\category{\textsc{Image QA}} & \category{22.7} & \category{2.7m} & \category{32m} & \category{2.4m} \\
\indent PixMo-Clocks & 1.9 & 800k & 800k & 800k \\
\indent Llava-665k-Multi & 1.5 & 280k & 2.5m & 160k \\
\indent TallyQA & 1.4 & 130k & 250k & 130k \\
\indent CoSyn-chart & 1.3 & 120k & 1.1m & 120k \\
\indent NLVR2 & 1.1 & 100k & 86k & 86k \\
\indent VQA v2 & 1.1 & 83k & 440k & 83k \\
\indent CoSyn-doc & 1.0 & 71k & 610k & 71k \\
\indent A-OKVQA & 1.0 & 33k & 34k & 34k \\
\indent CoSyn-math & 1.0 & 67k & 67k & 67k \\
\indent CoSyn-table & 0.8 & 47k & 420k & 47k \\
\indent DocVQA & 0.7 & 10k & 39k & 39k \\
\indent CoSyn-diagram & 0.7 & 35k & 300k & 35k \\
\indent TextQA & 0.7 & 22k & 35k & 35k \\
\indent \textcolor{molmocolor}{Molmo2-SynMultiImageQA-chart} & 0.7 & 100k & 330k & 33k \\
\indent ChartQA & 0.6 & 18k & 28k & 28k \\
\indent \textcolor{molmocolor}{Molmo2-SynMultiImageQA-doc} & 0.6 & 88k & 270k & 28k \\
\indent ST-VQA & 0.6 & 18k & 25k & 25k \\
\indent InfographicVQA & 0.6 & 4.4k & 24k & 24k \\
\indent TabWMP & 0.6 & 23k & 23k & 23k \\
\indent PlotQA & 0.5 & 160k & 20m & 160k \\
\indent AI2D & 0.5 & 6.2k & 15k & 15k \\
\indent \textcolor{molmocolor}{Molmo2-SynMultiImageQA-diagram} & 0.5 & 45k & 150k & 15k \\
\indent \textcolor{molmocolor}{Molmo2-SynMultiImageQA-table} & 0.4 & 47k & 140k & 14k \\
\indent CoSyn-music & 0.4 & 12k & 82k & 12k \\
\indent DVQA & 0.4 & 200k & 2.3m & 200k \\
\indent FigureQA & 0.4 & 100k & 1.3m & 100k \\
\indent OK-VQA & 0.4 & 9k & 9k & 9k \\
\indent CoSyn-chemical & 0.4 & 8.9k & 55k & 8.9k \\
\indent Spot-the-Difference & 0.3 & 15k & 14k & 7.5k \\
\indent ScienceQA & 0.3 & 6.2k & 6.2k & 6.2k \\
\indent \textcolor{molmocolor}{Molmo2-SynMultiImageQA-music} & 0.3 & 12k & 46k & 4.7k \\
\indent \textcolor{molmocolor}{Molmo2-SynMultiImageQA-chemical} & 0.2 & 8k & 23k & 2.4k \\
\\
\category{\textsc{Image Pointing}} & \category{9.1} & \category{510k} & \category{5.5m} & \category{1.1m} \\
\indent PixMo-Points & 4.6 & 220k & 4.6m & 530k \\
\indent \textcolor{molmocolor}{Molmo2-MultiImagePoint} & 2.0 & 180k & 470k & 470k \\
\indent PixMo-Count & 1.2 & 37k & 74k & 74k \\
\indent CoSyn-point & 1.2 & 68k & 320k & 68k \\
\\
\category{\textsc{Captions/Long QA}} & \category{13.6} & \category{1.2m} & \category{1.6m} & \category{1.2m} \\
\indent \textcolor{molmocolor}{Molmo2-Cap} & 3.4 & 100k & 280k & 100k \\
\indent PixMo-CapQa & 3.1 & 190k & 270k & 190k \\
\indent PixMo-Cap & 2.3 & 710k & 710k & 710k \\
\indent PixMo-AskModelAnything & 1.9 & 71k & 160k & 71k \\
\indent \textcolor{molmocolor}{Molmo2-MultiImageQA} & 1.5 & 98k & 73k & 45k \\
\indent \textcolor{molmocolor}{Molmo2-AskModelAnything} & 1.5 & 43k & 130k & 43k \\
\\
\category{\textsc{NLP}} & \category{9.1} & \category{0} & \category{980k} & \category{980k} \\
\indent Tulu & 9.1 & 0 & 980k & 980k \\
\\
\category{\textsc{Video Pointing}} & \category{13.6} & \category{260k} & \category{500k} & \category{370k} \\
\indent \textcolor{molmocolor}{Molmo2-VideoPoint} & 10.9 & 250k & 450k & 330k \\
\indent AcademicVideoPoint-MeViS & 1.2 & 1.6k & 20k & 20k \\
\indent AcademicVideoPoint-ReVOS & 0.7 & 3.4k & 11k & 11k \\
\indent AcademicVideoPoint-LV-VIS & 0.7 & 3.1k & 11k & 11k \\
\indent AcademicVideoPoint-OVIS & 0.05 & 600 & 880 & 880 \\
\indent AcademicVideoPoint-BURST & 0.04 & 310 & 680 & 680 \\
\indent AcademicVideoPoint-Ref-DAVIS17 & 0.03 & 58 & 450 & 450 \\
\end{tabular}

\end{minipage} 
\hfill 
\begin{minipage}{0.49\linewidth}
\begin{tabular}{l c c c c}
\centering
\header
\category{\textsc{Video QA}} & \category{18.2} & \category{2.3m} & \category{4.7m} & \category{2.4m} \\
\indent \textcolor{molmocolor}{Molmo2-CapQA} & 1.6 & 190k & 950k & 190k \\
\indent \textcolor{molmocolor}{Molmo2-SubtitleQA} & 1.2 & 100k & 470k & 100k \\
\indent Video Localized Narratives & 1.1 & 53k & 180k & 56k \\
\indent TGIF & 0.9 & 63k & 210k & 63k \\
\indent TVQA & 0.9 & 120k & 120k & 120k \\
\indent Paxion & 0.9 & 440k & 440k & 440k \\
\indent Moments In Time & 0.9 & 710k & 710k & 710k \\
\indent Kinentics & 0.9 & 420k & 420k & 420k \\
\indent LLaVA Academic & 0.9 & 11k & 62k & 31k \\
\indent Ego4D & 0.9 & 53k & 53k & 53k \\
\indent EPIC KITCHENS  & 0.7 & 37k & 37k & 37k \\
\indent COIN & 0.7 & 7.8k & 30k & 30k \\
\indent How2QA & 0.6 & 25k & 35k & 25k \\
\indent ActivityNet & 0.5 & 12k & 46k & 21k \\
\indent FunQA & 0.5 & 3.1k & 200k & 21k \\
\indent CLEVRER & 0.5 & 10k & 130k & 20k \\
\indent STAR & 0.5 & 3k & 91k & 19k \\
\indent YouCook2 & 0.4 & 1.2k & 18k & 10k \\
\indent SUTD-TrafficQA & 0.4 & 10k & 56k & 10k \\
\indent CinePile & 0.4 & 9.2k & 300k & 9.2k \\
\indent Charades STA & 0.4 & 5.3k & 12k & 9.2k \\
\indent QVHighlights & 0.3 & 6.8k & 7k & 7k \\
\indent MotionBench & 0.3 & 5k & 5k & 5k \\
\indent Countix & 0.2 & 3.9k & 4.4k & 4.4k \\
\indent NExT-QA & 0.2 & 3.9k & 34k & 3.9k \\
\indent Sports-QA & 0.2 & 3.6k & 56k & 3.6k \\
\indent IntentQA & 0.2 & 3.2k & 24k & 3.2k \\
\indent NewsVideoQA & 0.2 & 2.9k & 8.4k & 2.9k \\
\indent RoadTextVQA & 0.2 & 2.6k & 8.4k & 2.6k \\
\indent PerceptionTest & 0.2 & 2k & 7.4k & 2k \\
\indent CamaeraBench & 0.1 & 1.4k & 1.4k & 1.4k \\
\indent Social IQ 2 & 0.1 & 0.79k & 5k & 0.79k \\
\\
\category{\textsc{Video Tracking}} & \category{13.6} & \category{130k} & \category{800k} & \category{800k} \\
\indent \textcolor{molmocolor}{Molmo2-VideoTrack} & 4.6 & 8k & 220k & 220k \\
\indent AcademicVideoTrack-MeViS & 2.0 & 1.7k & 150k & 150k \\
\indent AcademicVideoTrack-ViCaS & 1.2 & 15k & 130k & 130k \\
\indent AcademicVideoTrack-ReVOS & 1.2 & 0.7k & 82k & 82k \\
\indent AcademicVideoTrack-TrackingNet & 1.1 & 29k & 29k & 29k \\
\indent AcademicVideoTrack-Ref-Youtube-VOS & 0.9 & 3.5k & 26k & 26k \\
\indent AcademicVideoTrack-VastTrack & 0.8 & 46k & 93k & 93k \\
\indent AcademicVideoTrack-LV-VIS & 0.8 & 3.1k & 38k & 38k \\
\indent AcademicVideoTrack-GOT-10k & 0.4 & 9.2k & 18k & 18k \\
\indent AcademicVideoTrack-WebUAV & 0.2 & 3.2k & 6.3k & 6.3k \\
\indent AcademicVideoTrack-BURST & 0.07 & 0.28k & 2.9k & 2.9k \\
\indent AcademicVideoTrack-LaSOT & 0.06 & 1.1k & 2.2k & 2.2k \\
\indent AcademicVideoTrack-TNL2K & 0.06 & 0.88k & 1.8k & 1.8k \\
\indent AcademicVideoTrack-WebUOT & 0.05 & 0.84k & 1.5k & 1.5k \\
\indent AcademicVideoTrack-LVOS V2 & 0.05 & 0.42k & 1.2k & 1.2k \\
\indent AcademicVideoTrack-lasot & 0.03 & 0.22k & 0.45k & 0.45k \\
\indent AcademicVideoTrack-UW-COT220 & 0.03 & 0.21k & 0.4k & 0.4k \\
\indent AcademicVideoTrack-LVOS V1 & 0.02 & 0.12k & 0.3k & 0.3k \\
\indent AcademicVideoTrack-TNLLT & 0.02 & 0.15k & 0.29k & 0.29k \\
\indent AcademicVideoTrack-Ref-DAVIS17 & 0.02 & 0.06k & 1.1k & 1.1k \\
\indent AcademicVideoTrack-YouTube-VIS & 0.02 & 1.2k & 1.4k & 1.4k \\
\indent AcademicVideoTrack-MoCA-Video & 0.01 & 0.13k & 0.4k & 0.4k \\
\\

\end{tabular}
    \end{minipage}
    \caption{\textbf{Full dataset list}. Columns show sampling rates, the number of videos or images, the number of annotations, and the number of training examples built after formatting the data into message trees.}
    \label{tab:detailed_datasets}
\end{table*}
\label{supp:training}

\paragraph{Packing.}
Our packing algorithm keeps a pool of $M = 48$ examples that have already been preprocessed and converted into a tokenized representation. If the pool is not full, examples are drawn from the training mixture and added to the pool. When the pool is full, we run a dynamic programming solver to find the optimal subset of examples that maximizes $T + I * w_i$ subject to $T \le 16384$ and $I \le 128$, where $T$ is the total number of text tokens in the selected subset, $I$ is the total number of crops, and $w_i = 30$ is a hyperparameter.
During long context training, we instead use a max of 384 images and 36864 tokens.
The selected examples are yielded as a single packed sequence and removed from the pool. In practice, we run the solver on a quantized version of the problem by rounding the number of tokens to the nearest multiple of 32. 

Increasing $M$ quickly leads to diminishing returns in terms of packing efficiency. We do not observe any gains from using more than 48. The algorithm is usually robust to $w_i$, but we observe that in some settings, if $w_i$ is too low, the pool can become filled with examples with 128 crops, which usually cannot be packed with anything else, thereby reducing efficiency.

Implementation-wise, we add this logic into torch's \textit{DataLoader} so that each data-worker runs this algorithm independently. This makes the algorithm easy to use, but it does add some unnecessary overhead when there are many data workers. This could be addressed in future work through a deeper integration into torch's data-loading logic. In practice, we find that packing still does not slow down the training speed. Loading and extracting frames from videos remains, by far, the most costly part of data loading.

\paragraph{Pre-training.}
During pre-training, we use response-only dropout, \ie, residual dropout on just the output tokens, of 0.1, length conditioning, and both the caption and transcript, following Molmo~\cite{molmov1}.

\paragraph{SFT.}
\begin{figure}
    \centering
    \includegraphics[width=.8\linewidth]{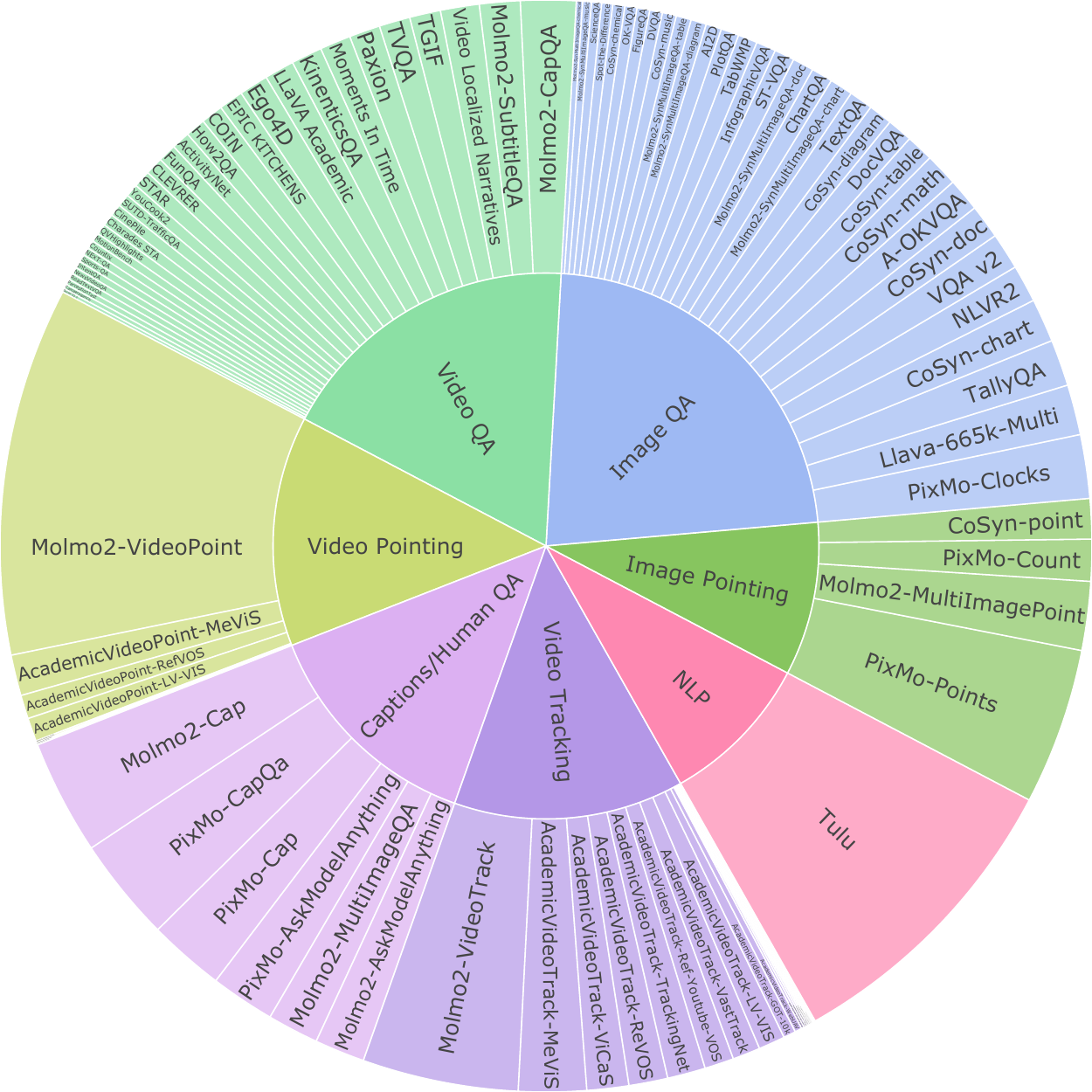}
    \caption{\textbf{Molmo2 SFT mixture.} Categories and datasets are shown in proportion to sampling rates in SFT mixture.}
    \label{fig:sft_mixture}
\end{figure}
The full list of datasets in our SFT mixture is shown in Table~\ref{tab:detailed_datasets}, and visualized in Figure~\ref{fig:sft_mixture}.
During SFT we use regular residual dropout of 0.1. 

\paragraph{Prompting.}
We use the human-written questions with long-form answers from PixMo-AskModelAnything, PixMo-CapQA, and \model{}-AskModelAnything directly. For captioning, all multiple-choice questions, and our various grounding tasks, we use prompt templates to generate a variety of ways to prompt the model for the target output. 
The remaining short-answer or captioning academic datasets typically have answer styles that are poorly suited for user-facing behaviors, either because they are too terse or have other idiosyncratic quirks due to how the data was collected.
For these datasets, we prompt the model with style tags (\eg "short\_video\_answer:") so that \model{} adopts those answer styles only if specifically prompted to do so.

\paragraph{Hyperparameters.}
Hyperparameters for AdamW~\cite{kingma2015adam} are in Table~\ref{tab:hyperparameters}. 
Following Molmo~\cite{molmoact2025}, during pre-training, we use a high learning rate for the connector and a long warmup for the ViT and LLM so that the first steps of training mostly train the connector. We use a cosine learning rate that decays to 10\% of the peak learning rate. We do not use weight decay.

\paragraph{Training time.}
\begin{table}[]
\small
    \centering
    \begin{tabular}{@{}l c c c c c c c c c c}
         \multirow{2}{*}{\textbf{Model}} & \multicolumn{3}{c}{Pre-train} & \multicolumn{3}{c}{SFT} & \multicolumn{3}{c}{Long-Context} \\
         & GPUs & time & GPU hr. & GPUs & time & GPU hr. & GPUs & time & GPU hr. \\ \hline
         4B & 32 & 15.2 & 490 & 128 & 58.8 & 7.5k & 128 & 25.3 & 3.2k \\
         7B & 64 & 11.3 & 720 & 128 & 59.3 & 7.6k & 128 & 25.7 & 3.3k \\
         8B & 64 & 12.1 & 780 & 128 & 63.0 & 8.1k & 128 & 26.0 & 3.3k \\ 
    \end{tabular}
    \caption{\textbf{Training times}. Training was done with Nvidia H100 GPUs.}
    \label{tab:gpu_hours}
\end{table}
We show the time and compute used for training \model{} in Table~\ref{tab:gpu_hours}.
During SFT, a high portion of the computation is from the ViT because, for videos, 9 patches in the ViT are processed for each visual token in the LLM. As a result, increasing the LLM size has a reduced effect on the training time.

\paragraph{Specialized models.}
Specialized models are pre-trained and then undergo a shorter SFT training round with a subset of our SFT data. 

For the  QA-specialized model, we start with an earlier version of the pre-trained \model-4B checkpoint and perform SFT on video caption and video QA data, excluding image, NLP, and video pointing/tracking datasets. We only train the model for 6k steps. For the captioning-specialized model, we only use the \model-Cap dataset and train the model for 5k steps. For the pointing-specialized model, we use a three-stage training pipeline in which the model is first pre-trained on image captioning for 22k steps, then further trained for 26k steps on the Molmo2 SFT mixture excluding video pointing and tracking data, and finally finetuned for 6k steps solely on video pointing data. For the tracking-specialized model, we use the same three-stage pipeline except that we finetune the model on video pointing and tracking data for 10k steps in the final stage. Finally, the image-specialized model is trained for 24k steps and a sequence length of 2560 on just the NLP, image pointing, image academic, and image datasets from the Captions/Long QA dataset groups, starting from \model{}-4B pre-trained checkpoint. We do not do long-context post-training for any specialized models.

\section{Evaluation Details}
Next, we provide more details about our evaluation setup.
\label{appendix:eval}

\paragraph{Captioning.} 
We evaluate video captioning quality on a set of 693 diverse videos using an F1 score designed to evaluate how accurate and detailed the captions are, similar to Molmo~\cite{molmov1}. We selected a small number of videos across diverse categories from creative-commons licensed Vimeo\footnote{\url{https://vimeo.com/creativecommons/cc0}} to ensure that the videos are disjoint from our training set, which is mostly composed of YouTube videos. 
The human captions of this evaluation set are collected using a protocol similar to Molmo2-Cap, but with annotators who were manually selected because they provided high-quality captions when collecting Molmo2-Cap. 
Each evaluation video has up to five human captions. For every model-generated caption and the human caption set, we first prompt GPT-4.1 to enumerate all distinct atomic statements. 
Precision is computed as the percentage of statements from the model-generated caption that were also stated in the human captions, using GPT-4.1 as a judge. Recall is computed through the opposite process, by matching statements from human captions to the model-generated captions. We average precision and recall across all videos and compute their harmonic mean to obtain our final summary metric: video caption F1.

We prompt \model{} and baseline models by asking for a long, detailed caption of the input video.

\begin{table}[!ht]
    \renewcommand{\arraystretch}{0.98}
    \centering
    \begin{tabular}{@{}lrrrrrr@{}}
 & \multicolumn{2}{@{}c}{Overall} 
 & \multicolumn{2}{@{}c}{Captioning} 
 & \multicolumn{2}{@{}c}{QA} \\ 
\textbf{Model}  & Score & Rank & Score & Rank & Score & Rank\\
\midrule
\multicolumn{7}{@{}l}{\textbf{\textit{API call only}}} \\

GPT-5~\cite{gpt5}                    
& 1031 & 10  
& 1136 & 2  
& 1019 & 11 \\

GPT-5 mini~\cite{gpt5}               
& 1076 & 4  
& 1086 & 5  
& 1075 & 4 \\

Gemini 3 Pro~\cite{gemini3}    
& 1082 & 3
& 1126 & 3
& 1076 & 3 \\

Gemini 2.5 Pro~\cite{comanici2025gemini}    
& 1096 & 1
& 1148 & 1
& 1090 & 1 \\

Gemini 2.5 Flash~\cite{comanici2025gemini}   
& 1084 & 2  
& 1109 & 4  
& 1082 & 2 \\

Claude Sonnet 4.5~\cite{anthropic2025sonnet}
& 1008 & 12 
& 1009 & 10  
& 1008 & 12 \\

\midrule
\multicolumn{7}{@{}l}{\textbf{\textit{Open weights only}}} \\

InternVL3.5-4B~\cite{wang2025internvl3} 
& 935  & 19 
& 817  & 19 
& 947  & 19 \\

InternVL3.5-8B~\cite{wang2025internvl3} 
& 941  & 18 
& 855  & 18 
& 951  & 17 \\

Qwen3-VL-4B~\cite{bai2025qwen3vltechnicalreport} 
& 1048 & 7  
& 1052 & 7  
& 1049 & 6 \\

Qwen3-VL-8B~\cite{bai2025qwen3vltechnicalreport} 
& 1054 & 6 
& 1105 & 5 
& 1048 & 7 \\

Keye-VL-1.5-8B~\cite{yang2025kwai}      
& 952  & 17 
& 957  & 15 
& 950  & 18 \\

GLM-4.1V-9B~\cite{glmv}                 
& 962  & 14 
& 1013 & 9  
& 956  & 15 \\

MiniCPM-V-4.5-8B~\cite{yu2025minicpmv45cookingefficient}  
& 975  & 13 
& 978  & 14 
& 975  & 13 \\

Eagle2.5-8B~\cite{eagle2_5}            
& 1019 & 11  
& 987  & 13 
& 1022 & 10 \\

\midrule
\multicolumn{7}{@{}l}{\textbf{\textit{Open models}}} \\

PLM-3B~\cite{cho2025PerceptionLM}    
& 841  & 21 
& 880  & 17 
& 836  & 21 \\

PLM-8B~\cite{cho2025PerceptionLM}    
& 853  & 20 
& 761  & 21 
& 863  & 20 \\

LLaVA-Video-7B~\cite{llava_video}    
& 959  & 15 
& 981  & 14 
& 955  & 16 \\

VideoChat-Flash-7B~\cite{li2024videochat} 
& 956  & 16 
& 932  & 16 
& 959  & 14 \\

\midrule
\multicolumn{7}{@{}l}{\textbf{\textit{Molmo2 family: Open weights, Open data, Open code}}} \\

\textcolor{molmocolor}{\model{}-4B}    
& 1041 & 8  
& 1004 & 11  
& 1045 & 8 \\

\textcolor{molmocolor}{\model{}-8B}    
& 1057 & 5 
& 1049 & 8  
& 1059 & 5 \\

\textcolor{molmocolor}{\model{}-O-7B}    
& 1033 & 9
& 1019 & 9
& 1034 & 9 \\

\end{tabular}
\caption{\textbf{Human evaluation results.} Scores updated using bootstrap Elo medians from overall, captioning, and QA evaluations.}
\label{tab:human_eval_results}
\end{table}

\begin{figure*}[!t]
    \centering
    \includegraphics[width=\textwidth]{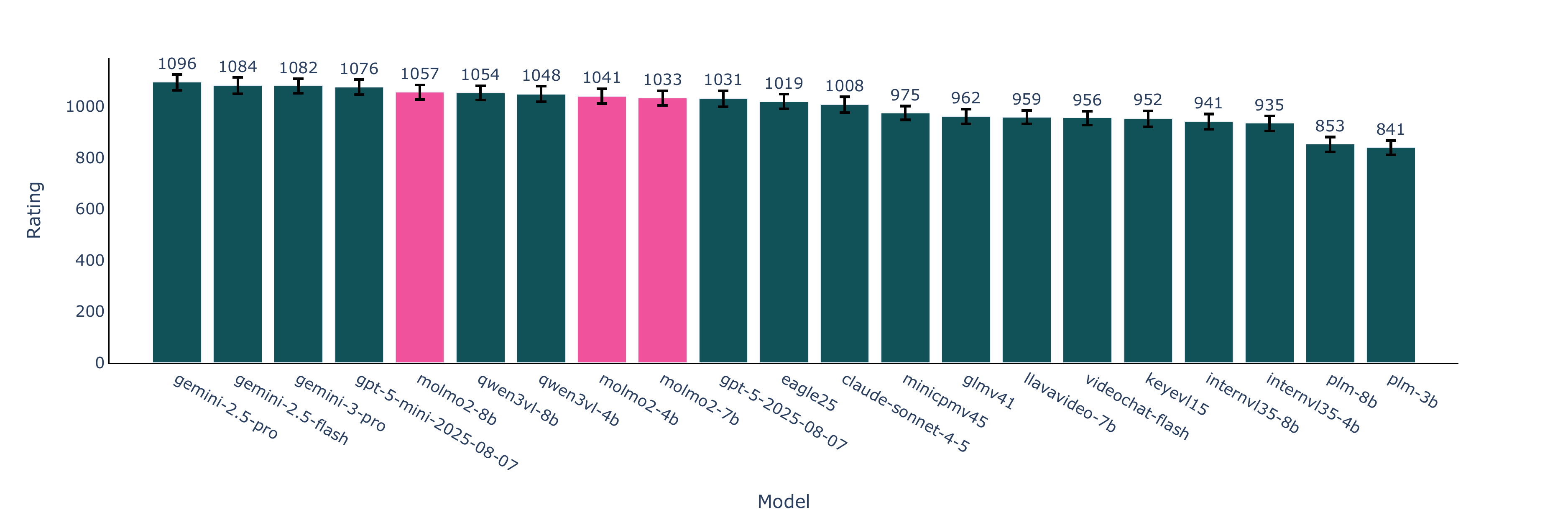}
    \caption{Elo ratings with confidence intervals}
    \label{fig:elo_ratings}
\end{figure*}

\begin{figure}[!t]
    \centering
    \includegraphics[width=.8\columnwidth]{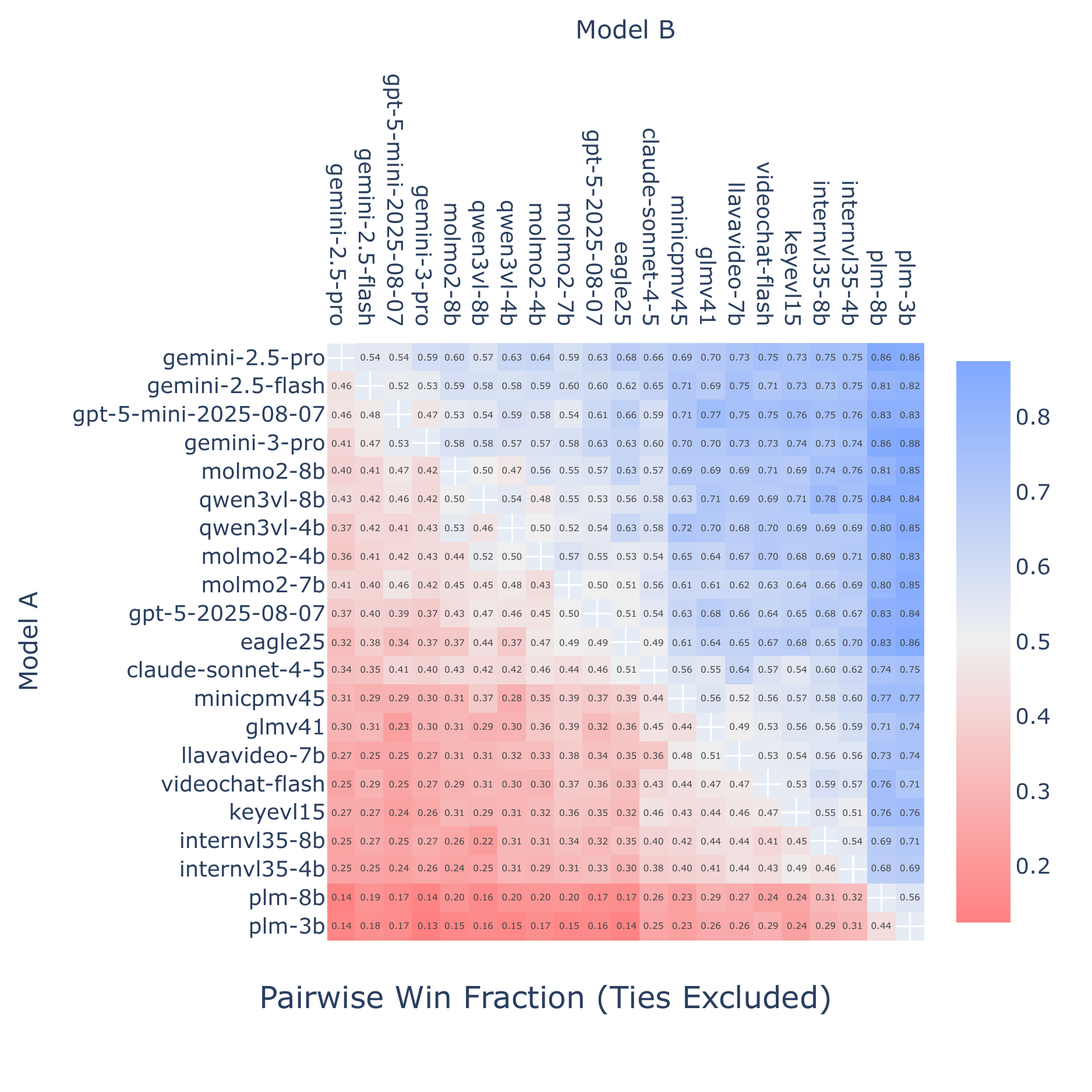}
    \caption{Pairwise win rates across all model pairs in human preference evaluation.}
    \label{fig:win_rates}
\end{figure}

\paragraph{Human Eval.}
Following the best practices from~\cite{chiang2024chatbot}, we use bootstrapping with 1000 rounds to get a more stable version of Elo ratings and estimate confidence intervals. We plot the Elo scores with confidence intervals in Figure~\ref{fig:elo_ratings}.

To better understand the results from human preference evaluation, we also analyze (1) fine-grained task-specific Elo ratings for diagnostic purposes~\cite{lirethinking} (Table~\ref{tab:human_eval_results}), (2) deterministic pairwise win rates (Figure~\ref{fig:win_rates}); and (3) human explanations of their preference. From the task-specific results, we learn that \model{} performs better than Qwen3-VL on the open-ended QA task, ranking first among open models. However, it underperforms Qwen3-VL and GLM-4.1V on captioning. Furthermore, we also examine the pairwise win rates across all model pairs, which are deterministic. We note that \model-8B's win rate against Qwen3-VL-8B is 53\%, and \model-4B's win rate against Qwen3-VL-4B is 51\%, suggesting that \model{} family of models is competitive against Qwen3-VL models. Lastly, from a qualitative analysis of human annotators' explanations of their preferences, we learn that our model performs well on QA because it provides a detailed explanation to its answer when needed and a concise one otherwise, However \model{} falls short on captioning because it sometimes outputs repetitive or non-sensical content at the end of the caption, which we believe is due to text-repetition issues when generating extremely long output (see Section~\ref{appendix:limitation}).

\paragraph{Counting and Pointing.}
For the video counting evaluation, we preprocess 2 fps videos and clip them to random intervals under 63 seconds. In addition to exact accuracy and close accuracy, we also track models' counting accuracy by query category (Table~\ref{tab:count_results_by_cat}) and by object count (Table~\ref{tab:count_results_by_count}). We find that \model-8B performs the best on Action/Event and Object counting, just behind Gemini 2.5 Pro and GPT-5. \model-8B also performs competitively on Animal counting, trailing slightly behind GPT-5 and Qwen3-VL-8B. Importantly, \model{} achieves similar accuracies to Qwen3-VL on low-count (0-10) queries while performing substantially better on high-count cases (10-60). Notably, Qwen3-VL obtains 0\% accuracy in the 25-60 range, whereas \model{} exceeds 10\%, placing it just behind Gemini 2.5 Pro.

For the video pointing evaluation, we use 2 fps videos with a maximum of 384 frames along with ground truth points and masks at 2 fps. For metrics, we compute recall, precision, F1, and valid accuracy (\ie, the percentage of predictions that are parsed correctly), reporting all metrics in Table~\ref{tab:video_count_and_point_results}. In contrast to the counting task, Qwen3-VL struggles to perform meaningful pointing: Qwen3-VL-8B achieves only 1.5 F1, indicating that it rarely produces correct points. Even the strongest proprietary model shows a significant gap relative to ours: Gemini 3 and 2.5 Pro reach 20.0 and 13.0 F1, whereas \model{}-4B and \model{}-8B achieve 39.9 and 38.4 F1, respectively. This highlights a substantial performance advantage of \model{} on fine-grained spatio-temporal localization.

To evaluate the performance of baseline models on counting and pointing, we adopt the following setups. For both counting and pointing, we feed the entire videos to Gemini and Qwen3-VL models and use their default setup for video preprocessing. For GPT and Claude models, we feed the video frames to them using the same max frames and fps in our models' video preprocessing. As for the prompt, we use a general counting prompt followed by a brief format instruction across all models: ``How many \{label\} are there? Output the integer number of the count only. The answer is:''. For pointing, we first try prompting baseline models with our pointing format, but find that they struggle to follow the instruction and produce sensible outputs. We then carefully review various cookbooks for the baseline models where available, and design prompts with the HH:MM:SS format for timestamps and the bounding box format (which we then calculate the center's coordinates and use those for evaluation). We present the prompts used in video pointing evaluation for models with video and image inputs in prompt~\ref{lst:video-prompt} and ~\ref{lst:image-prompt}, respectively.
\begin{table}[!t]
\renewcommand{\arraystretch}{0.98}
\centering
\small
\begin{tabular}{@{}lcccc@{}}
 & \multicolumn{4}{c}{\textbf{Query Catogery} } \\

\textbf{Model} &
Action/Event & Animal & Object & Avg. \\

\midrule
\multicolumn{5}{@{}l}{\textbf{\textit{API call only}}} \\

GPT-5~\cite{gpt5} & 
46.6 & 75.5 & 29.8 & 50.6 \\

GPT-5 mini~\cite{gpt5} &
36.2 & 63.3 & 25.1 & 41.5 \\

Gemini 3 Pro~\cite{gemini3} &
58.6 & 75.5 & 29.7 & 54.6 \\

Gemini 2.5 Pro~\cite{comanici2025gemini} &
53.4 & 63.3 & 30.0 & 48.9 \\

Gemini 2.5 Flash~\cite{comanici2025gemini} &
36.2 & 63.3 & 27.7 & 42.4 \\

Claude Sonnet 4.5~\cite{anthropic2025sonnet} &
26.3 & 53.1 & 24.3 & 34.6 \\

\midrule
\multicolumn{5}{@{}l}{\textbf{\textit{Open weights only}}} \\

Qwen3-VL-4B~\cite{bai2025qwen3vltechnicalreport} &
39.7 & 59.2 & 19.5 & 39.4 \\

Qwen3-VL-8B~\cite{bai2025qwen3vltechnicalreport} &
43.1 & \textbf{75.5} & 22.5 & \textbf{47.0} \\

\midrule
\multicolumn{5}{@{}l}{\textbf{\textit{Molmo2 family: Open weights, Open data, Open code}}} \\


\textcolor{molmocolor}{\model-4B} &
\textbf{51.7} & 59.2 & 29.1 & 46.7 \\

\textcolor{molmocolor}{\model-8B} &
50.0 & 69.4 & 29.6 & \textbf{49.7} \\
\textcolor{molmocolor}{\model-O-7B} &
50.0 & 63.3 & \textbf{27.5} & 46.9 
\end{tabular}

\caption{\textbf{Molmo2-VideoCount} accuracy by query category.}
\label{tab:count_results_by_cat}
\end{table}

\begin{table}[!t]
\renewcommand{\arraystretch}{0.98}
\centering
\small
\begin{tabular}{@{}lccccccc@{}}
 & \multicolumn{7}{c}{\textbf{Object Count}} \\
\textbf{Model} &
0--5 & 5--10 & 10--15 & 15--20 & 20--25 & 25--60 & Avg. \\

\midrule
\multicolumn{8}{@{}l}{\textbf{\textit{API call only}}} \\

GPT-5~\cite{gpt5} & 
64.4 & 34.1 & 31.3 & 16.2 & 11.1 & 10.5 & 27.9 \\

GPT-5 mini~\cite{gpt5} &
55.7 & 28.2 & 25.0 & 10.8 & 6.3 & 10.5 & 22.8 \\

Gemini 3 Pro~\cite{gemini3} &
69.5 & 34.1 & 24.1 & 16.2 & 14.3 & 12.5 & 28.5\\

Gemini 2.5 Pro~\cite{comanici2025gemini} &
61.5 & 31.3 & 31.5 & 15.7 & 17.5 & 13.0 & 28.4 \\

Gemini 2.5 Flash~\cite{comanici2025gemini} &
56.9 & 31.0 & 27.5 & 19.2 & 9.8 & 3.5 & 24.6 \\

Claude Sonnet 4.5~\cite{anthropic2025sonnet} &
48.0 & 24.7 & 20.3 & 14.9 & 15.9 & 5.4 & 21.5 \\

\midrule
\multicolumn{8}{@{}l}{\textbf{\textit{Open weights only}}} \\

Qwen3-VL-4B~\cite{bai2025qwen3vltechnicalreport} &
56.9 & 17.6 & 21.3 & 2.7 & 3.2 & 0.0 & 16.9 \\

Qwen3-VL-8B~\cite{bai2025qwen3vltechnicalreport} &
63.8 & 30.6 & 15.0 & 6.8 & 6.3 & 0.0 & 20.4 \\

\midrule
\multicolumn{8}{@{}l}{\textbf{\textit{Molmo2 family: Open weights, Open data, Open code}}} \\



\textcolor{molmocolor}{\model-4B} &
58.0 & 31.8 & 30.0 & 24.3 & \textbf{9.5} & \textbf{12.3} & \textbf{27.7} \\

\textcolor{molmocolor}{\model-8B} &
\textbf{64.4}	 & \textbf{32.9}	 & 26.3	& \textbf{25.7}	& 7.9	& 7.0 & 27.4 \\

\textcolor{molmocolor}{\model-O-7B} &
60.9	 & 32.9	 & \textbf{27.5}	 & 16.2	 & 6.3	 & 8.8	 & 25.4

\end{tabular}
\caption{\textbf{Molmo2-VideoCount} accuracy by object count. }
\label{tab:count_results_by_count}
\end{table}











\begin{lstlisting}[language=Prompt,caption={Video pointing prompt for baselines with video inputs},
                   label={lst:video-prompt}]
You are a video-analysis assistant that points to unique target objects in the video at 2FPS.

Goal:
Point to the timestamp and spatial coordinates of target objects, actions, or events in the input video.
- timestamp (as a string in `HH:MM:SS` format, where the second can be to the closest 0.5 seconds e.g. `00:01:23.5`)
- x_min, y_min, x_max, y_max (integer coordinates normalized to a 0-1000 scale)

Rules (strict):
- For actions/events spanning some time, pick the most representative / clear timestamp.
- Each instance should be a separate spatial-temporal point in "results".
- Do NOT point to the same object more than once.
- Return only valid JSON, without markdown code blocks, explanations, or extra text.

Output format (strict JSON):
{
  "results": [
    {
      "timestamp": <str>, `HH:MM:SS` format
      "x_min": <int>,
      "y_min": <int>,
      "x_max": <int>,
      "y_max": <int>
    },
    ...
  ]
}

Target: {label}
\end{lstlisting}

\begin{lstlisting}[language=Prompt,
                    caption={Video pointing prompt for baselines with image inputs},
                   label={lst:image-prompt}]
You are a video-analysis assistant that points to unique target objects in the video, (*@\textbf{represented as a sequence of image frames at 2FPS}@*).

Goal:
Point to the timestamp and spatial coordinates of target objects, actions, or events in the input (*@\textbf{video frames at 0.5 second intervals}@*).
- timestamp (as a string in `HH:MM:SS` format, where the second can be to the closest 0.5 seconds e.g. `00:01:23.5`)
- x_min, y_min, x_max, y_max (integer coordinates normalized to a 0-1000 scale)

Rules (strict):
- For actions/events spanning some time, pick the most representative / clear timestamp.
- Each instance should be a separate spatial-temporal point in "results".
- Do NOT point to the same object more than once.
- Return only valid JSON, without markdown code blocks, explanations, or extra text.

Output format (strict JSON):
{
"results": [
    {
    "timestamp": <str>, `HH:MM:SS` format
    "x_min": <int>,
    "y_min": <int>,
    "x_max": <int>,
    "y_max": <int>
    },
    ...
]
}

Target: {label}
\end{lstlisting}

\paragraph{Tracking.}
We explain the tracking evaluation setup used for Tables~\ref{tab:track_result_academic}--\ref{tab:molmo2_track_results}. Across all benchmarks, segmentation metrics are computed at the original video frame rate, while point-based metrics are evaluated at 1 fps and marked as correct if they fall inside the mask. For baselines, we evaluate specialized open segmentation models that output a single foreground mask per frame and report their segmentation quality. When a model can produce discrete points per object (e.g., VLMs), we additionally report its point-based metrics. We found that API models and generic VLMs are incapable of producing accurate point tracks, as shown in the video pointing task (Table~\ref{tab:video_count_and_point_results}), but their grounding performance improves substantially when prompted to output bounding boxes instead. Thus, for these models, we predict bounding boxes at 1-second intervals, use the boxes to prompt SAM 2 to generate segmentation masks, and take the box centers as representative points for point-based metrics. Our model, instead, can predict discrete point tracks with explicit IDs, and their points are directly fed to SAM 2 to obtain segmentation masks.

For metrics, we report their average $\boldsymbol{\mathcal{J}\&\mathcal{F}}$ over all objects and frames as a standard metric for segmentation quality. The Jaccard index $\mathcal{J}$ measures region overlap between predicted and ground-truth masks via intersection-over-union (IoU). The boundary F-score $\mathcal{F}$ measures how well predicted and ground-truth object contours align. \textbf{Point F1} is computed similarly to the video counting task but at 1 fps, and captures frame-wise detection performance. Since Point F1 is insensitive to identity swaps when the number of objects remains constant, we also report \textbf{HOTA}~\cite{luiten2021hota} ($\text{HOTA} = \sqrt{\text{DetA} \times \text{AssA}}$) to measure tracking quality, which jointly scores detection accuracy (DetA) and association accuracy (AssA). While originally designed for bounding box tracking, where similarity is measured via IoU, we adapt HOTA to point-based tracking by defining similarity as binary: a predicted point matches a ground-truth object if it falls within the object's segmentation mask. DetA then measures whether points are placed in correct masks, while AssA measures whether consistent object IDs are maintained over time based on their presence in the mask and penalizes identity switches if swapped. Since baseline models do not output stable track IDs but only counts, HOTA is only reported for \model{} that can perform tracking reliably.

Table~\ref{tab:track_result_academic} presents comprehensive results across all academic benchmarks and their splits. We see \model substantially outperforms API-based and open-source VLMs by a wide margin, suggesting the existing VLMs are not well-suited for object tracking tasks. Specialized open models that directly generate segmentation also fall behind our approach, indicating their inability to effectively ground object semantics despite being specifically trained for tracking. The most directly comparable baseline is VideoMolmo~\cite{ahmad2025videomolmo}, another video language model trained for point grounding in videos. While specialized models perform on par or outperform our model on Ref-Davis, which involves single objects with simple text queries, our model excels in more complex scenarios beyond basic tracking, where it significantly outperforms multi-object tracking supported in MeViS~\cite{ding2023mevis} and reasoning-intensive tasks in ReasonVOS~\cite{yan2024visa}.

Lastly, we report the performance on our proposed benchmark \model-Track in Table~\ref{tab:molmo2_track_results}, further broken down by video domains. Overall, \model comes out on top, outperforming other VLMs and even the specialized open video models. Across the board, API-based and open-source VLMs, including Molmo and VideoMolmo~\cite{ahmad2025videomolmo}, struggle to count and track consistent objects throughout videos, as indicated by their low F1 and HOTA scores. Interestingly, the Molmo variants and specialized models achieve a high segmentation score ($\mathcal{J} \& \mathcal{F}$), though we observe that for cluttered scenes--such as Pedestrians, Sports, and Dancers--models generate large, coarse masks covering entire people rather than precisely localizing individual objects. This results in high region overlap that inflates $\mathcal{J} \& \mathcal{F}$ while failing to accurately ground and track specific objects, as reflected in the substantially lower F1 and HOTA scores. This highlights the importance and necessity of our point-based F1 and identity-aware HOTA metrics, which more directly measure a model's ability to precisely ground and track the correct objects.





\section{Additional results}
In this section, we present several additional evaluations.

\label{appendix:add_eval}






\subsection{Additional model ablations}

\begin{table}[!ht]
    \centering
    \small
    \begin{tabular}{l c c c c}
        Pretrain  & Video QA & Molmo2 Video Cap. & Image QA & Image Pointing \\
\hline
\baseline{With pointing} & 66.8 & 31.8 & 80.9 & 73.0 \\
No pointing & 65.9 & 31.3 & 80.1 & 71.8 \\
    \end{tabular}
        \caption{\textbf{Pre-training ablations}. Columns show the average of our 12 video benchmarks, using validation sets for EgoSchema, PerceptionText, and MLVU, video captioning F1, the average of the 11 image benchmarks using validation sets for InfoQA, DocQA, ChartQA, VQA v2, and AI2D, and the average score in Point-Bench.}
    \label{tab:pretrain_ablation}
\end{table}

\paragraph{Pre-traing ablation}. We also present an ablation without image-pointing pre-training in Table~\ref{tab:pretrain_ablation}. This model is only trained on image captioning and NLP data. For the SFT stage, it uses 2x the sampling rate for the image pointing datasets and 28k steps of training instead of 25k to compensate for the fact that the image pointing data is not seen during pre-training. We observe a small decrease in the benchmarks in this setting, even for those not related to image pointing. We hypothesize that pointing pre-training simplifies the SFT stage for the model since it no longer needs to learn the basic pointing format and task, allowing for more focus on the non-pointing tasks.






\subsection{NLP Benchmarks}
\begin{table}[!h]
    \centering
    \small
    \begin{tabular}{@{}l c c c c@{}}
        Model & MMLU~\cite{mmlu} & GSM8K~\cite{gsm8k} & ARC-C~\cite{arc_c} & MBPP+~\cite{mbpp} \\
        \hline
        Qwen3-4B~\cite{qwen3technicalreport}   & 72.2 & 87.8 & 83.3 & 59.5 \\
        Qwen3-8B~\cite{qwen3technicalreport}   & 76.8 & 89.8 & 88.3 & 62.2 \\
        OLMo3-7B-Instruct~\cite{olmo3}   & 69.1 & 90.1 & 72.2 & 60.2 \\
        
        \textcolor{molmocolor}{\model{}-4B}  & 72.2 & 86.6 & 89.3 & 56.2 \\
        \textcolor{molmocolor}{\model{}-8B}  & 76.6 & 89.7 & 89.6 & 57.5 \\
         \textcolor{molmocolor}{\model{}-O-7B}  & 64.1 & 89.0 & 79.9 & 55.7 \\
    \end{tabular}
    \caption{\textbf{Results on selective NLP benchmarks}, including MMLU for general knowledge QA, GSM8K for math, ARC-C for reasoning, and MBPP+ for coding tasks.}
    \label{tab:nlp_results}
\end{table}

We evaluate \model{} on selective NLP benchmarks covering general knowledge QA, math, reasoning, and coding tasks and report their results compared to the base language models Qwen3 in Table~\ref{tab:nlp_results}. We run evaluations for all models following OLMo 3's evaluation protocol, except for OLMo3-7B-Instruct's MMLU and MBPP+ numbers, which we take directly from OLMo3's model card.  We find that \model{} achieves comparable numbers on the general knowledge QA and math benchmarks, MMLU and GSM8K, but suffers from some drops in coding on the MBPP+ coding benchmark~\cite{mbpp}. Interestingly, both \model-4B and \model-8B perform slightly better than their respective base language models in the ARC Challenge multiple-choice evaluation. 

\begin{figure}[!ht]
    \centering
    \includegraphics[width=.6\columnwidth]{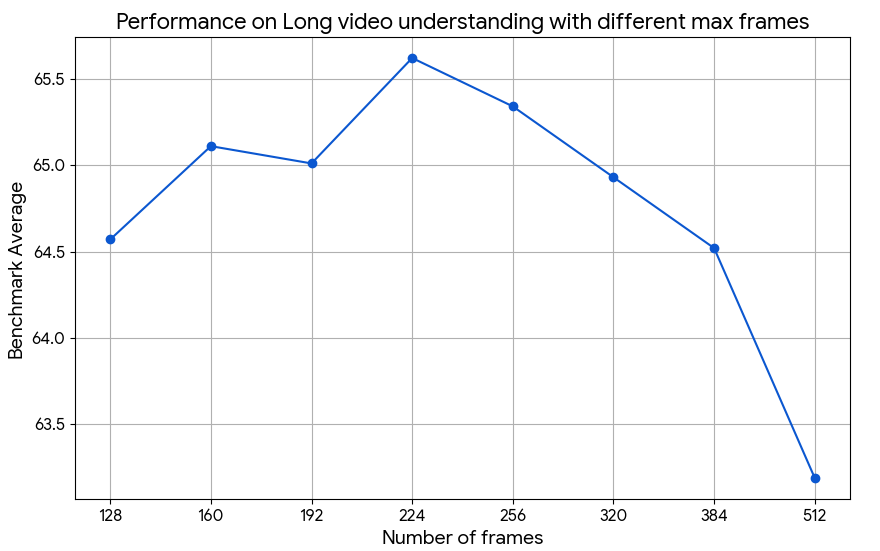}
    \caption{\textbf{Long video benchmark results with different max frames}, the average of our six long video benchmarks.}
    \label{fig:long_video_avg_more_frames}
\end{figure}

\section{Test time scaling with 128-frame model}
\label{appendix:test-time}
\begin{table*}[!ht]
    \centering
    \setlength{\tabcolsep}{3pt}
    \resizebox{\textwidth}{!}{
    \begin{tabular}{@{}lccccccccc@{}}
        \textbf{Model}
        & VTok
        & Video-MME
        & Video-MME-Sub
        & LongVideoBench
        & MLVU
        & LVBench
        & VideoEvalPro
        & Short QA avg
        & Long QA avg\\
        \midrule
        \colorbox{baselinecolor}{128 frames}             & 10.6k & 68.8 & 74.3 & 65.9 & 74.5 & 49.6 & 54.3 & \textbf{69.8} & 64.6 \\
        pool4, 216 frames    & 11k & 68.9 & \textbf{75.0} & 64.3 & 75.7 & 48.9 & 54.9 & 68.8 & 64.6 \\
        pool5, 332 frames    & 10.6k & 69.1 & 74.2 & 64.2 & \textbf{76.5} & 50.6 & 56.9 & 68.4 & 65.2 \\
        128 frames + SF-periodic    & 10.7k & 68.1 & 74.5 & 64.2 & 74.5 & 48.3 & 53.5 & 69.6 & 63.9 \\
        128 frames + SF-diff    & 10.7k & 68.4 & 74.1 & 64.7 & 75.7 & 48.7 & 54.8 & 69.6 & 64.4 \\
        128 frames + SF-query                            & 10.7k & 68.9 & 73.9 & \textbf{66.6} & 76.2 & \textbf{51.5} & \textbf{57.2} & 69.6 & \textbf{65.7} \\
        128 frames + SF-tr-0.1 & 10.7k & 69.1 & 74.3 & 65.4 & 75.0 & 48.6 & 54.3 & 69.8 & 64.4 \\
        128 frames + SF-tr-0.1 + SF-query                            & 10.7k & 68.9 & 74.3 & 65.5 & 75.4 & 51.5 & 57.1 & 69.8 & 65.5 \\        
        224 frames   & 18.6k & \textbf{69.2} & 74.6 & 66.1 & 76.4 & 50.7 & 56.7 & 69.7 & 65.6 \\
 \\
    \end{tabular}
    }
    \caption{\textbf{\model-8B with test time scaling / SlowFast (SF) encoding} 
    SF-query boosts long video understanding and matches using 224 frames while using $\sim43\%$ fewer visual tokens. Training without SF and then using SF-query marginally beats training with SF-tr-0.1 on long video understanding tasks. 
    All SlowFast models use a max of 368 frames. VTok denotes max vision tokens. SF-tr-0.1 denotes using SlowFast $10\%$ of the time in training. 
    }
    \label{tab:trained_slow_fast}
\end{table*}
In this section, we consider whether it is possible to scale the number of frames past 128 during inference without long-context training. We also test an approach using SlowFast~\cite{xu2024slowfast} to provide the model with a mix of high and low-resolution frames during inference, or during both training and inference.

\paragraph{Increasing max frames.} At test time, we scale the maximum number of frames for better long video understanding. We evaluate \model-8B after the SFT stage, but before long-context training, with $160$, $192$, $224$, $256$, $320$, and $512$ max frames and report the average on the val sets of our six long video understanding benchmarks in 
Figure~\ref{fig:long_video_avg_more_frames}. \model has the best performance with $224$ frames for long video benchmarks. For short video understanding benchmarks, the average is 69.8 for $128$ frames and 69.7 for all other settings as shown in Table~\ref{tab:trained_slow_fast}.

\paragraph{Keeping Vision tokens fixed.} However, increasing the maximum number of frames also increases the number of vision tokens fed into the model, which raises compute cost and may not be feasible on GPUs with limited memory. With the default setting of max $128$ frames, the maximum number of vision tokens is $83 * 128 \sim 10.6k$. We therefore evaluate alternative test-time strategies that keep the number of max vision tokens close to $10.6k$. Specifically, we evaluate different pooling strategies in the vision-language connector - $4 \times 4$ pooling with $216$ frames and $5 \times 5$ pooling with $332$ frames. The $5 \times 5$ pooling setting improves long video understanding by accessing more frames; however, both settings regress on short video understanding (Table~\ref{tab:trained_slow_fast}).

\paragraph{SlowFast encoding.} Since we find that our model can generalize to different pooling sizes at test time, we further explore a SlowFast video strategy~\cite{xu2024slowfast}. We build on the interleaved SlowFast variant used in~\cite{sfllava1_5, yang2025kwai, shen2024longvu}, which dynamically allocates computational resources across frames by varying their spatial pooling in the \model connector, with each frame represented exactly once -- either in the slow or the fast pathway. Frames are categorized as slow or fast based on a periodicity parameter $p$: every $p$-th frame is designated as a slow frame, while the remaining frames are fast frames. We refer to this approach as \textit{Slowfast-periodic}. Note that $p=1$ reduces to the default setting. Slow frames use the default pooling size of $3 \times 3$, whereas fast frames use $9 \times 9$ pooling. We use four different periodicities $p \in \{1, 2, 3, 4\}$ with corresponding max frames $M \in \{128, 224, 300, 368\}$. The max frame $M$ for each periodicity is chosen such that the maximum number of vision tokens input to the LLM is approximately $10.6k$. $10.6k$ is the maximum number of vision tokens used in the default setup of \model. When processing a video with SlowFast encoding, after we sample $F_t$ frames, $p$ is selected to maximize the tokens in the slow pathway. For example, when $F_t \le 128$, we use $p=1$ and all the frames are in the slow pathway, or when $128 < F_t \le 224$, we use $p=2$ and every other frame is in the slow pathway. In practice, that leads to stepwise changes in selected $p$ as the number of frames ranges from $1$ to $368$.

We explore two strategies to score the frames' relevance for inclusion in the slow pathway. First, we embed both the query and all the frames using SigLIP 2~\cite{siglip2} and calculate per frame cosine similarity scores. Second, we calculate the average of the absolute similarity difference of the embedded frames with their neighboring frames. In either strategy, we use the per-frame score to select the relevant frames for the slow pathway. Our formulation when selecting $F_s$ slow pathway frames from $F_t$ sampled frames is to include both frames that globally have the highest scores and frames that have high scores in their local neighborhoods. To select locally high scoring frames, we first select $F_s/2$ frames by choosing the single highest scoring frame from temporally ordered groups of size $F_t \div F_s/2$. To select globally relevant frames, we select the remaining $F_s/2$ frames that have the highest scores from all the remaining frames. Additionally, we don't use score based selection and use Slowfast-periodic when the frames per second $F_r$ is high. This follows the intuition that frame selection is useful when selecting amongst sparser frames for long videos with multiple scenes, but not for shorter videos that get densely sampled and tend to have only one scene. In practice, we fall back to Slowfast-periodic when $F_r \geq 2$.

With Slowfast-periodic, the model regresses on the long video understanding, contrary to the finding in ~\cite{xu2024slowfast}. Using the frame difference improves over using periodic sampling, but still lags behind the default setting. However, using the query to select frames for the slow pathway achieves the best performance. It provides a boost to long video understanding with minor regression in short video understanding. It closes the gap to the optimal setting of using 224 frames while having $\sim43\%$ fewer visual tokens (Table~\ref{tab:trained_slow_fast}).

\paragraph{Training with SlowFast.} Due to the improvement on long video understanding tasks using SlowFast encoding in the training-free regime, we explore training with SlowFast. We report results for training in a combined single stage starting from the image captioner. We keep the max frames the same $128$ and sample using the SlowFast setup with a probability $P_{sf}$ while randomly sampling different $p \in {2, 4, 8}$. We use the default sampling with a probability if $1 - P_{sf}$ and use $P_{sf} = 0.1$. When training with a SlowFast setup, we randomize the slow frames. Concretely, to select $F_s$ frames from $F_t$ sampled frames, 1 frame in ordered groups of size $F_t \div F_s$ is selected randomly. Even though the max frames is not increased, the goal is to familiarize the video model with the SlowFast encoding similar to score-based Slow frame selection, but without increasing the training cost by requiring the use of more frames. At test time, we evaluate with and without the query based SlowFast setup described above. Surprisingly, training without SF and then using the query to select Slow frames beats training with SF $10\%$ of the time as shown in Table~\ref{tab:trained_slow_fast}. This suggests \model{} can frame using $9 \times 9$ pooling even though such frames were not seen during training.


\section{Dataset details}
In this section, we provide additional details about our data collection methodology.
\label{appendix:data}

\subsection{Dataset statistics}

\begin{figure}[htbp]
  \centering
  \begin{minipage}[t]{0.48\linewidth}
    \centering
    \includegraphics[width=\linewidth]{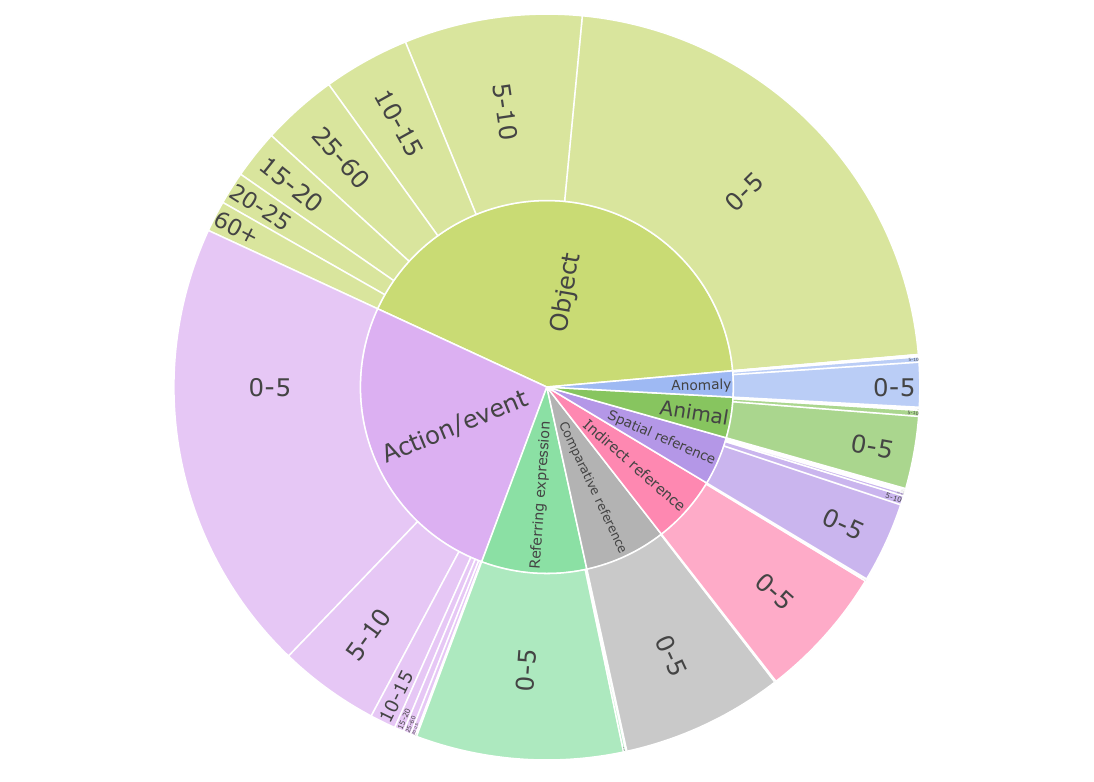}
    \caption{The distribution of categories and counts across pointing queries in \textbf{\model-VideoPoint}.}
    \label{fig:points_cat_dist}
  \end{minipage}
  \hfill
  \begin{minipage}[t]{0.48\linewidth}
    \centering
    \includegraphics[width=\linewidth]{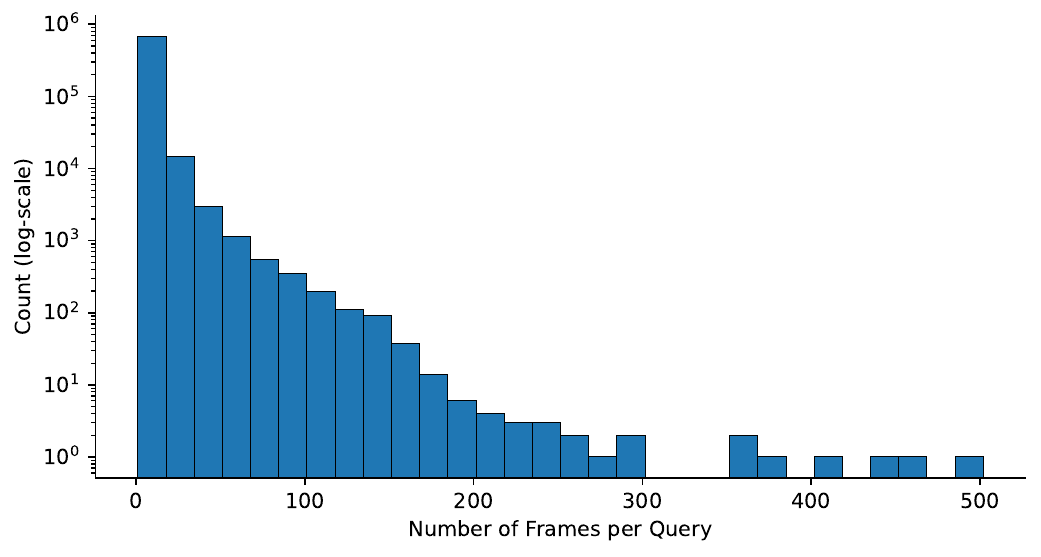}
    \caption{The distribution of annotated frame count per query in \textbf{\model-VideoPoint}.}
    \label{fig:points_frame_dist}
  \end{minipage}
\end{figure}

\begin{figure}[htbp]
  \centering
  \begin{minipage}[t]{0.48\linewidth}
    \centering
    \includegraphics[width=\linewidth]{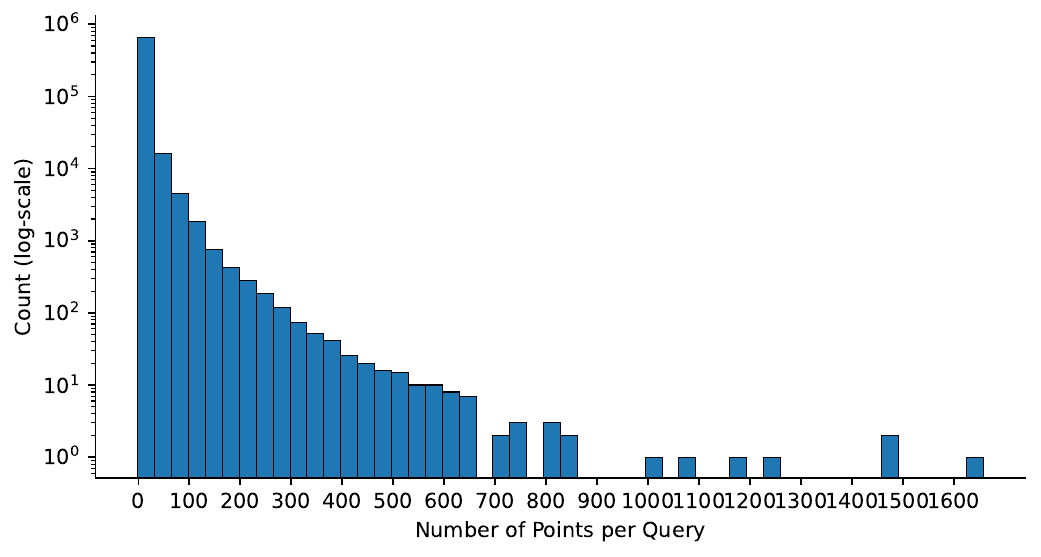}
    \caption{The distribution of annotated point count per query in \textbf{\model-VideoPoint}.}
    \label{fig:points_point_dist}
  \end{minipage}
  \hfill
  \begin{minipage}[t]{0.48\linewidth}
    \centering
    \includegraphics[width=\linewidth]{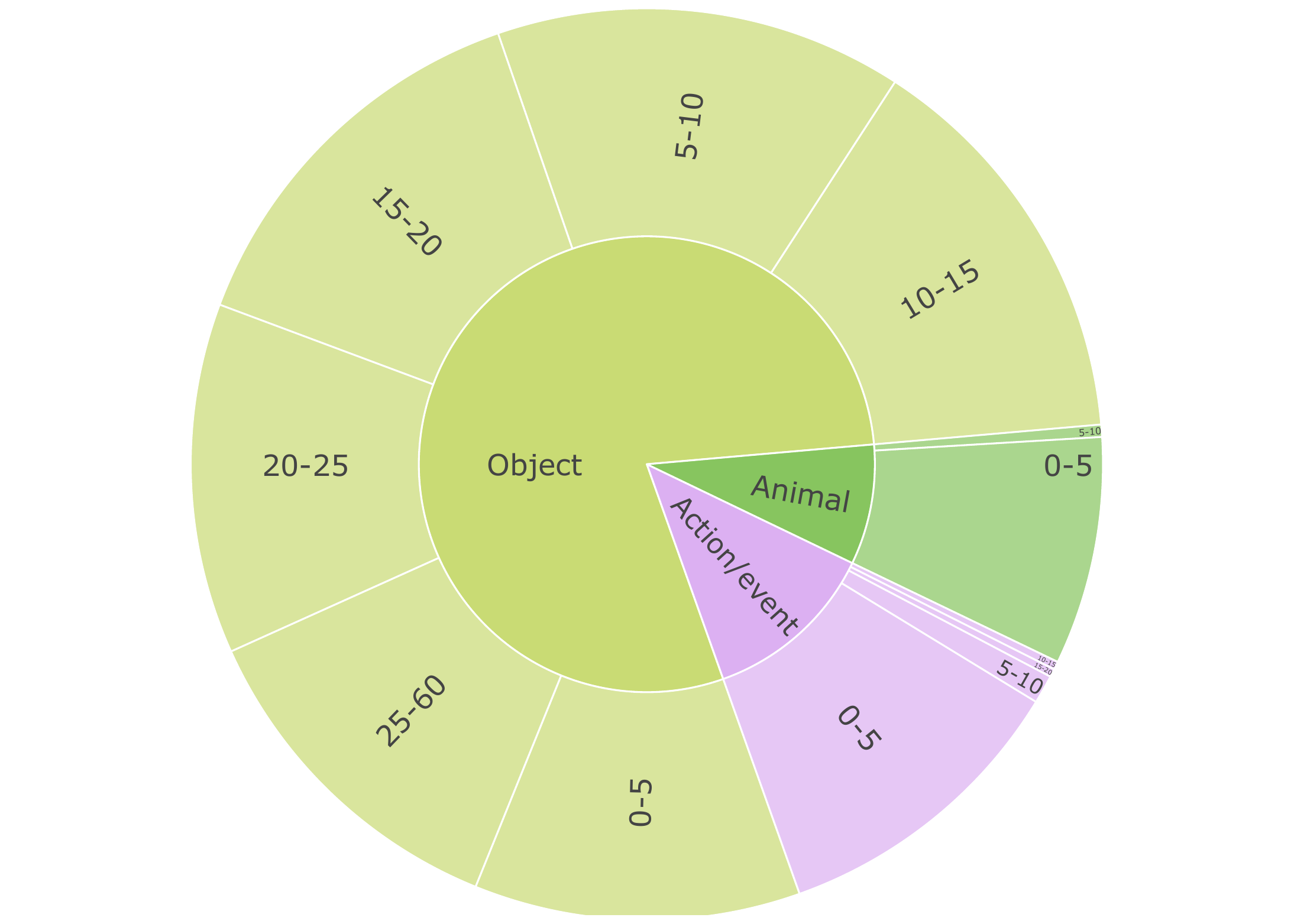}
    \caption{The distribution of categories and counts across queries in the \textbf{\model-VideoCount} evaluation.}
    \label{fig:points_count_val_dist}
  \end{minipage}
\end{figure}

\begin{figure}[htbp]
  \centering
  \includegraphics[width=.48\linewidth]{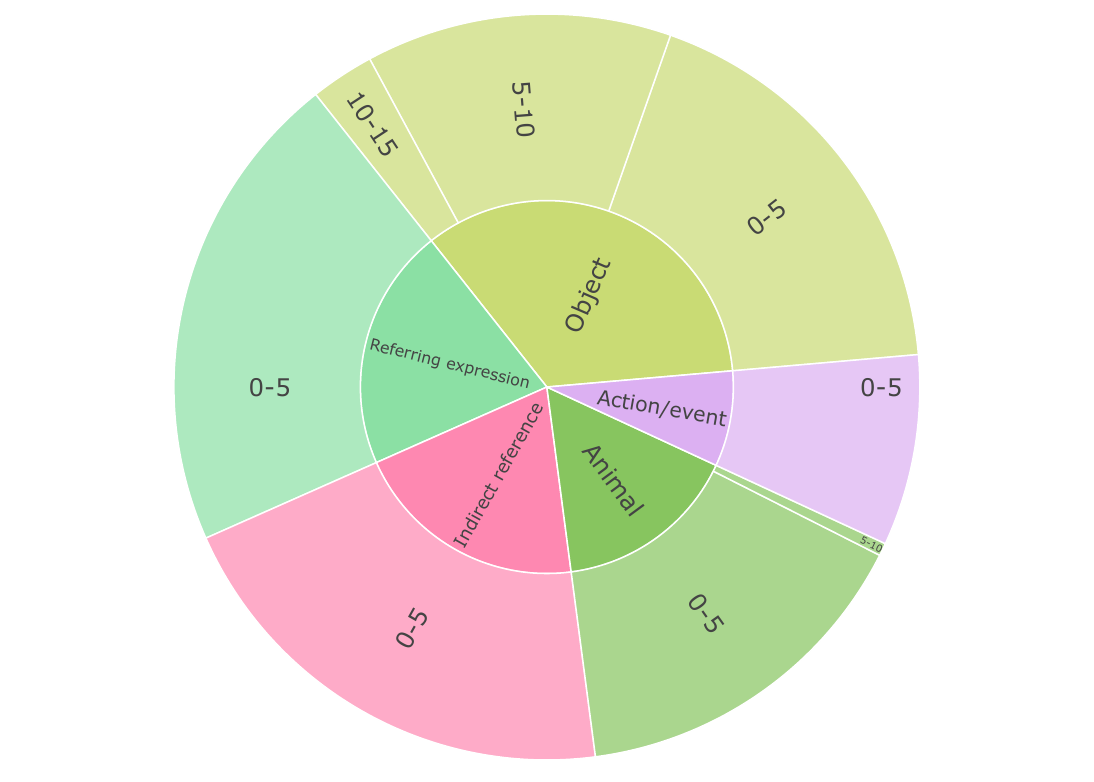}
  \caption{The distribution of categories and counts across queries in the \textbf{\model-VideoPoint} evaluation.}
  \label{fig:points_point_val_dist}
\end{figure}

\paragraph{Pointing.} We report the statistics on the \model-VideoPoint training and validation sets. Overall, the \model-VideoPoint dataset contains diverse pointing queries across seven categories (Figure~\ref{fig:points_cat_dist}). There are more queries in Action/Event, Object, and Referring expression, as we expect these to be harder for the model to learn. We also see that the distribution is skewed towards low-count examples with 0 to 5 counts (Figure~\ref{fig:points_cat_dist} and~\ref{fig:points_point_dist}). We mitigate this bias by upsampling medium- and high-count examples during training, and plan to collect more high-count examples in the future. Similarly, the distribution of frames annotated per query is also heavily skewed to the left (Figure~\ref{fig:points_frame_dist}). 

For the validation sets used in \model-VideoCount and \model-VideoPoint evaluations, we carefully build them by (1) collecting double annotations on some queries and selecting high-confidence examples where two different annotators provide the same answer; and (2) sampling queries across diverse categories and counts (Figure~\ref{fig:points_count_val_dist} and ~\ref{fig:points_point_val_dist}). For video counting, we mostly sample queries from the object category, as there are significantly more high-count examples in this category than in others (Figure~\ref{fig:points_count_val_dist}). For video pointing evaluation, we intentionally pick queries in the more difficult categories -- referring expression and indirect reference (Figure~\ref{fig:points_point_val_dist}) -- orthogonal to the ones in the counting evaluation, so that we have a comprehensive evaluation of our model's counting and pointing capabilities.

\begin{figure}[!ht]
    \centering
    \begin{minipage}[t]{0.48\linewidth}
        \centering
        \includegraphics[width=\linewidth]{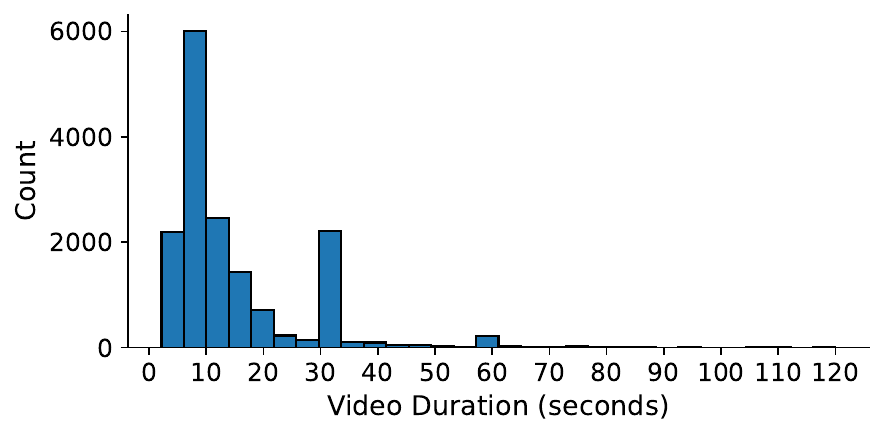}
        \caption{Distribution of video clip duration in \textbf{\model-VideoTrack} and \textbf{\model-Track}.}
        \label{fig:vid_track_clip_lengths}
    \end{minipage}
    \hfill
    \begin{minipage}[t]{0.48\linewidth}
        \centering
        \includegraphics[width=\linewidth]{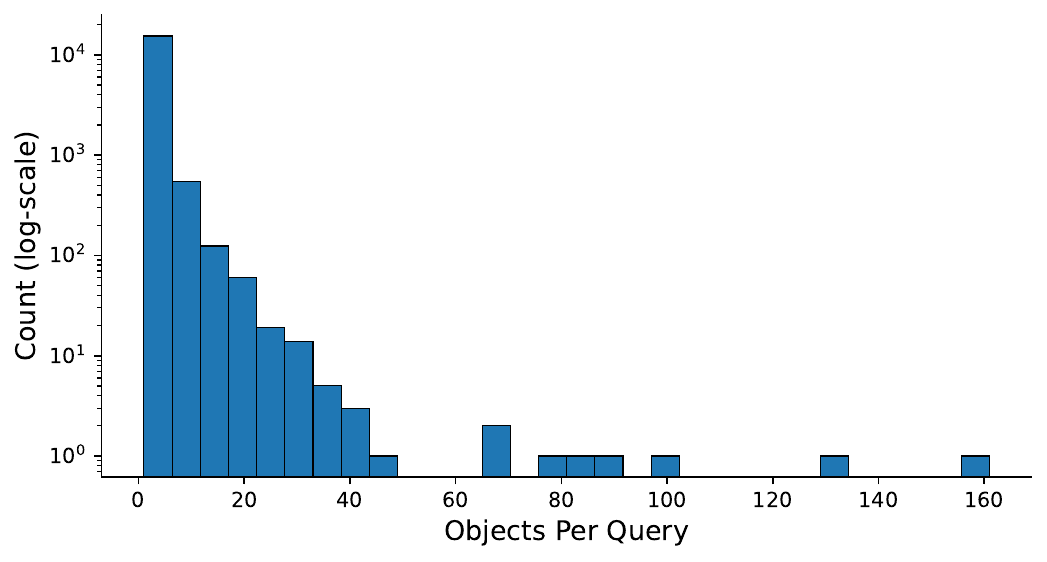}
        \caption{Distribution of objects described by text queries in \textbf{\model-VideoTrack} and \textbf{\model-Track}.}
        \label{fig:vid_track_objects_per_query}
    \end{minipage}
    \hfill
    \begin{minipage}[t]{0.48\linewidth}
        \centering
        \includegraphics[width=\linewidth]{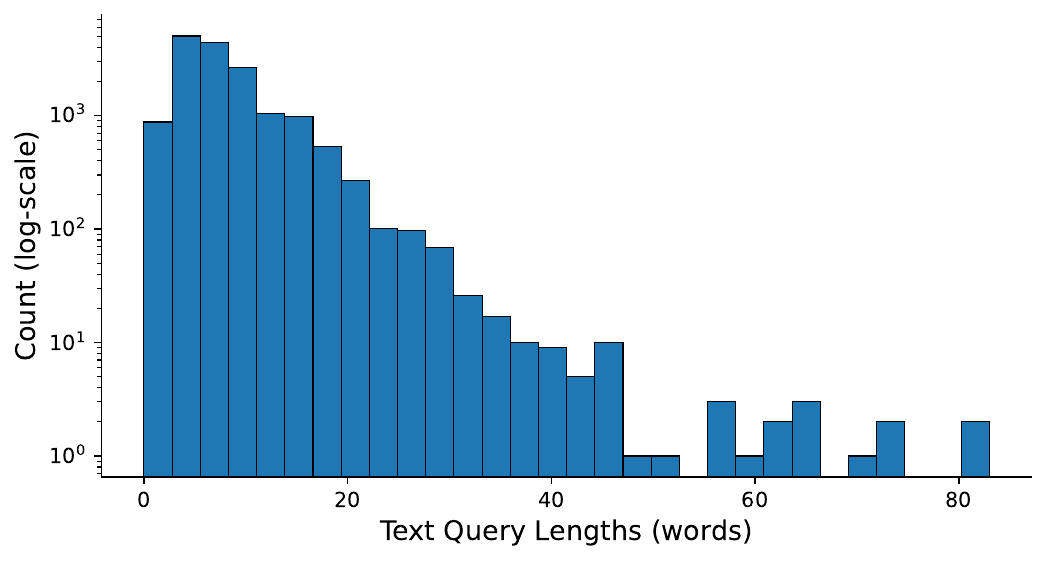}
        \caption{Distribution of text query lengths in \textbf{\model-VideoTrack} and \textbf{\model-Track}.}
        \label{fig:vid_track_query_lengths}
    \end{minipage}
\end{figure}

\paragraph{Tracking.}
We report statistics on the videos and text queries in \model-VideoTrack and the \model-Track benchmark. The two datasets have a total of 8k video clips, with 6.6k for training and 1.3k for evaluation. Both datasets provide segmentation masks, text queries, and metadata for each video. On average, there are 6.08 annotated objects per video, and the videos are up to 2 minutes long, with most being around 10-30 seconds. The distribution of video durations is shown in Figure ~\ref{fig:vid_track_clip_lengths}. 

Our dataset contains a total of 29k diverse text queries covering a wide variety of categories, bringing an average of 1.33 text queries per video. The distribution of categories is detailed in Figure~\ref{fig:tracking-training-categories} and Figure~\ref{fig:tracking-benchmark-categories}. Multi-object tracking is a primary focus in the tracking capabilities of \model, so we strived to find text queries that describe many objects within a video. The dataset has an average of 3.31 objects described per text query, with many queries describing far more than that. The distribution is shown in Figure~\ref{fig:vid_track_objects_per_query}. Each text query is on average 8.21 words long, but there is a wide range. The exact distribution across all text queries is shown in Figure~\ref{fig:vid_track_query_lengths}. 



\subsection{Data collection}
Here, we detail how we collect videos and synthesize annotations for most of \model video datasets.

\paragraph{Video collection for \model-Cap.}
We first source videos less than 3 minutes from multiple large-scale datasets~\cite{zellers2022merlotreserve,wang2024koala36mlargescalevideodataset,wang2023internvid,llava_video} and YouTube videos searched with keywords used in MetaCLIP~\cite{xudemystifying} to form a pool of over 10M videos.

Then, we perform one step of filtering based on the informativeness of the video: we first discard the audio track and uniformly sample the video at 1 fps; Then the sampled frames are encoded using H.264; The total size of the resulting encoded stream (in bits) is divided by the product of the video duration and spatial resolution (duration × W × H) to obtain a normalized video informativeness score. After collecting scores for all videos in the pool, we discard those whose score falls below (mean - 1 standard deviation), effectively removing videos with unusually low visual or temporal diversity.

After this filtering, we conduct a diversity-based sampling to obtain a final set of videos for human annotation: for each remaining video, we uniformly sample 5 frames and apply SAM 2~\cite{ravi2024sam2} to segment each frame, computing the average number of segments as a proxy for visual complexity. We further use Molmo to caption each sampled frame and follow MetaCLIP’s processing pipeline to extract a set of keywords that characterize its semantic content. To select a diverse subset, we perform a greedy sampling procedure that aims to maximize the entropy of both the segment-count distribution and the keyword distribution. At each step, we score all candidate videos using a two-stage ranking: (1) we compute a “what-if” entropy gain for the keyword distribution if the candidate were selected, and rank candidates accordingly; (2) we compute a density-based score that favors videos contributing to underrepresented segment-count regions. The final score is obtained by summing the two ranks, and we select the top-ranked candidate. For efficiency, we approximate this process by scanning the pool in chunks of 1,000 candidates at a time, rather than evaluating the entire pool at each iteration. This procedure yields a video subset that is both semantically diverse and visually varied, providing a strong foundation for high-quality human annotations.
Finally, we set the sampling ratio to be 1\% and obtained around 100k videos.

\paragraph{Video and synthetic annotation collection for \model-CapQA, -SubtitleQA, -VideoPoint, and -AskModelAnything.}
 We first source 500k videos with Creative Commons license from YT-Temporal~\cite{zellers2022merlotreserve} and YouTube keyword search. Then we use a video captioner trained on \model-Cap to caption these videos. In particular, we segment each video into multiple scenes and caption each scene instead of the entire video to encourage detailed descriptions. 
Since model-generated captions can sometimes be low-quality, we apply a heuristic rule-based filter to remove captions with repetition patterns.
The final set of videos and synthetic captions is used to curate \model-CapQA, -SubtitleQA, and -VideoPoint datasets.

For \model-CapQA and \model-SubtitleQA, we prompt an LLM to generate both the question and the answer. For \model-VideoPoint, we prompt an LLM to generate the queries and solicit human answers. For \model-AskModelAnything, we elicit questions from human annotators and generate the corresponding answers using an LLM with human feedback.

\subsection{Data annoation}
\begin{figure*}[!t]
  \centering
  \includegraphics[width=0.8\linewidth]{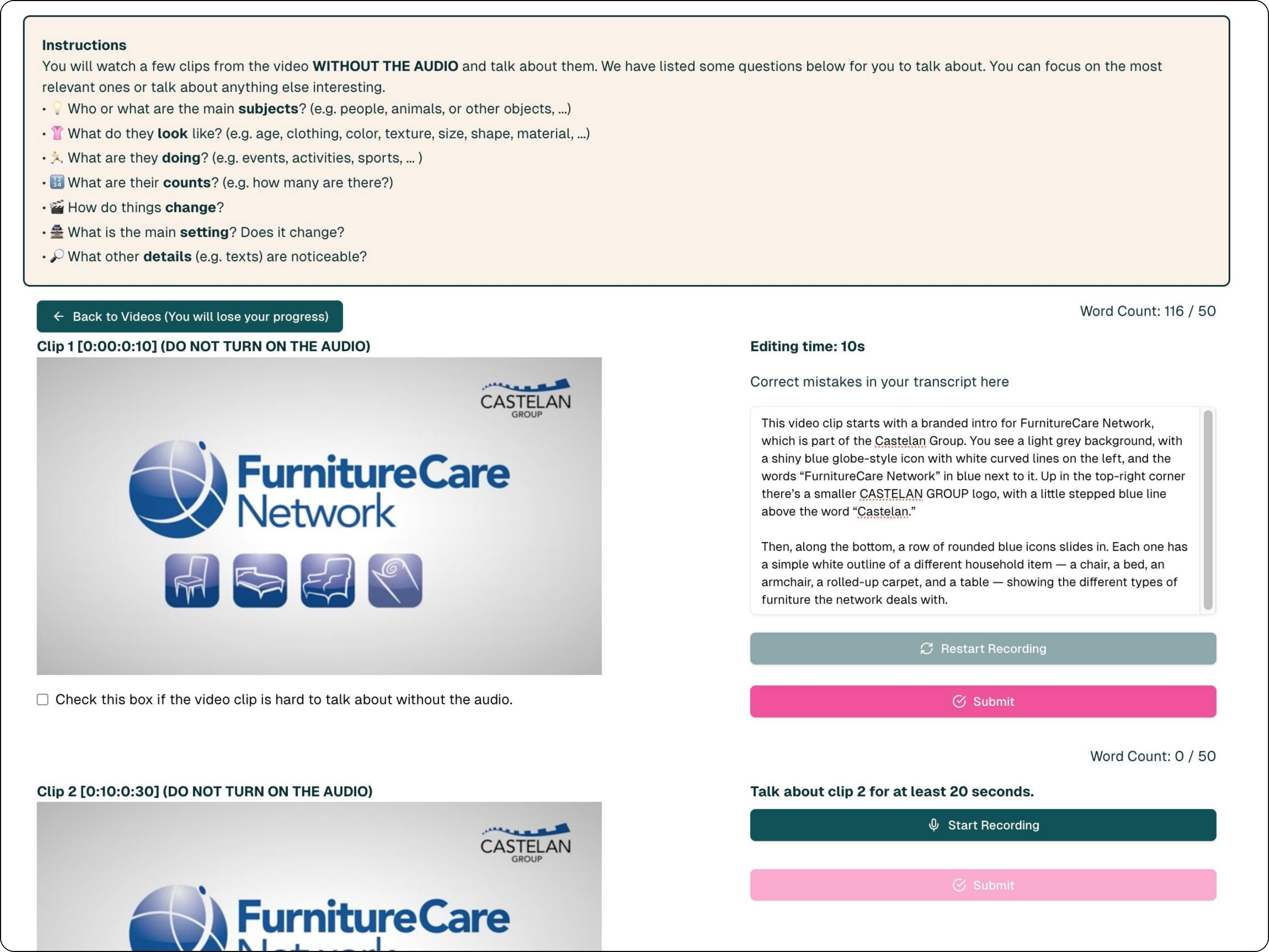}
  \caption{Video clip captioning interface. Crowdworkers are instructed to annotate captions for video clips in sequence.}
  \label{fig:interface-clipcaption-0}
\end{figure*}

\begin{figure*}[!t]
  \centering
  \includegraphics[width=0.8\linewidth]{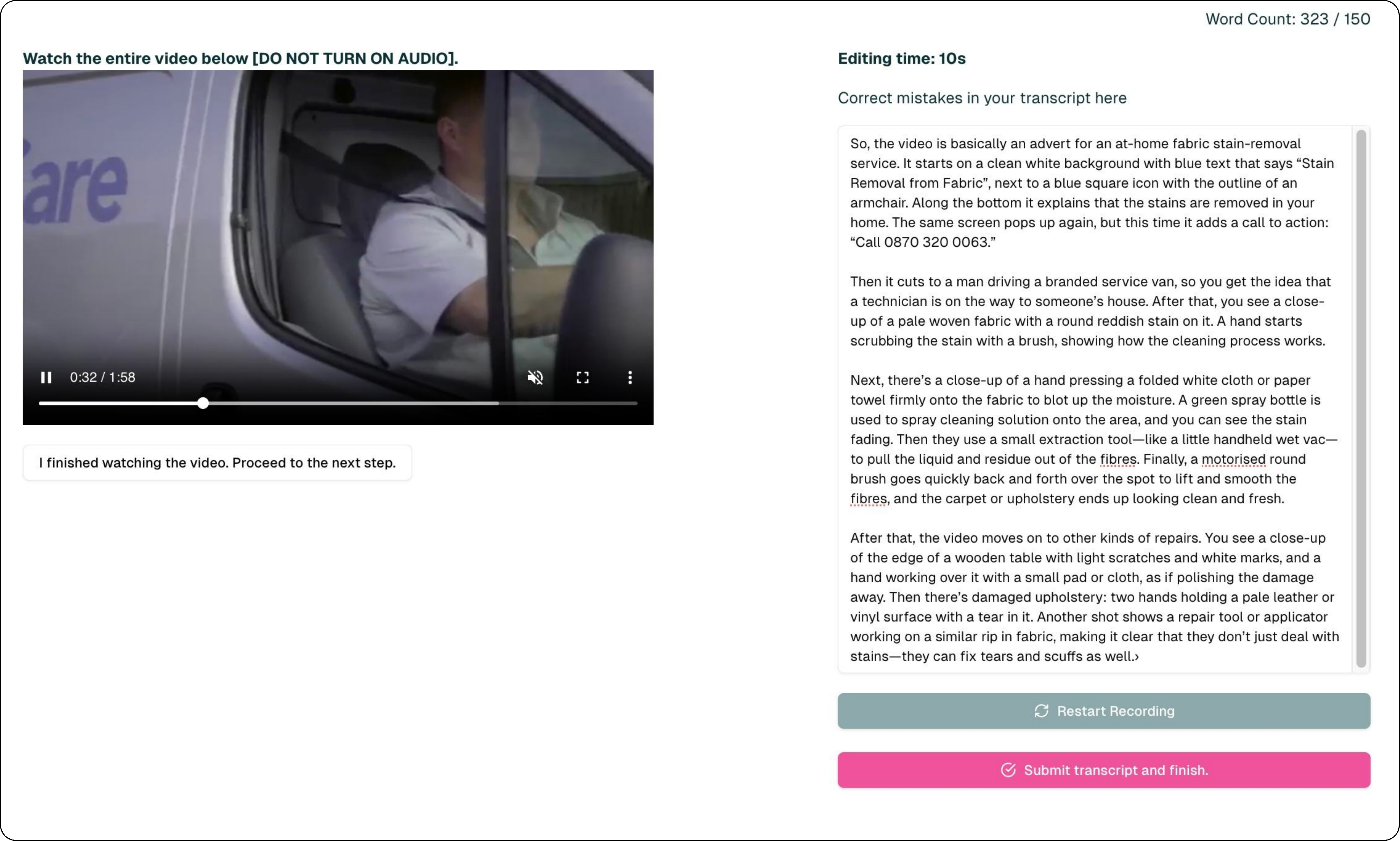}
  \caption{Video captioning interface. Crowdworkers are instructed to annotate captions for complete videos.} 
  \label{fig:interface-clipcaption-1}
\end{figure*}

\paragraph{\model-Cap.}  To obtain clips for the first-stage captioning, we develop an algorithm to split a video into clips of variable lengths between 10 and 30 seconds based on their information density so that a more informative clip has a shorter duration. This algorithm minimizes the highest information density of a video clip across all clips. Overall, videos are split into 4-5 clips on average. We then deploy the video-description task to online crowdworkers (see Figure~\ref{fig:interface-clipcaption-0} for the task interface). For each full video, workers are first shown a sequence of shorter clips split by our algorithm from the original video with audio muted. At the top of the interface, we provide instructions to guide their descriptions. For each clip, workers verbally describe what is happening on the screen, and their speech is automatically converted to text via real-time transcription. They then edit the transcript to correct recognition errors before submitting it. After completing all clips, workers are asked to provide a comprehensive description of the full video (see Figure~\ref{fig:interface-clipcaption-1}).

\begin{figure*}[!t]
  \centering
  \includegraphics[width=0.8\linewidth]{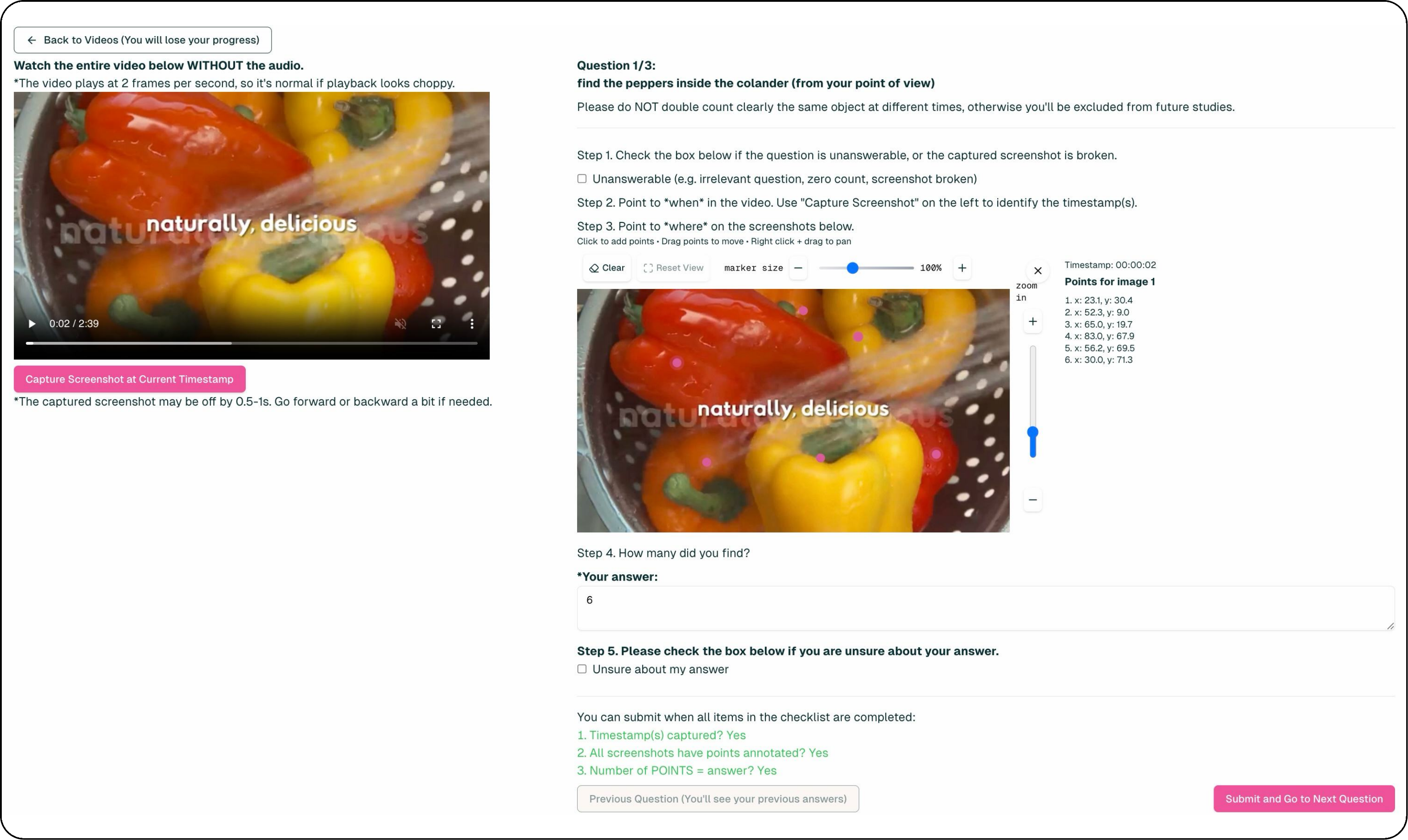}
  \caption{Video pointing interface. Crowdworkers are instructed to  annotate points for object instances to answer visual questions.} 
  \label{fig:interface-videopoints-0}
\end{figure*}

\paragraph{\model-VideoPoint.} 
For each video, we design several visual questions that require workers to answer using evidence from a single or several frames (see Figure~\ref{fig:interface-videopoints-0} for the task interface). Crowdworkers first watch the full video clip without audio. For each question, they capture screenshots from the video at the moments when the relevant content is visible. On the screenshot, workers annotate points on object instances that satisfy the question, and we record both the video timestamp and the $(x, y)$ coordinates of all points. Then they answer the corresponding questions in a required format. Workers could mark a question as Unanswerable (e.g., if the content is missing or ambiguous) or flag that they are unsure about their answer. This process is repeated for all questions associated with the video.

To collect annotations for anomaly identification queries in \model-VideoPoint, we first need to construct a dataset of generative videos exhibiting visual defects.  We begin by leveraging two publicly available datasets: the ViBe dataset~\cite{vibe} and the Broken Video Detection Dataset~\cite{broken}. The Broken Video Detection Dataset provides high-quality, frame-level annotations of defective regions, allowing us to directly incorporate its pixel-accurate defect masks. From the ViBe dataset, we selectively retain only videos labeled as Vanishing Subject, Physical Incongruity, or Temporal Dysmorphia. These categories correspond to defects intrinsic to the generated video itself rather than issues arising from ill-posed or misleading prompts, ensuring our dataset focuses on model-induced visual failures.
To complement these sources with realistic user prompts, we sample 2,000 human-written prompts from the VidProM dataset~\cite{vidprom}. For each prompt, we generate videos using 10 T2V models and manually filter the outputs to retain only those containing clear and salient defects. This step introduces diversity in both content and failure types and reflects real-world usage patterns of contemporary text-to-video systems.
In total, our final training set for generative video anomaly pointing consists of 10k videos, covering a broad range of defective generations produced by around 25 T2V models.

\paragraph{\model-VideoTrack.}
Directly reusing the \model-VideoPoint annotation strategy for tracking is infeasible, as it would require point annotations on every sampled frame. One could use off-the-shelf tracking models, such as Co-Tracker~\cite{karaev2024cotracker3} or SAM 2~\cite{ravi2024sam2}, with point prompts; however, we found them to yield incomplete or unstable trajectories and are therefore not reliable sources for generating accurate training data for tracking. We thus resort to existing human-annotated tracks and focus on expanding coverage to video domains and object categories underrepresented in standard training datasets. 

As our base pool, we use a set of videos in video object segmentation (VOS) datasets: SAM-V~\cite{ravi2024sam2}, VIPSeg~\cite{vpseg}, MOSE~\cite{MOSE}, and MOSEv2~\cite{mosev2}, which are not as densely supported in existing academic video track datasets. We discard videos that are shorter than 3 seconds or that contain fewer than three object tracks. We additionally decontaminate videos in MOSE~\cite{MOSE} with respect to the MeViS validation set~\cite{ding2023mevis}; we sample 8 frames per video, extract CLIP ViT-L/14 features~\cite{clip}, and remove any videos whose maximum pairwise frame similarity exceeds 0.95. We then extract points from segmentation masks by computing an alpha-weighted score that combines centroid distance and distance to mask boundaries, which keeps the points near the center while minimizing flickering.

\begin{figure*}[!ht]
    \centering
    \includegraphics[width=.8\textwidth]{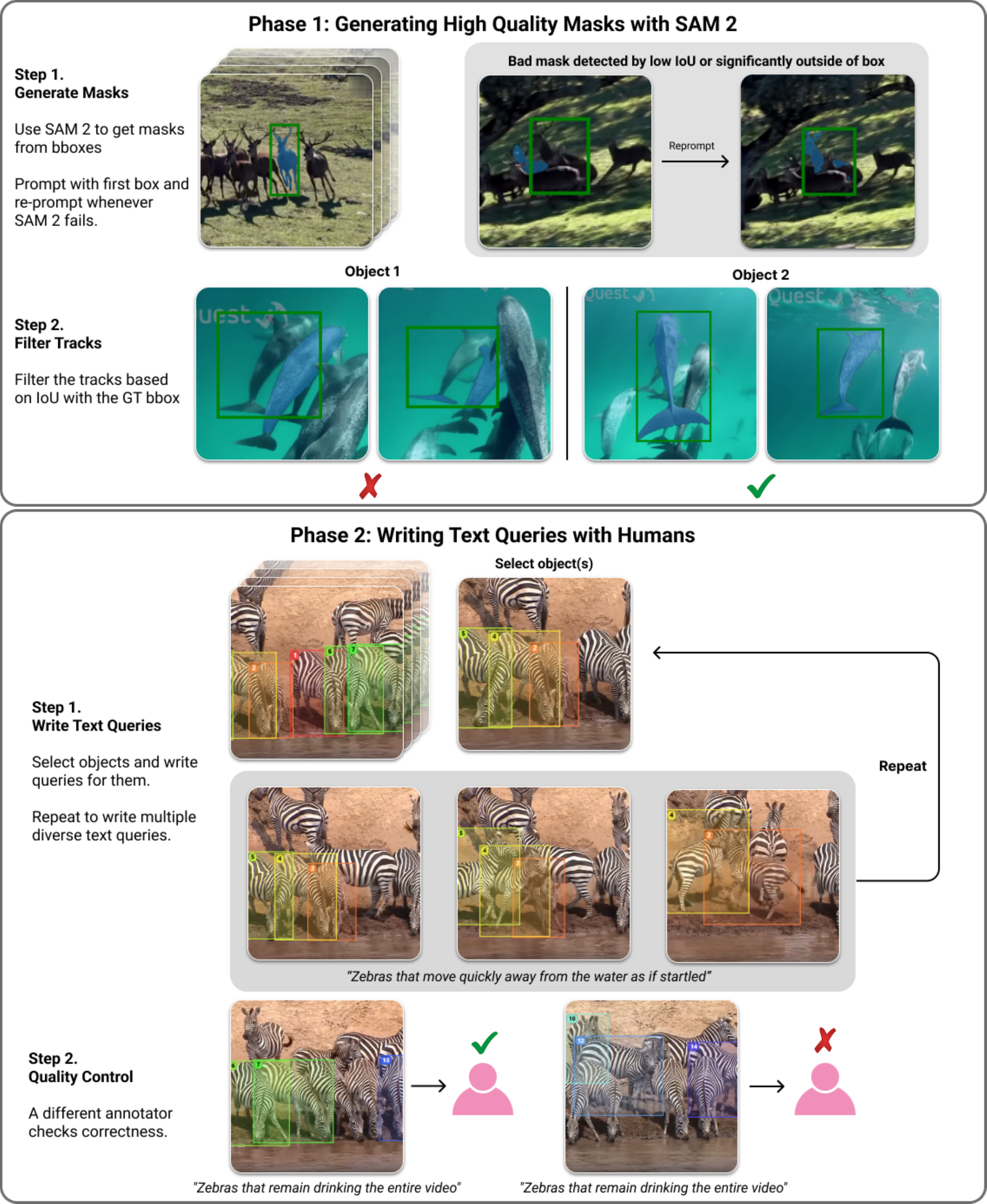}
    \caption{Overview of the annotation pipeline for \textbf{\model-VideoTrack} and the \textbf{\model-Track} benchmark.}
    \label{fig:vid_track_annotation}
\end{figure*}

\begin{figure*}[!t]
    \centering
    \includegraphics[width=0.8\textwidth]{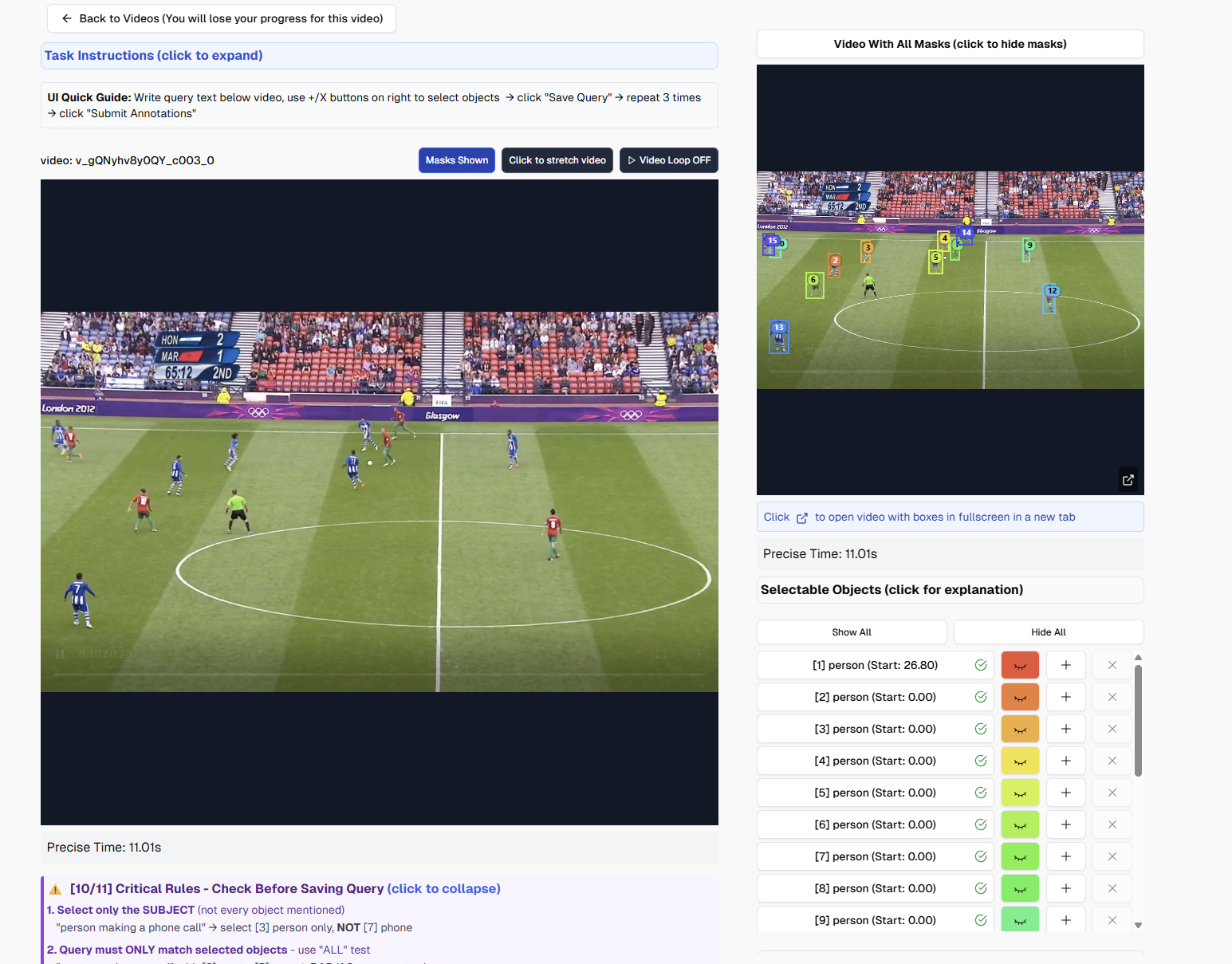}
    \caption{Crowdworkers annotating object text queries. }
    \label{fig:vid_track_write_screenshot}
    \centering
    \includegraphics[width=0.8\textwidth]{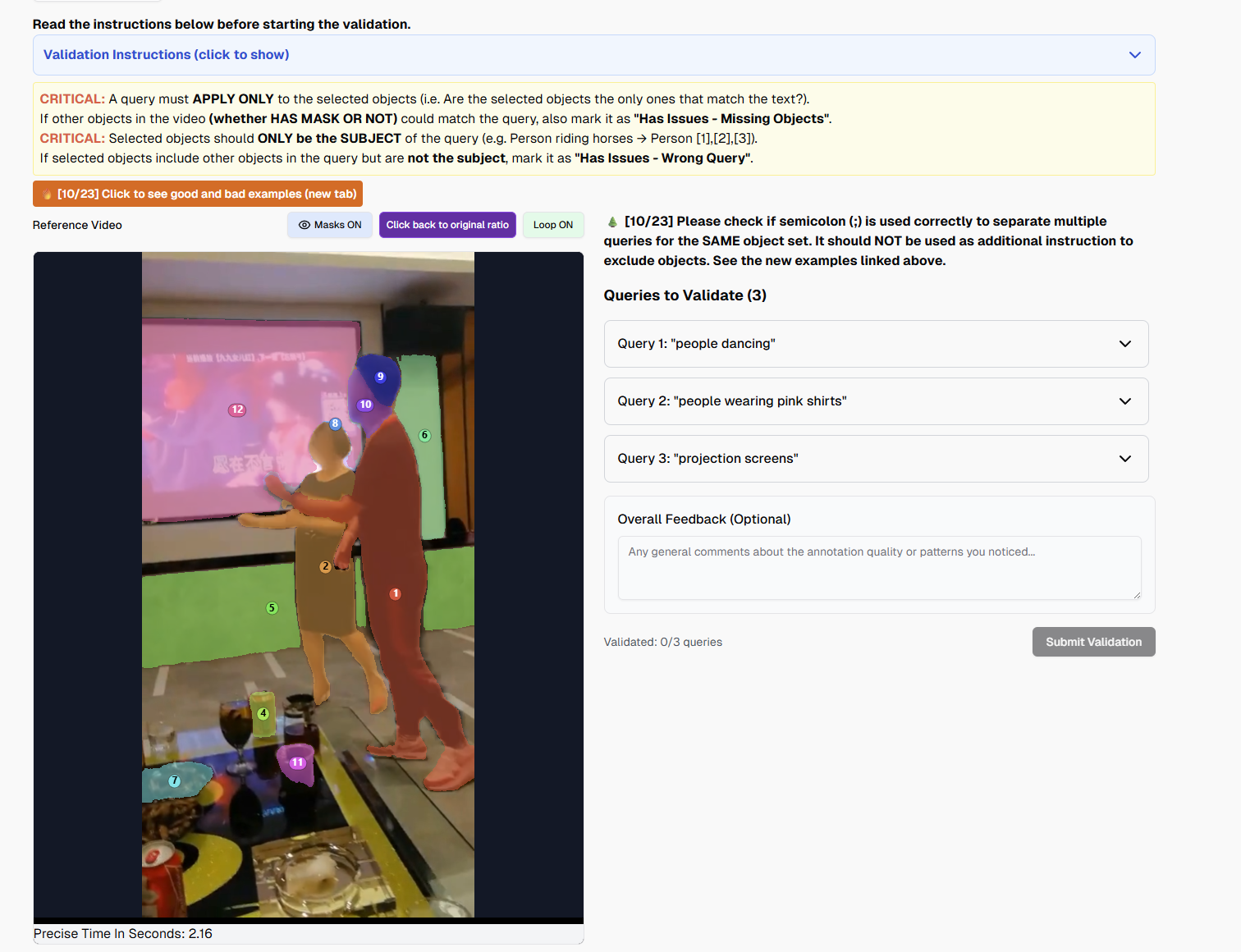}
    \caption{Crowdworkers validating object text queries. }
    \label{fig:vid_track_val_screenshot}
\end{figure*}

We further extend our pool with datasets that provide video object tracks in the form of bounding boxes. These datasets span diverse domains and challenging multi-object scenarios with occlusion, including pedestrians, dancers, autonomous vehicles, animals, athletes, and UAV footage. Unlike in segmentation tracks, naively sampling a (center) point from a bounding box does not guarantee that the point lies on the object. Thus, we convert each bounding-box track into a segmentation task to obtain reliable point tracks. We prompt SAM 2 with the first available bounding box for an object to generate a mask tracklet and propagate this segmentation through the rest of the video. We re-prompt SAM 2 with a new box if the predicted mask has low IoU with the ground truth bounding box or if more than $20\%$ of the mask is outside the bounding box. We filter out object tracks whose predicted segmentation masks have an average IoU below a threshold $0.5$ across all frames. We then apply the same point-sampling procedure on these generated segmentation masks to obtain point tracks. This process is depicted in the first panel of Figure~\ref{fig:vid_track_annotation}, and the annotator interface for this step is shown in Figure~\ref{fig:vid_track_write_screenshot}.

Text descriptions for these tracks are acquired with human annotators. The annotation procedure is illustrated in the second panel of Figure~\ref{fig:vid_track_annotation}, where human annotators are given a video and its list of object tracks and are asked to select one or more objects to write text queries for. The query should describe the selected objects only. The process is repeated $N$ times per video, while ensuring that the set of selected objects is unique for each query. A separate validation round performs quality checks on the annotated text queries. After this filtering, we retain approximately 70\% of the queries on average. This process yields both our training set and the \model-Track benchmark. The annotator interface for validation is shown in Figure~\ref{fig:vid_track_val_screenshot}.

\begin{figure}[!t]
  \centering
  \begin{minipage}[t]{0.48\linewidth}
    \centering
    \includegraphics[width=\linewidth]{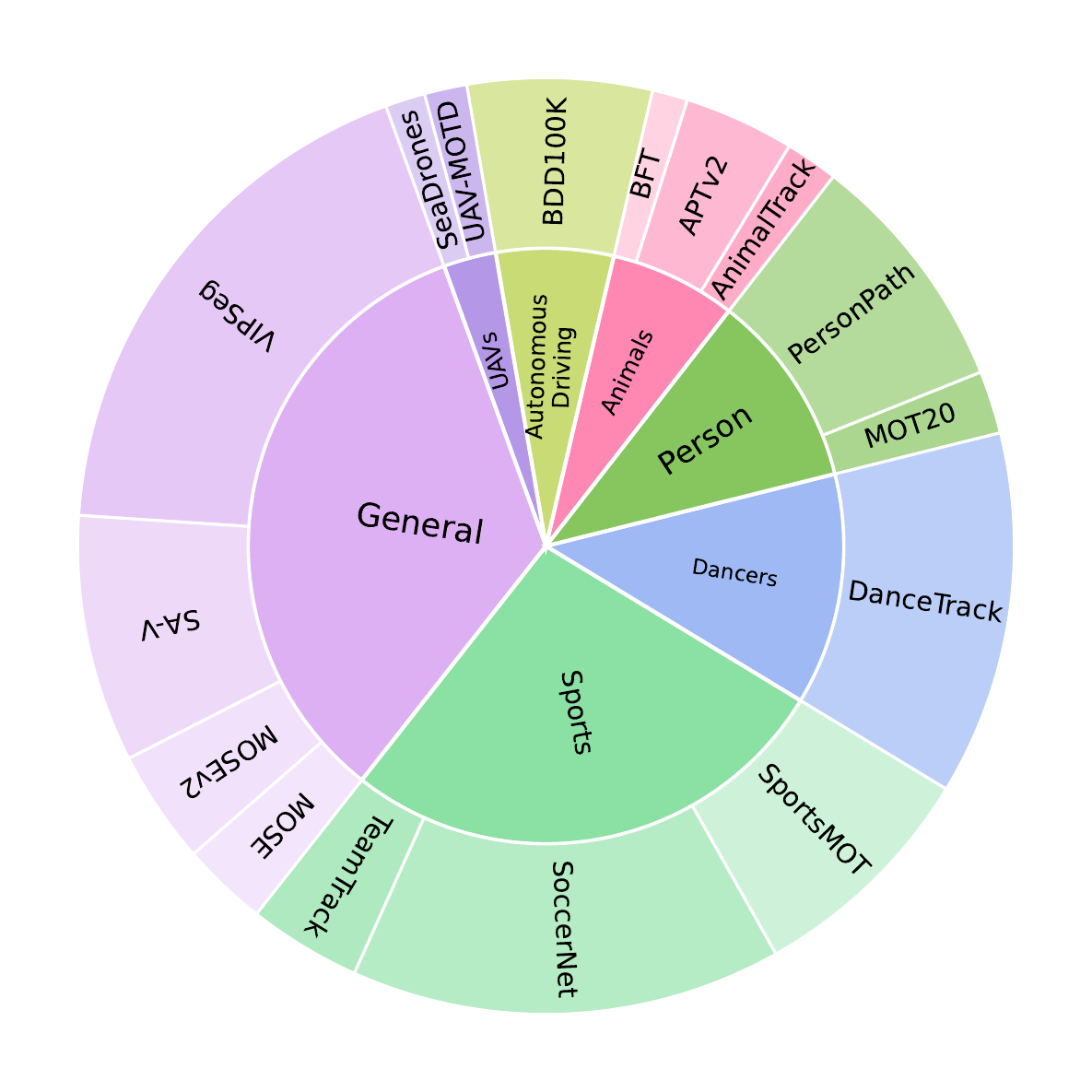}
    \caption{\textbf{\model-VideoTrack} dataset}
    \label{fig:tracking-training-categories}
  \end{minipage}
  \hfill
  \begin{minipage}[t]{0.48\linewidth}
    \centering
    \includegraphics[width=\linewidth]{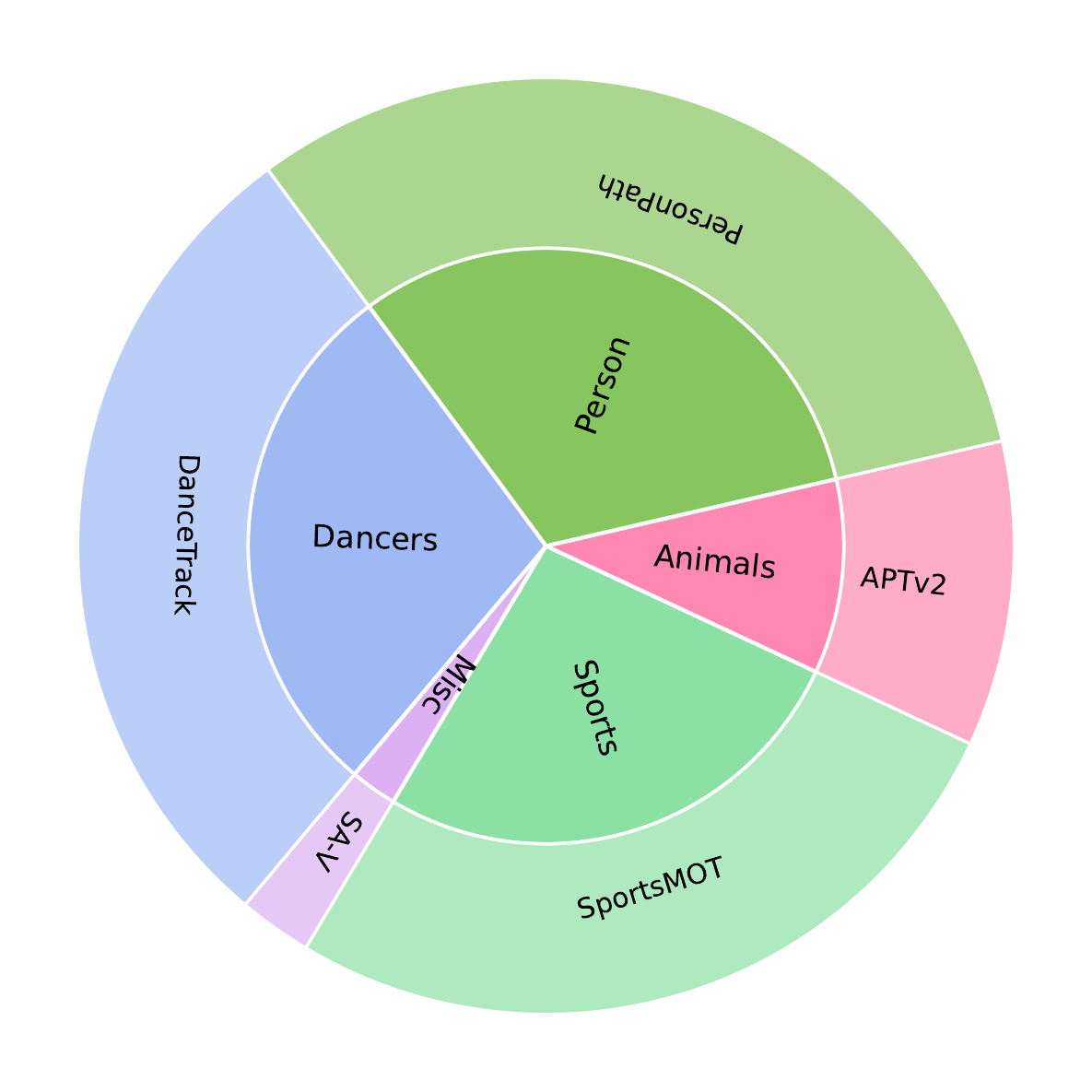}
    \caption{\textbf{\model-Track} benchmark}
    \label{fig:tracking-benchmark-categories}
  \end{minipage}
\end{figure}

Table~\ref{tab:molmo2_video_track_statistics} summarizes the dataset statistics, and Figures~\ref{fig:tracking-training-categories} and~\ref{fig:tracking-benchmark-categories} break down the distribution of queries and objects per semantic category for both training data and \model-Track. The segmentation datasets provide general object tracking across diverse categories, while the bounding-box datasets contribute domain-specific tracking scenarios. Together, these complementary data sources yield a large-scale and diverse corpus for object tracking. 

\begin{table}[!t]
    \centering
    \small

    \begin{subtable}{\columnwidth}
    \centering
    \begin{tabular}{l | c cccc}
        \textbf{Data Source} & \textbf{Type (Ann.)} & \textbf{\# Clips} & \textbf{\# Tracks} & \textbf{\# Queries} & \textbf{Avg \# Obj/Q} \\
        \toprule
        VIPSeg & General (Segm) & 675 & 2,150 & 5,466 & 2.65 \\
        SAM-V & General (Segm) & 1,090 & 2,282 & 2,537 & 1.43 \\
        MOSEv2 & General (Segm) & 463 & 1,107 & 1,168 & 2.08 \\
        MOSE & General (Segm) & 337 & 863 & 880 & 1.91 \\
        TeamTrack & Sports (Bbox) & 154 & 899 & 1,158 & 2.13 \\
        SoccerNet & Sports (Bbox) & 610 & 4109 & 4420 & 6.60 \\
        SportsMOT & Sports (Bbox) & 396 & 2,150 & 2,420 & 4.48 \\
        BDD100K & Auto. Driving (Bbox) & 450 & 1,810 & 1892 & 3.10 \\
        APTv2 & Animals (Bbox) & 401 & 1,051 & 1,132 & 2.68 \\
        AnimalTrack & Animals (Bbox) & 52 & 413 & 542 & 3.59 \\
        BFT & Animals (Bbox) & 30 & 214 & 364 & 2.38 \\
        UAV-MOTD & UAV (Bbox) & 142 & 426 & 437 & 3.43 \\
        SeaDrones & UAV (Bbox) & 79 & 368 & 408 & 2.25 \\
        MOT20 & Person (Bbox) & 147 & 603 & 643 & 2.68 \\
        PersonPath & Person (Bbox) & 1,146 & 2,383 & 2,502 & 1.86 \\
        DanceTrack & Dancers (Bbox) &  704 & 3,199 & 3,735 & 4.07\\
        \midrule
        \textbf{Total} & \textbf{All} & \textbf{6,624} & \textbf{25,437} & \textbf{29,704} & \textbf{3.38} \\
        \bottomrule
    \end{tabular}
    \vspace{2mm}
    \caption{Statistics for the \model-VideoTrack dataset.}
    \label{tab:training_data}
    \vspace{5mm}
\end{subtable}
\begin{subtable}{\columnwidth}
    \centering
    \begin{tabular}{l | c cccc}
        \textbf{Data Source} & \textbf{Type (Ann.)} & \textbf{\# Clips} & \textbf{\# Tracks} & \textbf{\# Queries} & \textbf{Avg \# Obj/Q} \\
        \toprule
        
        APTv2 & Animals (Bbox) & 188 & 331 & 332 & 1.57 \\
        PersonPath & Person (Bbox) & 487 & 958 & 992 & 1.58 \\
        SportsMOT & Sports (Bbox) & 323 & 825 & 838 & 4.03 \\
        DanceTrack & Dancers (Bbox) & 360 & 885 & 905 & 3.11 \\
        SAM-V & Misc (Segm) & 28 & 63 & 80 & 1.21 \\       
        \midrule
        \textbf{Total} & \textbf{All} & \textbf{1,386} & \textbf{3,062} & \textbf{3,147} & \textbf{2.66} \\
        \bottomrule
    \end{tabular}
    \caption{Statistics for the \model-Track benchmark}
    \label{tab:benchmark}
\end{subtable}
    
\caption{\textbf{Distribution of tracking dataset for \model-VideoTrack (train) and \model-Track (benchmark)}. We report the number of unique video clips, unique tracks, total queries, and average number of objects per query (Avg \# Obj/Q) for each dataset. Type indicates video category; Ann. indicates original tracking annotation format (Segm: segmentation masks, Bbox: bounding boxes).}   
\label{tab:molmo2_video_track_statistics}
\end{table}

\paragraph{Academic-VideoTrack.}
We additionally construct an Academic-VideoTrack dataset by aggregating existing academic VOS datasets and bounding-box tracking datasets with referring expressions. Similar to the bounding-box processing for \model-VideoTrack, we convert bounding-box tracks into segmentation mask tracklets by running them through the same pipeline (bounding-box–prompted SAM 2 followed by propagation and IoU-based filtering). 

We also accommodate datasets with non-exhaustive labels, where objects mentioned in the text queries lack corresponding tracks despite appearing in the video. 
 Since these missing objects cannot be used directly for general multi-object tracking, we repurpose them for the ``single-point'' task (Section~\ref{sec:training}), where the model receives a single point on the target object with the associated query and generates its track. This allows us to augment non-exhaustive tracking datasets to our training data and have the model be exposed to diverse, challenging tracking scenarios.

Table~\ref{tab:detailed_datasets} shows the detailed composition of the Academic-VideoTrack dataset used for training.

\begin{figure*}[!t]
  \centering
  \includegraphics[width=0.8\linewidth]{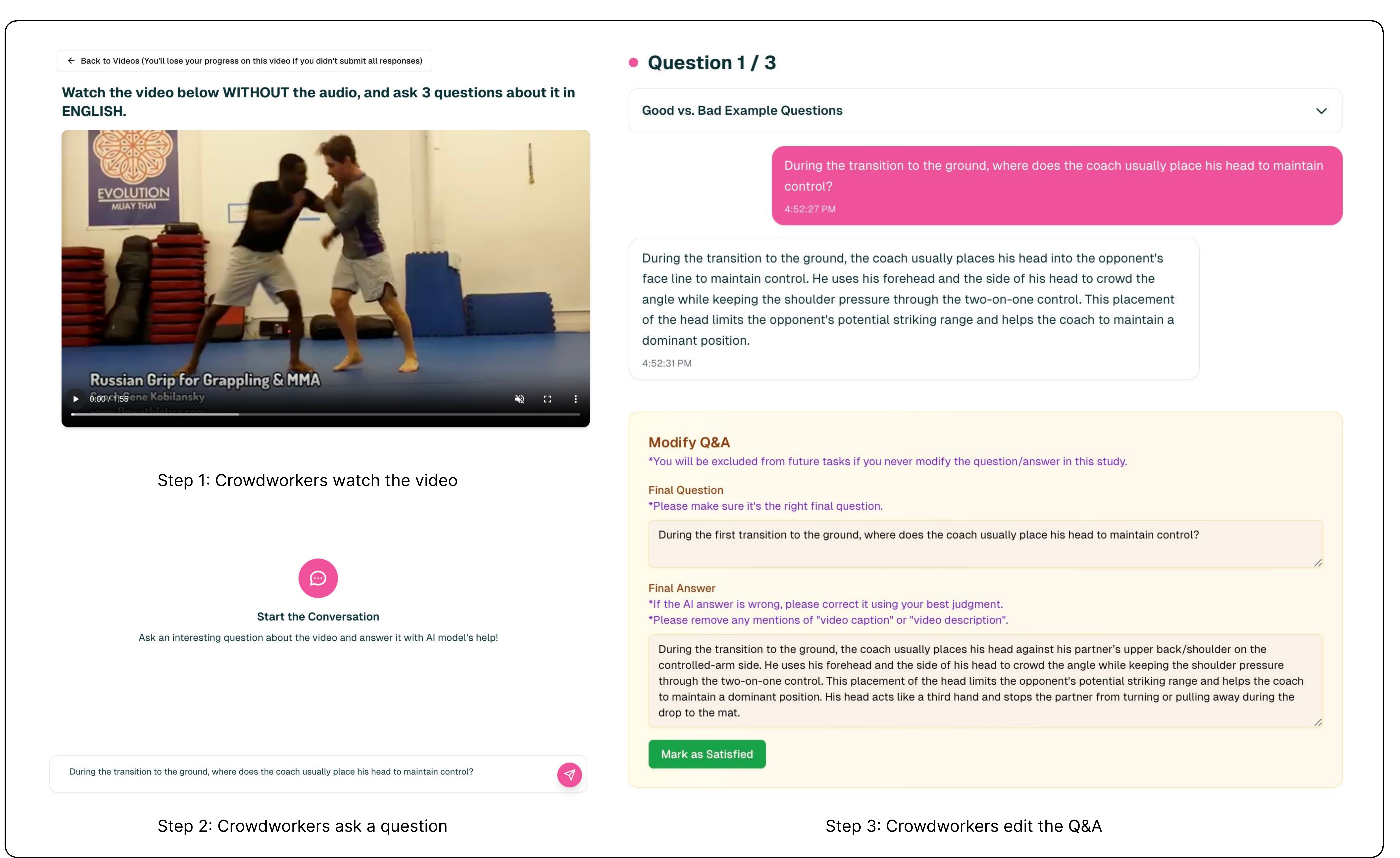}
  \caption{AskModelAnything interface. Crowdworkers are instructed to  ask model non-trivial visual questions and finalize Q\&A.} 
  \label{fig:interface-humanqa-0}
\end{figure*}

\paragraph{\model-AskModelAnything.}
For each video, we first ask crowdworkers to watch the clip without audio and write questions in English that require non-trivial visual reasoning, such as temporal understanding, reading on-screen text details, or identifying fine-grained visual details. We discourage questions that were too vague, too easy or low-level, subjective with no clear ground-truth answer, dependent on unverifiable information such as names or identities, or simple counting questions, which we do not collect for this task. We then feed the full video caption together with the worker's question into a backend language model, which produces an initial answer. Workers are then instructed to slightly edit the question to form a valid query and to carefully edit the model answer to form a final answer. Once they are satisfied, they submit the final Q\&A pair, which we used as our annotation (see Figure~\ref{fig:interface-humanqa-0} for the task interface).


\section{Data examples}
\begin{figure*}
    \includegraphics[width=\linewidth]{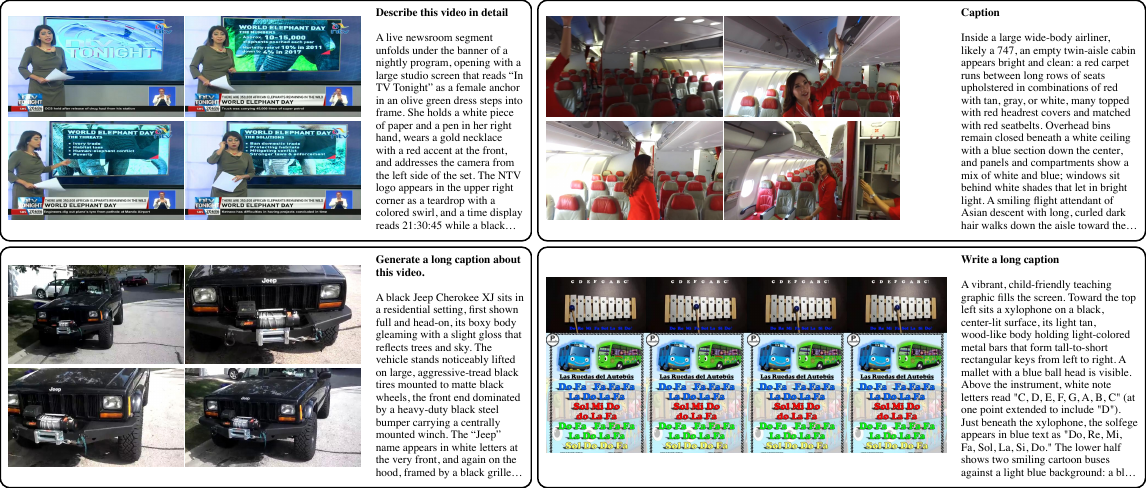}
    \caption{Random examples from \model-Cap. Prompts are generated from our captioning prompt templates.}
    \label{fig:qual_molmo2_cap}
\end{figure*}

\begin{figure*}
    \includegraphics[width=\linewidth]{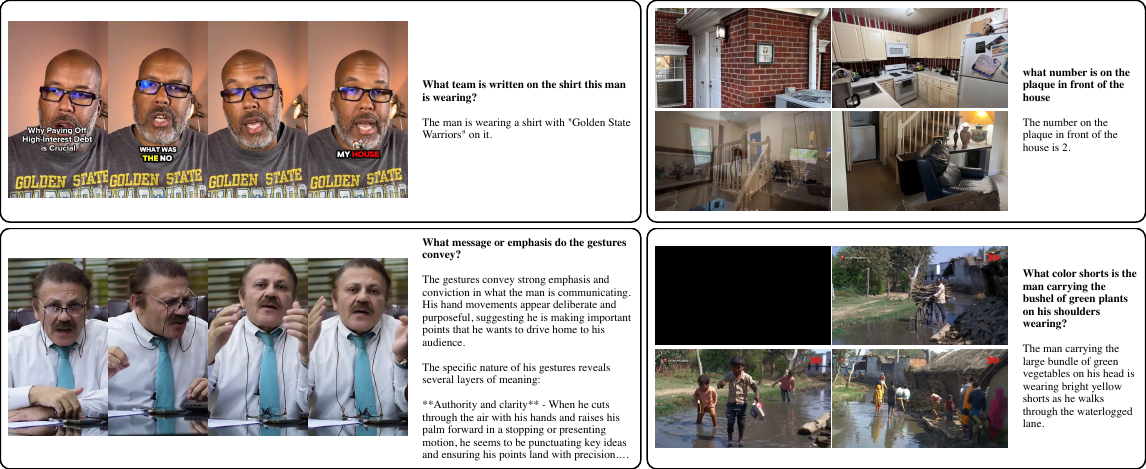}
    \caption{Random examples from \model-AskModelAnything.}
    \label{fig:qual_ask_model_anything}
\end{figure*}

\begin{figure*}
    \includegraphics[width=\linewidth]{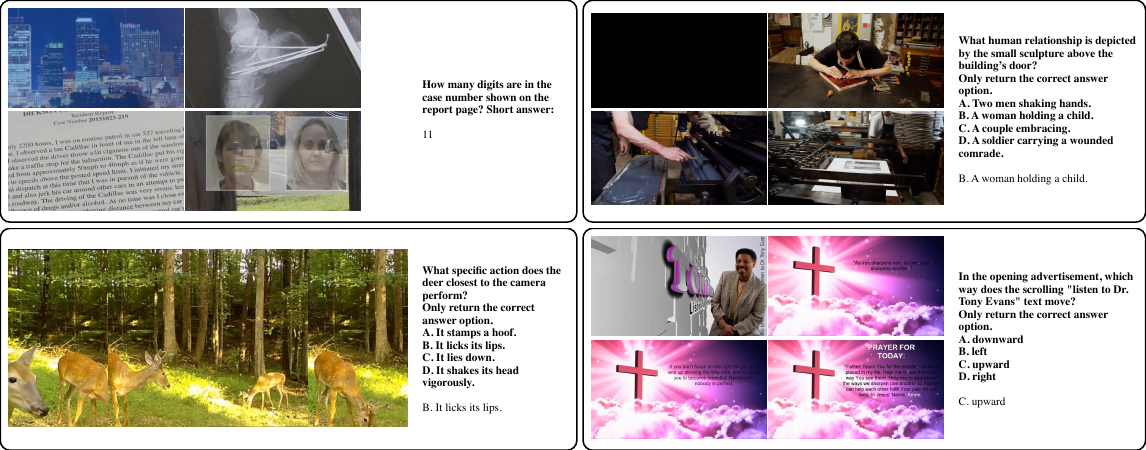}
    \caption{Random examples from \model-CapQA.}
    \label{fig:qual_capqa}
\end{figure*}

\begin{figure*}
    \includegraphics[width=\linewidth]{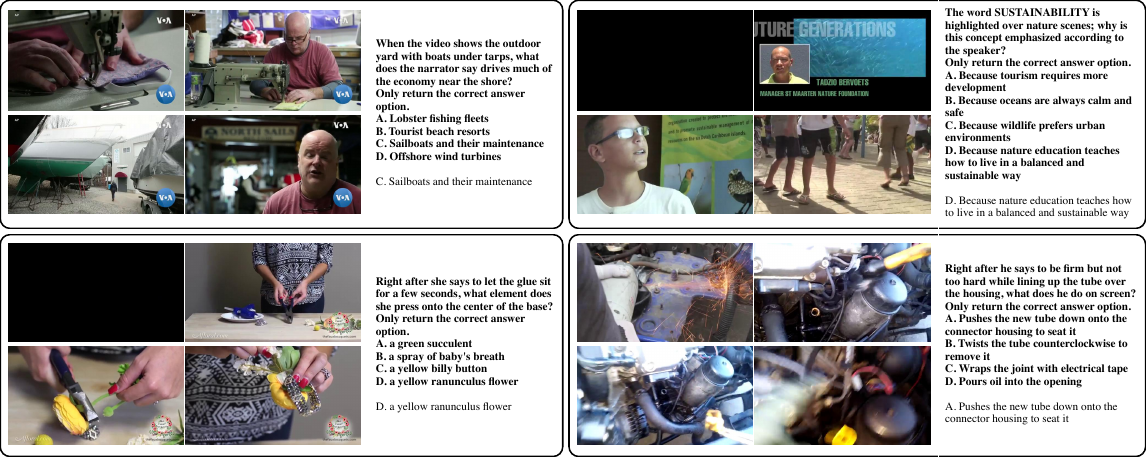}
    \caption{Random examples from \model-SubtitleQA.}
    \label{fig:qual_subtitle_qa}
\end{figure*}

\begin{figure*}
    \includegraphics[width=\linewidth]{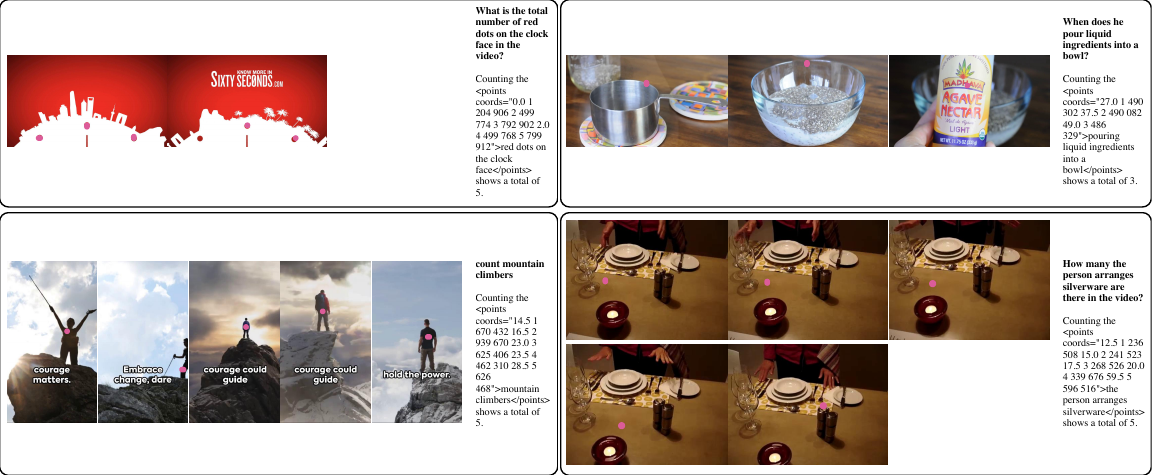}
    \caption{Random examples from \model-VideoPoint. Points are shown in pink, output text follows \model{}'s point formatting.}
        \label{fig:qual_video_pointing}
\end{figure*}

\begin{figure*}
    \includegraphics[width=\linewidth]{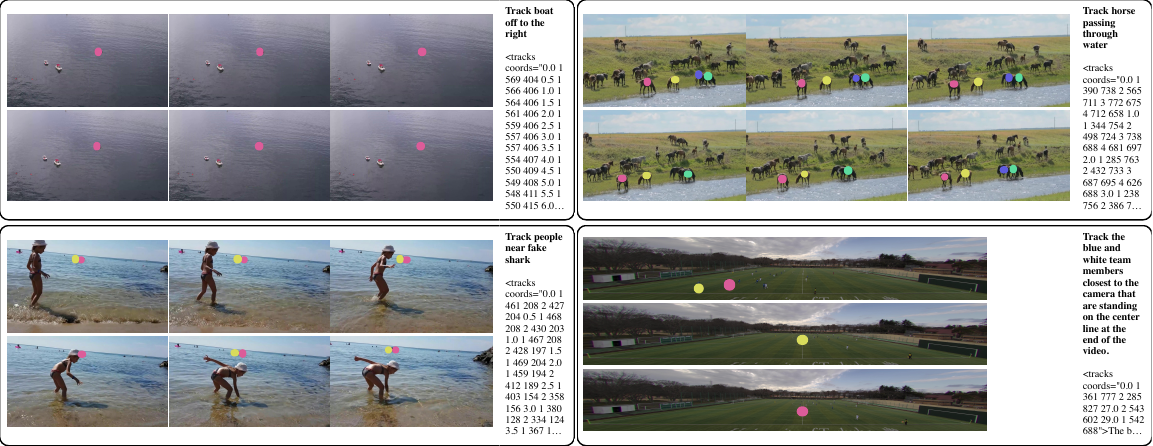}
    \caption{Random examples from \model-VideoTrack. Points are shown in different colors that are shared between the same objects, output text follows \model{}'s point formatting.}
    \label{fig:qual_video_track}
\end{figure*}

\begin{figure*}
    \includegraphics[width=\linewidth]{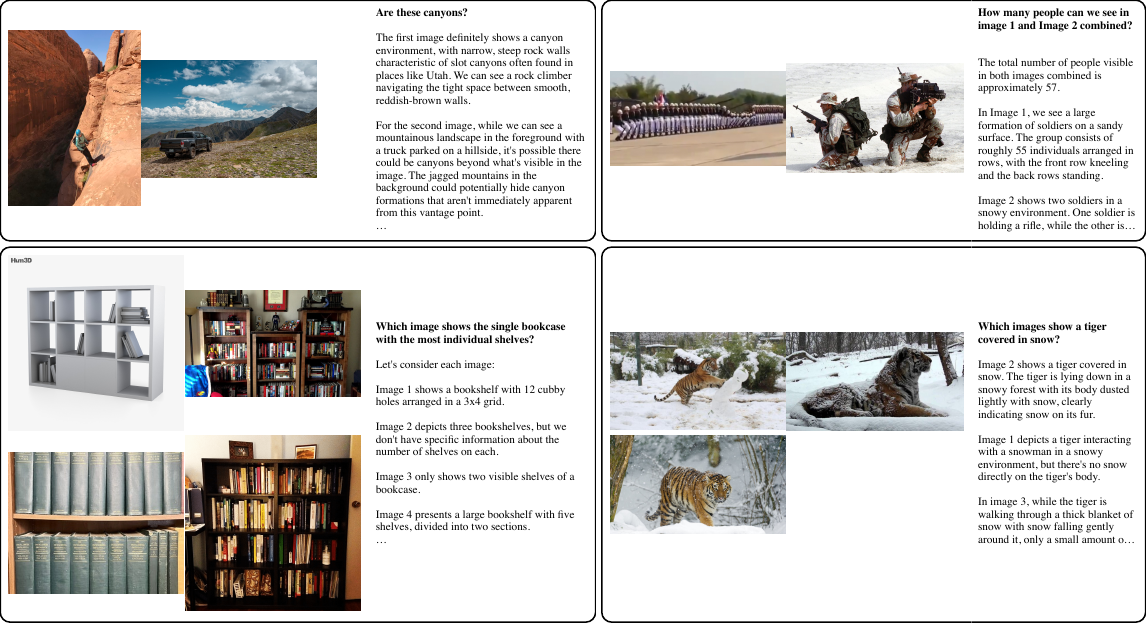}
    \caption{Random examples from \model-MultiImageQA.}
    \label{fig:qual_multi_image_qa}
\end{figure*}

\begin{figure*}
    \includegraphics[width=\linewidth]{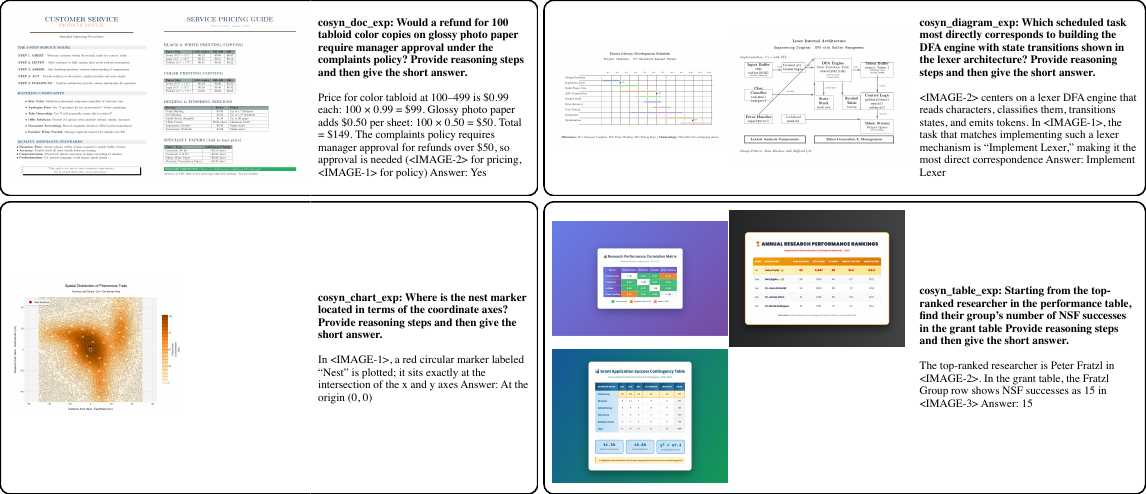}
    \caption{Random examples from \model-SynMultiImageQA.}
    \label{fig:qual_syn_multi_image_qa}
\end{figure*}

\begin{figure*}
    \includegraphics[width=\linewidth]{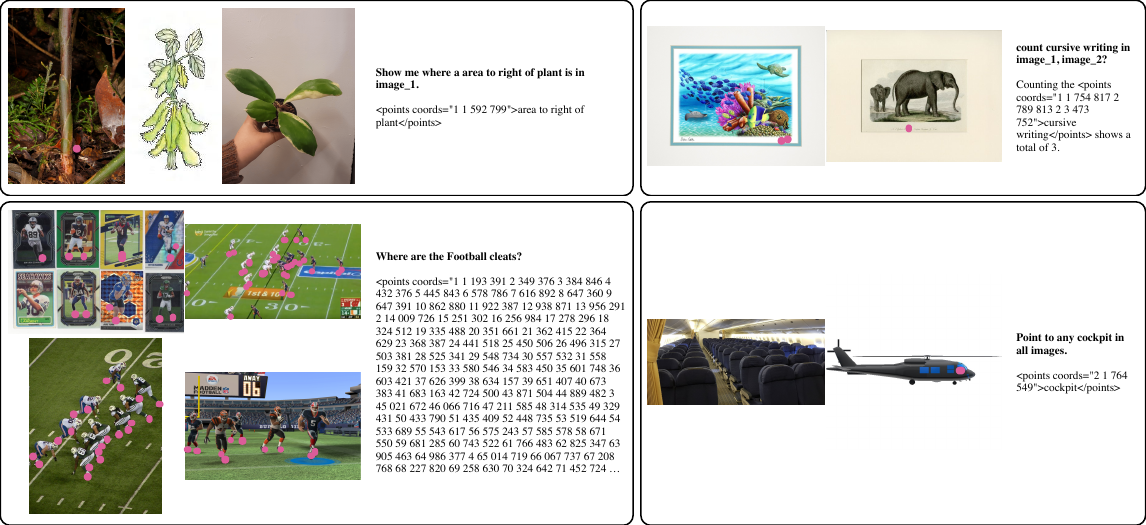}
    \caption{Random examples from \model-MultiImagePoint. Points are shown in pink, output text follows \model{}'s point formatting.}
    \label{fig:qual_multi_image_point}
\end{figure*}

Here, we present qualitative examples from the \model{} datasets. For datasets, we show \textbf{randomly} selected examples. Prompts are in bold, and the target output text is below. Videos are shown using a small number of sampled frames. Examples can be found in:

\begin{itemize}
    \item \model-Cap: Figure~\ref{fig:qual_molmo2_cap}
    \item \model-AskModelAnything: Figure~\ref{fig:qual_ask_model_anything}
    \item \model-CapQA: Figure~\ref{fig:qual_capqa}
    \item \model-SubtitleQA: Figure~\ref{fig:qual_subtitle_qa}
    \item \model-VideoPoint: Figure~\ref{fig:qual_video_pointing}
    \item \model-VideoTrack: Figure~\ref{fig:qual_video_track}
    \item \model-MultiImageQA: Figure~\ref{fig:qual_multi_image_qa}
    \item \model-SynMultiImageQA: Figure~\ref{fig:qual_syn_multi_image_qa}
    \item \model-MultiImagePoint: Figure~\ref{fig:qual_multi_image_point}
\end{itemize}
\label{appendix:qualitative_dataset}


\section{Limitations}
\label{appendix:limitation}
Here we discuss some of the limitations of the \model{} models.

\paragraph{Closed image ViT}. Even with OLMo 3 as the LLM, our models still utilize a closed-data SigLIP 2 image encoder~\cite{siglip2}. We chose to use SigLIP 2 because there are currently no competitive open-data encoders. We call upon the open-source community to explore such alternatives in future work.

\paragraph{Use of closed LLMs}. We use closed text-only LLMs for data generation, as is common practice~\cite{liu2023llava}.
This reduces the transparency of our data collection pipeline. However, we believe that future open LLMs will become sufficiently proficient to be used
in place of closed ones to reproduce this dataset in a fully open manner.
It is still important that we avoid using closed \textit{VLM}s, which would create a circular dependency (training our VLMs would require first building a VLM to generate the training data) and therefore cannot lead to a fully open system in the same way.

\paragraph{Video grounding repeating points}.
For both video tracking and pointing, we sometimes observe that the model produces degenerate outputs, such as a long line of points on one frame or the same point for every frame. 
This is particularly common when pointing to high-frequency objects or on long videos, so this could likely be mitigated by sourcing more training data to better cover these cases.
We also observe the issue is less common in specialized models, so we hypothesize that there might be some interference between the tasks in the joint training mixture which leads to this behavior.

\paragraph{Video grounding}. Video grounding is less consistent than image grounding. Our metrics reflect this, with none of the models we tested reaching more than 40\% on either our counting or pointing metrics, while image models often achieve 70-90\% on image grounding metrics like PointBench. 

We believe this is partly due to the inherent complexity of the task. Video grounding typically requires looking at much more visual content, and pointing at more things, than image grounding. Video grounding also requires re-identification, meaning understanding whether two objects in two different frames are the same object or not, which can be challenging. We also think that the lower resolution typically used when processing long videos, and the fact that the vision encoders are often not pre-trained on videos, could be contributing factors.

\paragraph{Long video grounding}. Grounding has limited support for long (3 minutes+) videos because our grounding training is limited to that length. Handling longer videos is complicated by the fact that we would have to lower the fps when sampling frames to $<2$. This would result in our annotations, which are always at 2 fps, not being aligned with the selected frames. A possible solution is to customize how frames are sampled in these cases to ensure that all grounding annotations are selected.

\paragraph{Point tracking}. 
\model{}'s generated tracks will sometimes change the location of its output point on the target object. This is likely because our tracking data generation pipeline does not always ensure that the point is consistently placed within the target object for every frame. Future improvements in generating points from bounding box or segment mask data could mitigate this issue.

\paragraph{Captioning.} We observe that \model{} can sometimes generate repeating text when generating a very long video caption using greedy decoding.
This is a known issue with LLMs~\cite{holtzman2019curious}, including Qwen3\footnote{\url{https://huggingface.co/Qwen/Qwen3-4B} Best Practices}.
However, we also think that the limited captioning training data contributed, as well as the high length of the captions (we observe that this typically occurs after generating thousands of tokens).
We do not observe this behavior for other tasks.

\section{Qualitative results}
\label{appendix:qualitative}
\begin{figure*}
    \centering
    \includegraphics[width=0.8\linewidth]{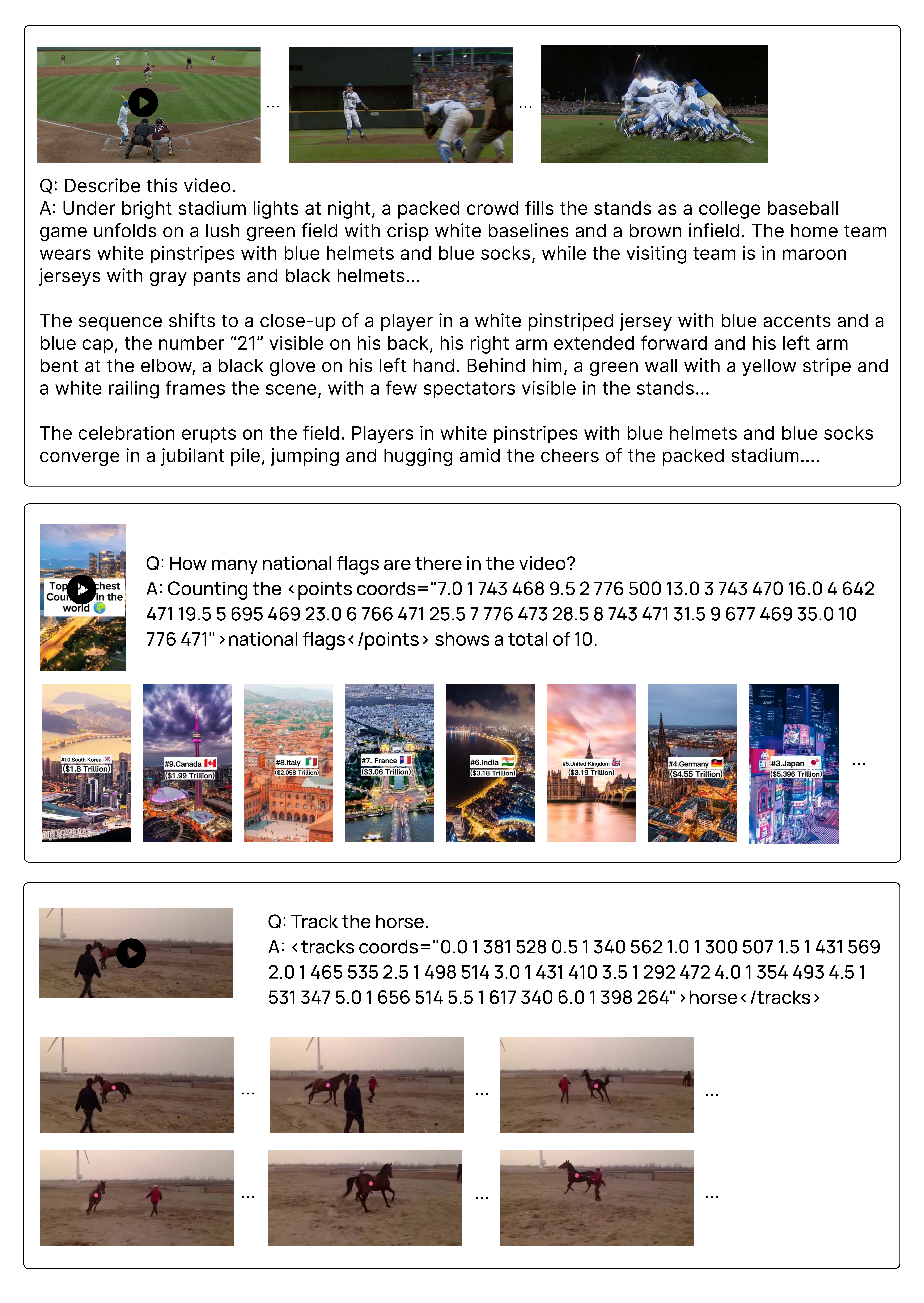}
    \caption{Qualitative examples of captioning, counting, and tracking from \textcolor{molmocolor}{\model{}-8B}}
    \label{fig:model_qual1}
\end{figure*}

\begin{figure*}
    \centering
    \includegraphics[width=0.8\linewidth]{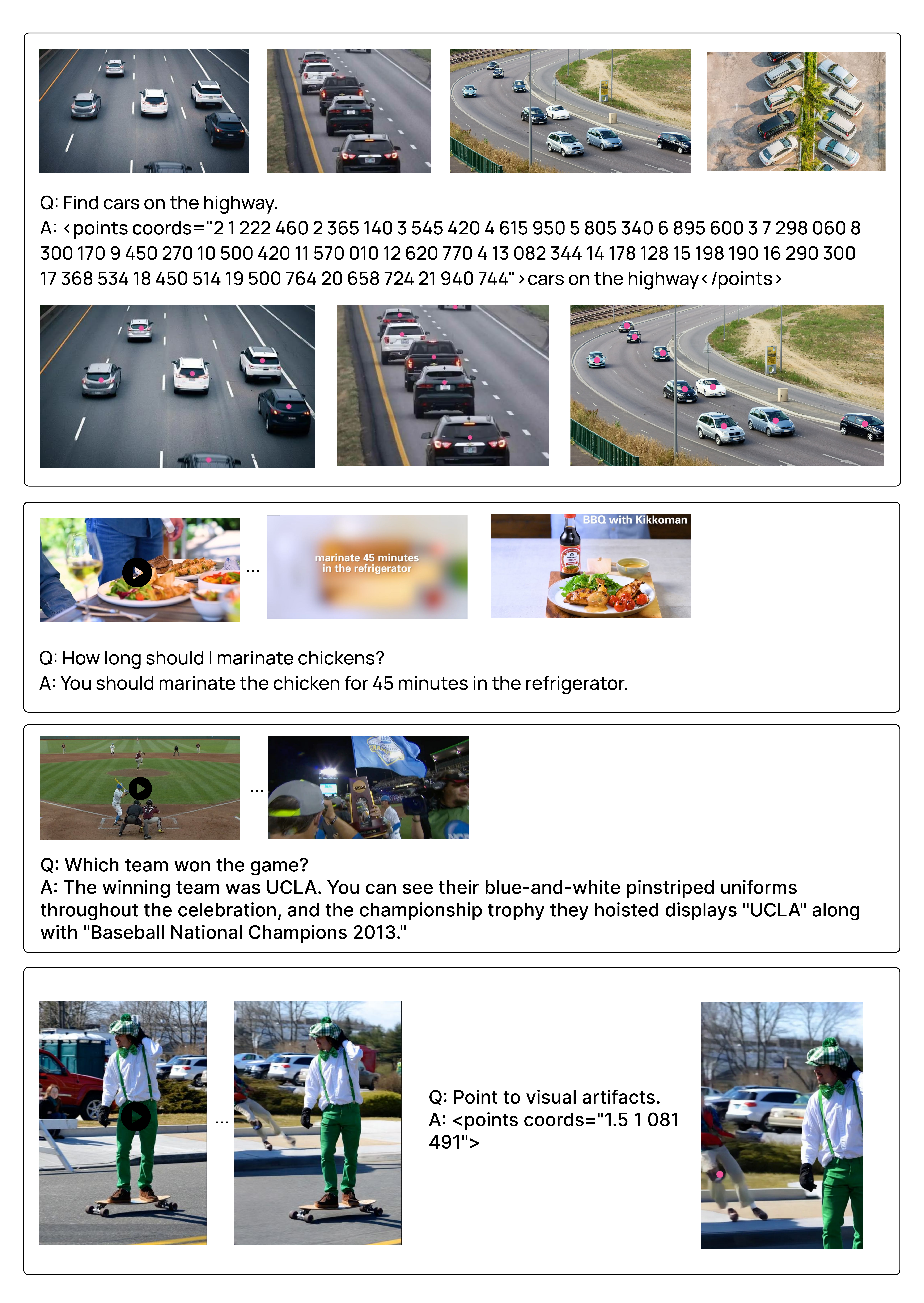}
    \caption{Qualitative examples of pointing and QA from \textcolor{molmocolor}{\model{}-8B}}
    \label{fig:model_qual2}
\end{figure*}

\begin{figure*}
    \centering
    \includegraphics[width=0.8\linewidth]{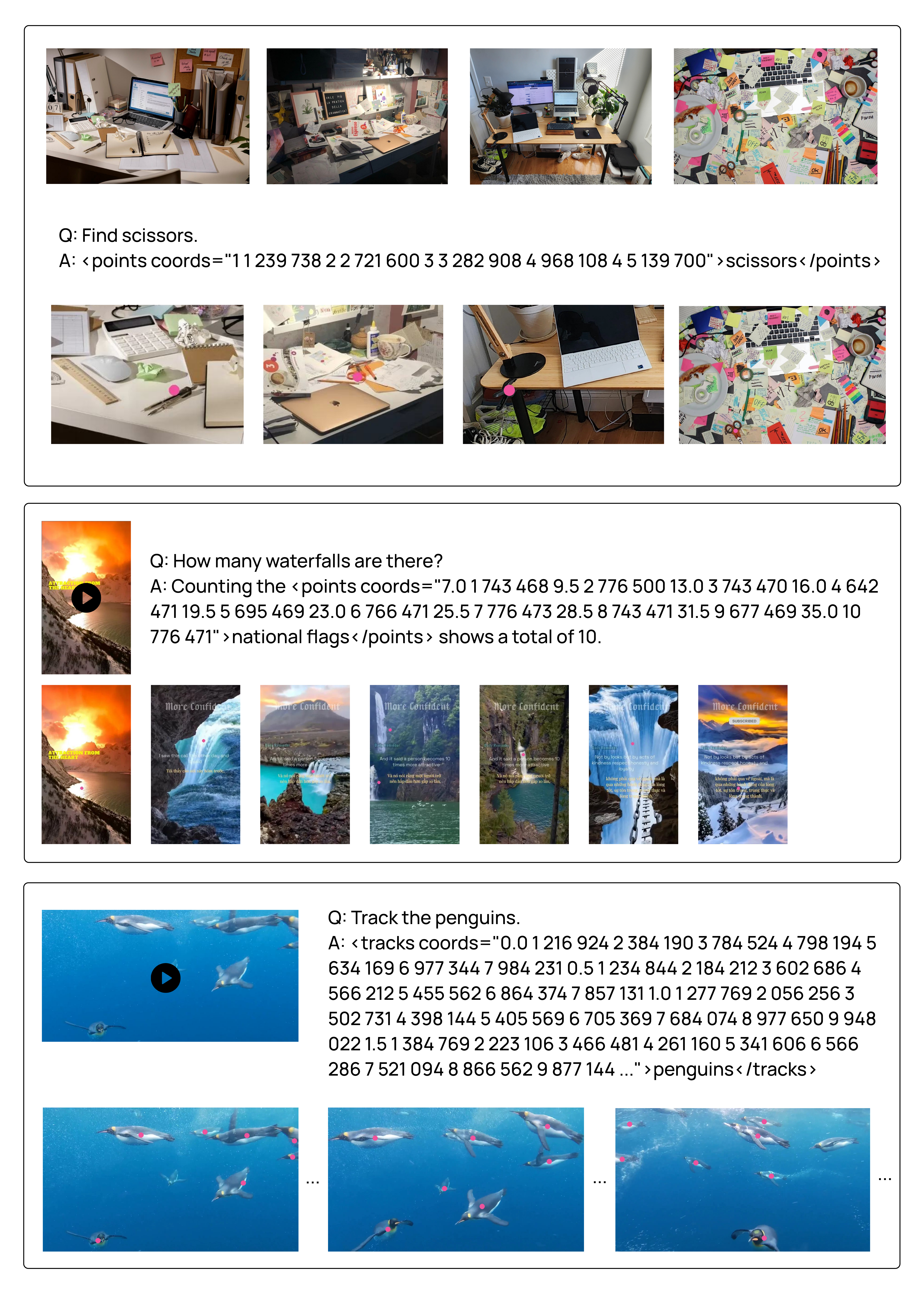}
    \caption{Qualitative failure cases from \textcolor{molmocolor}{\model{}-8B}. The model identifies false positives in the first two examples and misses several of the penguins in the bottom example.}
    \label{fig:model_qual3}
\end{figure*}

We show qualitative examples from \model{}-8B. Each figure shows a query, the response from the model, and selected frames from the input video. The returned points are annotated with pink dots. Successful examples are shown in Figure~\ref{fig:model_qual1} and Figure~\ref{fig:model_qual2}. We also show some failure cases in Figure~\ref{fig:model_qual3}.

\clearpage

\label{appendix:qualitative}

\label{appendix:qualitative}





\end{document}